\documentclass{article}

\PassOptionsToPackage{hyphens}{url}
\PassOptionsToPackage{table}{xcolor}

\usepackage[table,dvipsnames]{xcolor}

\usepackage[preprint]{corl_2026}    

\usepackage{microtype}
\usepackage{amssymb}
\usepackage{amsmath}
\usepackage{mathtools}
\usepackage{amsthm}
\usepackage{booktabs}
\usepackage{graphicx}
\usepackage{caption}
\usepackage{subcaption}

\usepackage{float}
\usepackage{stfloats} 
\usepackage{multirow}
\usepackage{xspace}
\usepackage{pifont}
\usepackage{tabularx}
\usepackage{threeparttable}
\usepackage{array}
\usepackage{wrapfig}
\usepackage[T1]{fontenc}
\usepackage[utf8]{inputenc}
\usepackage{kotex}
\usepackage[table,dvipsnames]{xcolor}
\usepackage{longtable}
\usepackage{lscape}
\usepackage{arydshln} 
\usepackage{xfp} 
\usepackage{siunitx} 
\usepackage{tikz} 

\newif\ifcomments
\commentsfalse   

\definecolor{yspurple}{RGB}{128,0,200}

\makeatletter
\newcommand{\finalonly}[1]{\if@conferencefinal #1\fi\if@preprinttype #1\fi}
\makeatother

\usepackage{xfp} 

\newcommand{\PRobot}{8000}
\newcommand{\PLidar}{3000}
\newcommand{\PGPS}{2000}
\newcommand{\TRobotLife}{2}
\newcommand{\NDelivery}{25}
\newcommand{\NRobotLifeRun}{\fpeval{\TRobotLife * 365 *\NDelivery}}
\newcommand{\PDataCollector}{33}

\newcommand{\PMktRobotDeli}{3.49}
\newcommand{\PRefund}{35.42}
\newcommand{\PMktFood}{\fpeval{\PRefund - \PMktRobotDeli}}
\newcommand{\NFleet}{4}

\newcommand{\CDeliveryFailure}{0.06}
\newcommand{\CRemoteAssistance}{80}
\newcommand{\CPhysicalAssistance}{0.002}
\newcommand{\CRepair}{10--20}
\newcommand{\CRepairCalc}{0.15}
\newcommand{\POperator}{24}
\newcommand{\PElec}{0.2704}
\newcommand{\PMAISReportVehicleWeight}{4535.924}

\newcommand{\PPedDamageAISZero}{380}
\newcommand{\PPedDamageAISOne}{8487}
\newcommand{\PPedDamageAISTwo}{60464}
\newcommand{\PPedDamageAISThree}{261200}

\newcommand{\PPropBollardDamage}{65}
\newcommand{\PPropBuildingGlassDamage}{300}
\newcommand{\PPropMailBoxDamage}{50}
\newcommand{\PPropTrashBinDamage}{50}

\newcommand{\PPropPlantDamage}{10}
\newcommand{\PPropFenceDamage}{50}

\newcommand{\CRobotSpeed}{6.00} 
\newcommand{\CRobotMaxSpeed}{12.80} 
\newcommand{\CRollingResistanceCoeff}{0.0371} 
\newcommand{\CRollingResistanceForce}{18.179} 
\newcommand{\CRobotWeight}{50} 
\newcommand{\TAvgDeliveryTime}{\fpeval{22*(1/60)}} 
\newcommand{\TAvgDeliveryDistance}{4} 
\newcommand{\TMaxDeliveryTime}{1} 
\newcommand{\CAvgDeliveryDistance}{6} 
\newcommand{\EtaCharge}{0.95} %
\newcommand{\EtaBatteryRoundtrip}{0.95} %
\newcommand{\EtaInverter}{0.99} %
\newcommand{\EtaMotor}{0.82} %
\newcommand{\CEnergyConvert}{
    \fpeval{
        \EtaCharge
        * \EtaInverter
        * \EtaBatteryRoundtrip
        * \EtaMotor
    }
} 

\newcommand{\SEvalEpisode}{100}

\newcommand{\SAvgVelocityAMCL}{1.1704}
\newcommand{\SAvgMechanicalPowerAMCL}{0.0212}
\newcommand{\SCollisionImpulseAMCL}{116.35}
\newcommand{\SCollisionDeltaVAMCL}{0.83}

\newcommand{\SAvgVelocityGPS}{1.2625}
\newcommand{\SAvgMechanicalPowerGPS}{0.0229}
\newcommand{\SCollisionImpulseGPS}{82.42}
\newcommand{\SCollisionDeltaVGPS}{0.58}

\newcommand{\SAvgVelocityViNT}{0.3622}
\newcommand{\SAvgMechanicalPowerViNT}{0.0065}
\newcommand{\SCollisionImpulseViNT}{175.61}
\newcommand{\SCollisionDeltaVViNT}{0.85}

\newcommand{\SAvgVelocityGNM}{0.1933}
\newcommand{\SAvgMechanicalPowerGNM}{0.0035}
\newcommand{\SCollisionImpulseGNM}{50.10}
\newcommand{\SCollisionDeltaVGNM}{0.32}

\newcommand{\SAvgVelocityNoMaD}{0.1427}
\newcommand{\SAvgMechanicalPowerNoMaD}{0.0025}
\newcommand{\SCollisionImpulseNoMaD}{579.84}
\newcommand{\SCollisionDeltaVNoMaD}{0.27}

\newcommand{\SAvgVelocityNavDP}{0.0667}
\newcommand{\SAvgMechanicalPowerNavDP}{0.0011}
\newcommand{\SCollisionImpulseNavDP}{31.40}
\newcommand{\SCollisionDeltaVNavDP}{0.26}

\newcommand{\SAvgVelocityCANVAS}{1.1003}
\newcommand{\SAvgMechanicalPowerCANVAS}{0.0200}
\newcommand{\SCollisionImpulseCANVAS}{149.36}
\newcommand{\SCollisionDeltaVCANVAS}{0.41}

\newcommand{\SEpisodeTermSLAAMCL}{0.43}
\newcommand{\SEpisodeTermSpoiledAMCL}{0.12}
\newcommand{\SEpisodeTermTimeoutRobotAMCL}{0.12}
\newcommand{\SEpisodeTermPhysicalAssistanceAMCL}{0.33}

\newcommand{\SEpisodeTermSLAGPS}{0.46}
\newcommand{\SEpisodeTermSpoiledGPS}{0.29}
\newcommand{\SEpisodeTermTimeoutRobotGPS}{0.03}
\newcommand{\SEpisodeTermPhysicalAssistanceGPS}{0.22}

\newcommand{\SEpisodeTermSLAViNT}{0.10}
\newcommand{\SEpisodeTermSpoiledViNT}{0.00}
\newcommand{\SEpisodeTermTimeoutRobotViNT}{0.54}
\newcommand{\SEpisodeTermPhysicalAssistanceViNT}{0.36}

\newcommand{\SEpisodeTermSLAGNM}{0.00}
\newcommand{\SEpisodeTermSpoiledGNM}{0.00}
\newcommand{\SEpisodeTermTimeoutRobotGNM}{0.84}
\newcommand{\SEpisodeTermPhysicalAssistanceGNM}{0.16}

\newcommand{\SEpisodeTermSLANoMaD}{0.01}
\newcommand{\SEpisodeTermSpoiledNoMaD}{0.00}
\newcommand{\SEpisodeTermTimeoutRobotNoMaD}{0.78}
\newcommand{\SEpisodeTermPhysicalAssistanceNoMaD}{0.21}

\newcommand{\SEpisodeTermSLANavDP}{0.00}
\newcommand{\SEpisodeTermSpoiledNavDP}{0.00}
\newcommand{\SEpisodeTermTimeoutRobotNavDP}{0.91}
\newcommand{\SEpisodeTermPhysicalAssistanceNavDP}{0.09}

\newcommand{\SEpisodeTermSLACANVAS}{0.70}
\newcommand{\SEpisodeTermSpoiledCANVAS}{0.10}
\newcommand{\SEpisodeTermTimeoutRobotCANVAS}{0.00}
\newcommand{\SEpisodeTermPhysicalAssistanceCANVAS}{0.20}

\newcommand{\SPedInjuryAMCL}{29.32}
\newcommand{\SPedInjuryGPS}{20.58}
\newcommand{\SPedInjuryViNT}{29.89}
\newcommand{\SPedInjuryGNM}{11.64}
\newcommand{\SPedInjuryNoMaD}{9.76}
\newcommand{\SPedInjuryNavDP}{9.30}
\newcommand{\SPedInjuryCANVAS}{14.38}

\newcommand{\SPropMailBoxContactAMCL}{0}
\newcommand{\SPropTrashBinContactAMCL}{0}
\newcommand{\SPropBuildingGlassContactAMCL}{0.0}
\newcommand{\SPropBollardContactAMCL}{0}

\newcommand{\SPropMailBoxContactGPS}{0.02}
\newcommand{\SPropTrashBinContactGPS}{0}
\newcommand{\SPropBuildingGlassContactGPS}{0}
\newcommand{\SPropBollardContactGPS}{0}

\newcommand{\SPropMailBoxContactViNT}{0}
\newcommand{\SPropTrashBinContactViNT}{0}
\newcommand{\SPropBuildingGlassContactViNT}{0.02}
\newcommand{\SPropBollardContactViNT}{0}

\newcommand{\SPropMailBoxContactGNM}{0.00}
\newcommand{\SPropTrashBinContactGNM}{0.00}
\newcommand{\SPropBuildingGlassContactGNM}{0.01}
\newcommand{\SPropBollardContactGNM}{0.01}

\newcommand{\SPropMailBoxContactNoMaD}{0}
\newcommand{\SPropTrashBinContactNoMaD}{0}
\newcommand{\SPropBuildingGlassContactNoMaD}{0.04}
\newcommand{\SPropBollardContactNoMaD}{0}

\newcommand{\SPropMailBoxContactNavDP}{0}
\newcommand{\SPropTrashBinContactNavDP}{0}
\newcommand{\SPropBuildingGlassContactNavDP}{0}
\newcommand{\SPropBollardContactNavDP}{0}

\newcommand{\SPropMailBoxContactCANVAS}{0.00}
\newcommand{\SPropTrashBinContactCANVAS}{0.00}
\newcommand{\SPropBuildingGlassContactCANVAS}{0.02}
\newcommand{\SPropBollardContactCANVAS}{0.00}

\newcommand{\SRobotSpeed}{7.2}
\newcommand{\STimeout}{0.067}
\newcommand{\SMaxDeliveryDistance}{0.4}
\newcommand{\SRobotMaxMomentum}{640}

\newcommand{\SAvgRunTimeAMCL}{0.026}
\newcommand{\SAvgDeliveryDistanceAMCL}{0.04977}

\newcommand{\SAvgRunTimeGPS}{\fpeval{70.91/3600}}
\newcommand{\SAvgDeliveryDistanceGPS}{\fpeval{86.23/1000}}

\newcommand{\SAvgRunTimeViNT}{\fpeval{106.01/3600}}
\newcommand{\SAvgDeliveryDistanceViNT}{\fpeval{36.81/1000}}

\newcommand{\SAvgRunTimeGNM}{\fpeval{129.63/3600}}
\newcommand{\SAvgDeliveryDistanceGNM}{\fpeval{23.27/1000}}

\newcommand{\SAvgRunTimeNoMaD}{\fpeval{132.50/3600}}
\newcommand{\SAvgDeliveryDistanceNoMaD}{\fpeval{18.48/1000}}

\newcommand{\SAvgRunTimeNavDP}{\fpeval{171.63/3600}}
\newcommand{\SAvgDeliveryDistanceNavDP}{\fpeval{10.90/1000}}

\newcommand{\SAvgRunTimeCANVAS}{\fpeval{103.27/3600}}
\newcommand{\SAvgDeliveryDistanceCANVAS}{\fpeval{90.15/1000}}

\newcommand{\KInjuryAdjustment}{\fpeval{\CRobotWeight / \PMAISReportVehicleWeight}}

\newcommand{\SDataCollectorWorkingTime}{65.63}
\newcommand{\SDataCollectionTime}{40.71}
\newcommand{\SDataCollectionEff}{\fpeval{\SDataCollectionTime / \SDataCollectorWorkingTime}}

\newcommand{\CHardware}{
    \fpeval{ \PRobot + \PLidar + \PGPS}
}

\newcommand{\CHardwareGPS}{
    \fpeval{ \PRobot + \PLidar + \PGPS}
}

\newcommand{\CHardwareViNT}{
    \fpeval{ \PRobot }
}

\newcommand{\CHardwareGNM}{
    \fpeval{ \PRobot }
}

\newcommand{\CHardwareNoMaD}{
    \fpeval{ \PRobot }
}

\newcommand{\CHardwareNavDP}{
    \fpeval{ \PRobot }
}

\newcommand{\CHardwareCANVAS}{
    \fpeval{ \PRobot + \PGPS }
}

\newcommand{\CDataCollection}{
    \fpeval{\PDataCollector * \SDataCollectorWorkingTime}
}

\newcommand{\CElectricityRunAMCL}{
    \fpeval{
        \PElec *
        (\SAvgMechanicalPowerAMCL / \CEnergyConvert) *
        \SAvgRunTimeAMCL
    }
}

\newcommand{\CElectricityRunGPS}{
    \fpeval{
        \PElec *
        (\SAvgMechanicalPowerGPS / \CEnergyConvert) *
        \SAvgRunTimeGPS
    }
}

\newcommand{\CElectricityRunViNT}{
    \fpeval{
        \PElec *
        (\SAvgMechanicalPowerViNT / \CEnergyConvert) *
        \SAvgRunTimeViNT
    }
}

\newcommand{\CElectricityRunGNM}{
    \fpeval{
        \PElec *
        (\SAvgMechanicalPowerGNM / \CEnergyConvert) *
        \SAvgRunTimeGNM
    }
}

\newcommand{\CElectricityRunNoMaD}{
    \fpeval{
        \PElec *
        (\SAvgMechanicalPowerNoMaD / \CEnergyConvert) *
        \SAvgRunTimeNoMaD
    }
}

\newcommand{\CElectricityRunNavDP}{
    \fpeval{
        \PElec *
        (\SAvgMechanicalPowerNavDP / \CEnergyConvert) *
        \SAvgRunTimeNavDP
    }
}

\newcommand{\CElectricityRunCANVAS}{
    \fpeval{
        \PElec *
        (\SAvgMechanicalPowerCANVAS / \CEnergyConvert) *
        \SAvgRunTimeCANVAS
    }
}

\newcommand{\CServiceCompRunAMCL}{
    \fpeval{
        (\SEpisodeTermSpoiledAMCL * \PMktFood) +
        ((\SEpisodeTermTimeoutRobotAMCL + \SEpisodeTermPhysicalAssistanceAMCL) * \PMktRobotDeli)
    }
}

\newcommand{\CServiceCompRunGPS}{
    \fpeval{
        (\SEpisodeTermSpoiledGPS * \PMktFood) +
        ((\SEpisodeTermTimeoutRobotGPS + \SEpisodeTermPhysicalAssistanceGPS) * \PMktRobotDeli)
    }
}

\newcommand{\CServiceCompRunViNT}{
    \fpeval{
        (\SEpisodeTermSpoiledViNT * \PMktFood) +
        ((\SEpisodeTermTimeoutRobotViNT + \SEpisodeTermPhysicalAssistanceViNT) * \PMktRobotDeli)
    }
}

\newcommand{\CServiceCompRunGNM}{
    \fpeval{
        (\SEpisodeTermSpoiledGNM * \PMktFood) +
        ((\SEpisodeTermTimeoutRobotGNM + \SEpisodeTermPhysicalAssistanceGNM) * \PMktRobotDeli)
    }
}

\newcommand{\CServiceCompRunNoMaD}{
    \fpeval{
        (\SEpisodeTermSpoiledNoMaD * \PMktFood) +
        ((\SEpisodeTermTimeoutRobotNoMaD + \SEpisodeTermPhysicalAssistanceNoMaD) * \PMktRobotDeli)
    }
}

\newcommand{\CServiceCompRunNavDP}{
    \fpeval{
        (\SEpisodeTermSpoiledNavDP * \PMktFood) +
        ((\SEpisodeTermTimeoutRobotNavDP + \SEpisodeTermPhysicalAssistanceNavDP) * \PMktRobotDeli)
    }
}

\newcommand{\CServiceCompRunCANVAS}{
    \fpeval{
        (\SEpisodeTermSpoiledCANVAS * \PMktFood) +
        ((\SEpisodeTermTimeoutRobotCANVAS + \SEpisodeTermPhysicalAssistanceCANVAS) * \PMktRobotDeli)
    }
}

\newcommand{\CPedestrianRunAMCL}{
    \fpeval{
        \SPedInjuryAMCL
    }
}

\newcommand{\CPedestrianRunGPS}{
    \fpeval{
        \SPedInjuryGPS
    }
}

\newcommand{\CPedestrianRunViNT}{
    \fpeval{
        \SPedInjuryViNT
    }
}

\newcommand{\CPedestrianRunGNM}{
    \fpeval{
        \SPedInjuryGNM
    }
}

\newcommand{\CPedestrianRunNoMaD}{
    \fpeval{
        \SPedInjuryNoMaD
    }
}

\newcommand{\CPedestrianRunNavDP}{
    \fpeval{
        \SPedInjuryNavDP
    }
}

\newcommand{\CPedestrianRunCANVAS}{
    \fpeval{
        \SPedInjuryCANVAS
    }
}

\newcommand{\CPropertyRunAMCL}{
    \fpeval{
        (
        (\SPropBollardContactAMCL     * \PPropBollardDamage) +
        (\SPropMailBoxContactAMCL    * \PPropMailBoxDamage) +
        (\SPropTrashBinContactAMCL   * \PPropTrashBinDamage) +
        (\SPropBuildingGlassContactAMCL * \PPropBuildingGlassDamage)
        )
    }
}

\newcommand{\CPropertyRunGPS}{
    \fpeval{
        (
        (\SPropBollardContactGPS     * \PPropBollardDamage) +
        (\SPropMailBoxContactGPS    * \PPropMailBoxDamage) +
        (\SPropTrashBinContactGPS   * \PPropTrashBinDamage) +
        (\SPropBuildingGlassContactGPS * \PPropBuildingGlassDamage)
        )
    }
}

\newcommand{\CPropertyRunViNT}{
    \fpeval{
        (
        (\SPropBollardContactViNT     * \PPropBollardDamage) +
        (\SPropMailBoxContactViNT    * \PPropMailBoxDamage) +
        (\SPropTrashBinContactViNT   * \PPropTrashBinDamage) +
        (\SPropBuildingGlassContactViNT * \PPropBuildingGlassDamage)
        )
    }
}

\newcommand{\CPropertyRunGNM}{
    \fpeval{
        (
        (\SPropBollardContactGNM     * \PPropBollardDamage) +
        (\SPropMailBoxContactGNM    * \PPropMailBoxDamage) +
        (\SPropTrashBinContactGNM   * \PPropTrashBinDamage) +
        (\SPropBuildingGlassContactGNM * \PPropBuildingGlassDamage)
        )
    }
}

\newcommand{\CPropertyRunNoMaD}{
    \fpeval{
        (
        (\SPropBollardContactNoMaD     * \PPropBollardDamage) +
        (\SPropMailBoxContactNoMaD    * \PPropMailBoxDamage) +
        (\SPropTrashBinContactNoMaD   * \PPropTrashBinDamage) +
        (\SPropBuildingGlassContactNoMaD * \PPropBuildingGlassDamage)
        )
    }
}

\newcommand{\CPropertyRunNavDP}{
    \fpeval{
        (
        (\SPropBollardContactNavDP     * \PPropBollardDamage) +
        (\SPropMailBoxContactNavDP    * \PPropMailBoxDamage) +
        (\SPropTrashBinContactNavDP   * \PPropTrashBinDamage) +
        (\SPropBuildingGlassContactNavDP * \PPropBuildingGlassDamage)
        )
    }
}

\newcommand{\CPropertyRunCANVAS}{
    \fpeval{
        (
        (\SPropBollardContactCANVAS     * \PPropBollardDamage) +
        (\SPropMailBoxContactCANVAS    * \PPropMailBoxDamage) +
        (\SPropTrashBinContactCANVAS   * \PPropTrashBinDamage) +
        (\SPropBuildingGlassContactCANVAS * \PPropBuildingGlassDamage)
        )
    }
}

\newcommand{\CRepairRunAMCL}{
    \fpeval{
        (\PRobot / \NRobotLifeRun) *
        \CRepairCalc *
        (\SEpisodeTermPhysicalAssistanceAMCL / \CPhysicalAssistance)
    }
}

\newcommand{\CRepairRunGPS}{
    \fpeval{
        (\PRobot / \NRobotLifeRun) *
        \CRepairCalc *
        (\SEpisodeTermPhysicalAssistanceGPS / \CPhysicalAssistance)
    }
}

\newcommand{\CRepairRunViNT}{
    \fpeval{
        (\PRobot / \NRobotLifeRun) *
        \CRepairCalc *
        (\SEpisodeTermPhysicalAssistanceViNT / \CPhysicalAssistance)
    }
}

\newcommand{\CRepairRunGNM}{
    \fpeval{
        (\PRobot / \NRobotLifeRun) *
        \CRepairCalc *
        (\SEpisodeTermPhysicalAssistanceGNM / \CPhysicalAssistance)
    }
}

\newcommand{\CRepairRunNoMaD}{
    \fpeval{
        (\PRobot / \NRobotLifeRun) *
        \CRepairCalc *
        (\SEpisodeTermPhysicalAssistanceNoMaD / \CPhysicalAssistance)
    }
}

\newcommand{\CRepairRunNavDP}{
    \fpeval{
        (\PRobot / \NRobotLifeRun) *
        \CRepairCalc *
        (\SEpisodeTermPhysicalAssistanceNavDP / \CPhysicalAssistance)
    }
}

\newcommand{\CRepairRunCANVAS}{
    \fpeval{
        (\PRobot / \NRobotLifeRun) *
        \CRepairCalc *
        (\SEpisodeTermPhysicalAssistanceCANVAS / \CPhysicalAssistance)
    }
}

\newcommand{\CCAPEXAMCL}{
    \fpeval{ \PRobot + \PLidar + \PGPS}
}

\newcommand{\CCAPEXGPS}{
    \fpeval{ \PRobot + \PLidar + \PGPS}
}

\newcommand{\CCAPEXViNT}{
    \fpeval{ \PRobot + \CDataCollection }
}

\newcommand{\CCAPEXGNM}{
    \fpeval{ \PRobot + \CDataCollection }
}

\newcommand{\CCAPEXNoMaD}{
    \fpeval{ \PRobot + \CDataCollection }
}

\newcommand{\CCAPEXNavDP}{
    \fpeval{ \PRobot + \CDataCollection }
}

\newcommand{\CCAPEXCANVAS}{
    \fpeval{ \PRobot + \PGPS + \CDataCollection }
}

\newcommand{\COPEXRunAMCL}{
    \fpeval{
        \PElec *
        (\SAvgMechanicalPowerAMCL / \CEnergyConvert) *
        \SAvgRunTimeAMCL
        +
        (\SEpisodeTermSpoiledAMCL * \PMktFood) +
        ((\SEpisodeTermTimeoutRobotAMCL + \SEpisodeTermPhysicalAssistanceAMCL) * \PMktRobotDeli)
        +
        \SPedInjuryAMCL
        +
        (
        (\SPropBollardContactAMCL     * \PPropBollardDamage) +
        (\SPropMailBoxContactAMCL    * \PPropMailBoxDamage) +
        (\SPropTrashBinContactAMCL   * \PPropTrashBinDamage) +
        (\SPropBuildingGlassContactAMCL * \PPropBuildingGlassDamage)
        )
        +
        (\PRobot / \NRobotLifeRun) *
        \CRepairCalc *
        (\SEpisodeTermPhysicalAssistanceAMCL / \CPhysicalAssistance)
    }
}

\newcommand{\COPEXRunGPS}{
    \fpeval{
        \PElec *
        (\SAvgMechanicalPowerGPS / \CEnergyConvert) *
        \SAvgRunTimeGPS
        +
        (\SEpisodeTermSpoiledGPS * \PMktFood) +
        ((\SEpisodeTermTimeoutRobotGPS + \SEpisodeTermPhysicalAssistanceGPS) * \PMktRobotDeli)
        +
        \SPedInjuryGPS
        +
        (
        (\SPropBollardContactGPS     * \PPropBollardDamage) +
        (\SPropMailBoxContactGPS    * \PPropMailBoxDamage) +
        (\SPropTrashBinContactGPS   * \PPropTrashBinDamage) +
        (\SPropBuildingGlassContactGPS * \PPropBuildingGlassDamage)
        )
        +
        (\PRobot / \NRobotLifeRun) *
        \CRepairCalc *
        (\SEpisodeTermPhysicalAssistanceGPS / \CPhysicalAssistance)
    }
}

\newcommand{\COPEXRunViNT}{
    \fpeval{
        \PElec *
        (\SAvgMechanicalPowerViNT / \CEnergyConvert) *
        \SAvgRunTimeViNT
        +
        (\SEpisodeTermSpoiledViNT * \PMktFood) +
        ((\SEpisodeTermTimeoutRobotViNT + \SEpisodeTermPhysicalAssistanceViNT) * \PMktRobotDeli)
        +
        \SPedInjuryViNT
        +
        (
        (\SPropBollardContactViNT     * \PPropBollardDamage) +
        (\SPropMailBoxContactViNT    * \PPropMailBoxDamage) +
        (\SPropTrashBinContactViNT   * \PPropTrashBinDamage) +
        (\SPropBuildingGlassContactViNT * \PPropBuildingGlassDamage)
        )
        +
        (\PRobot / \NRobotLifeRun) *
        \CRepairCalc *
        (\SEpisodeTermPhysicalAssistanceViNT / \CPhysicalAssistance)
    }
}

\newcommand{\COPEXRunGNM}{
    \fpeval{
        \PElec *
        (\SAvgMechanicalPowerGNM / \CEnergyConvert) *
        \SAvgRunTimeGNM
        +
        (\SEpisodeTermSpoiledGNM * \PMktFood) +
        ((\SEpisodeTermTimeoutRobotGNM + \SEpisodeTermPhysicalAssistanceGNM) * \PMktRobotDeli)
        +
        \SPedInjuryGNM
        +
        (
        (\SPropBollardContactGNM     * \PPropBollardDamage) +
        (\SPropMailBoxContactGNM    * \PPropMailBoxDamage) +
        (\SPropTrashBinContactGNM   * \PPropTrashBinDamage) +
        (\SPropBuildingGlassContactGNM * \PPropBuildingGlassDamage)
        )
        +
        (\PRobot / \NRobotLifeRun) *
        \CRepairCalc *
        (\SEpisodeTermPhysicalAssistanceGNM / \CPhysicalAssistance)
    }
}

\newcommand{\COPEXRunNoMaD}{
    \fpeval{
        \PElec *
        (\SAvgMechanicalPowerNoMaD / \CEnergyConvert) *
        \SAvgRunTimeNoMaD
        +
        (\SEpisodeTermSpoiledNoMaD * \PMktFood) +
        ((\SEpisodeTermTimeoutRobotNoMaD + \SEpisodeTermPhysicalAssistanceNoMaD) * \PMktRobotDeli)
        +
        \SPedInjuryNoMaD
        +
        (
        (\SPropBollardContactNoMaD     * \PPropBollardDamage) +
        (\SPropMailBoxContactNoMaD    * \PPropMailBoxDamage) +
        (\SPropTrashBinContactNoMaD   * \PPropTrashBinDamage) +
        (\SPropBuildingGlassContactNoMaD * \PPropBuildingGlassDamage)
        )
        +
        (\PRobot / \NRobotLifeRun) *
        \CRepairCalc *
        (\SEpisodeTermPhysicalAssistanceNoMaD / \CPhysicalAssistance)
    }
}

\newcommand{\COPEXRunNavDP}{
    \fpeval{
        \PElec *
        (\SAvgMechanicalPowerNavDP / \CEnergyConvert) *
        \SAvgRunTimeNavDP
        +
        (\SEpisodeTermSpoiledNavDP * \PMktFood) +
        ((\SEpisodeTermTimeoutRobotNavDP + \SEpisodeTermPhysicalAssistanceNavDP) * \PMktRobotDeli)
        +
        \SPedInjuryNavDP
        +
        (
        (\SPropBollardContactNavDP     * \PPropBollardDamage) +
        (\SPropMailBoxContactNavDP    * \PPropMailBoxDamage) +
        (\SPropTrashBinContactNavDP   * \PPropTrashBinDamage) +
        (\SPropBuildingGlassContactNavDP * \PPropBuildingGlassDamage)
        )
        +
        (\PRobot / \NRobotLifeRun) *
        \CRepairCalc *
        (\SEpisodeTermPhysicalAssistanceNavDP / \CPhysicalAssistance)
    }
}

\newcommand{\COPEXRunCANVAS}{
    \fpeval{
        \PElec *
        (\SAvgMechanicalPowerCANVAS / \CEnergyConvert) *
        \SAvgRunTimeCANVAS
        +
        (\SEpisodeTermSpoiledCANVAS * \PMktFood) +
        ((\SEpisodeTermTimeoutRobotCANVAS + \SEpisodeTermPhysicalAssistanceCANVAS) * \PMktRobotDeli)
        +
        \SPedInjuryCANVAS
        +
        (
        (\SPropBollardContactCANVAS     * \PPropBollardDamage) +
        (\SPropMailBoxContactCANVAS    * \PPropMailBoxDamage) +
        (\SPropTrashBinContactCANVAS   * \PPropTrashBinDamage) +
        (\SPropBuildingGlassContactCANVAS * \PPropBuildingGlassDamage)
        )
        +
        (\PRobot / \NRobotLifeRun) *
        \CRepairCalc *
        (\SEpisodeTermPhysicalAssistanceCANVAS / \CPhysicalAssistance)
    }
}

\newcommand{\CRevenueRunAMCL}{
    \fpeval{ \PMktRobotDeli * \SEpisodeTermSLAAMCL }
}

\newcommand{\CRevenueRunGPS}{
    \fpeval{ \PMktRobotDeli * \SEpisodeTermSLAGPS }
}

\newcommand{\CRevenueRunViNT}{
    \fpeval{ \PMktRobotDeli * \SEpisodeTermSLAViNT }
}

\newcommand{\CRevenueRunGNM}{
    \fpeval{ \PMktRobotDeli * \SEpisodeTermSLAGNM }
}

\newcommand{\CRevenueRunNoMaD}{
    \fpeval{ \PMktRobotDeli * \SEpisodeTermSLANoMaD }
}

\newcommand{\CRevenueRunNavDP}{
    \fpeval{ \PMktRobotDeli * \SEpisodeTermSLANavDP }
}

\newcommand{\CRevenueRunCANVAS}{
    \fpeval{ \PMktRobotDeli * \SEpisodeTermSLACANVAS }
}

\newcommand{\CMarginAMCL}{
    \fpeval{
        (\PMktRobotDeli * \SEpisodeTermSLAAMCL)
        -
        (
        \PElec *
        (\SAvgMechanicalPowerAMCL / \CEnergyConvert) *
        \SAvgRunTimeAMCL
        +
        (\SEpisodeTermSpoiledAMCL * \PMktFood) +
        ((\SEpisodeTermTimeoutRobotAMCL + \SEpisodeTermPhysicalAssistanceAMCL) * \PMktRobotDeli)
        +
        \SPedInjuryAMCL
        +
        (
        (\SPropBollardContactAMCL     * \PPropBollardDamage) +
        (\SPropMailBoxContactAMCL    * \PPropMailBoxDamage) +
        (\SPropTrashBinContactAMCL   * \PPropTrashBinDamage) +
        (\SPropBuildingGlassContactAMCL * \PPropBuildingGlassDamage)
        )
        +
        (\PRobot / \NRobotLifeRun) *
        \CRepairCalc *
        (\SEpisodeTermPhysicalAssistanceAMCL / \CPhysicalAssistance)
        )
    }
}

\newcommand{\CMarginGPS}{
    \fpeval{
        (\PMktRobotDeli * \SEpisodeTermSLAGPS)
        -
        (
        \PElec *
        (\SAvgMechanicalPowerGPS / \CEnergyConvert) *
        \SAvgRunTimeGPS
        +
        (\SEpisodeTermSpoiledGPS * \PMktFood) +
        ((\SEpisodeTermTimeoutRobotGPS + \SEpisodeTermPhysicalAssistanceGPS) * \PMktRobotDeli)
        +
        \SPedInjuryGPS
        +
        (
        (\SPropBollardContactGPS     * \PPropBollardDamage) +
        (\SPropMailBoxContactGPS    * \PPropMailBoxDamage) +
        (\SPropTrashBinContactGPS   * \PPropTrashBinDamage) +
        (\SPropBuildingGlassContactGPS * \PPropBuildingGlassDamage)
        )
        +
        (\PRobot / \NRobotLifeRun) *
        \CRepairCalc *
        (\SEpisodeTermPhysicalAssistanceGPS / \CPhysicalAssistance)
        )
    }
}

\newcommand{\CMarginViNT}{
    \fpeval{
        (\PMktRobotDeli * \SEpisodeTermSLAViNT)
        -
        (
        \PElec *
        (\SAvgMechanicalPowerViNT / \CEnergyConvert) *
        \SAvgRunTimeViNT
        +
        (\SEpisodeTermSpoiledViNT * \PMktFood) +
        ((\SEpisodeTermTimeoutRobotViNT + \SEpisodeTermPhysicalAssistanceViNT) * \PMktRobotDeli)
        +
        \SPedInjuryViNT
        +
        (
        (\SPropBollardContactViNT     * \PPropBollardDamage) +
        (\SPropMailBoxContactViNT    * \PPropMailBoxDamage) +
        (\SPropTrashBinContactViNT   * \PPropTrashBinDamage) +
        (\SPropBuildingGlassContactViNT * \PPropBuildingGlassDamage)
        )
        +
        (\PRobot / \NRobotLifeRun) *
        \CRepairCalc *
        (\SEpisodeTermPhysicalAssistanceViNT / \CPhysicalAssistance)
        )
    }
}

\newcommand{\CMarginGNM}{
    \fpeval{
        (\PMktRobotDeli * \SEpisodeTermSLAGNM)
        -
        (
        \PElec *
        (\SAvgMechanicalPowerGNM / \CEnergyConvert) *
        \SAvgRunTimeGNM
        +
        (\SEpisodeTermSpoiledGNM * \PMktFood) +
        ((\SEpisodeTermTimeoutRobotGNM + \SEpisodeTermPhysicalAssistanceGNM) * \PMktRobotDeli)
        +
        \SPedInjuryGNM
        +
        (
        (\SPropBollardContactGNM     * \PPropBollardDamage) +
        (\SPropMailBoxContactGNM    * \PPropMailBoxDamage) +
        (\SPropTrashBinContactGNM   * \PPropTrashBinDamage) +
        (\SPropBuildingGlassContactGNM * \PPropBuildingGlassDamage)
        )
        +
        (\PRobot / \NRobotLifeRun) *
        \CRepairCalc *
        (\SEpisodeTermPhysicalAssistanceGNM / \CPhysicalAssistance)
        )
    }
}

\newcommand{\CMarginNoMaD}{
    \fpeval{
        (\PMktRobotDeli * \SEpisodeTermSLANoMaD)
        -
        (
        \PElec *
        (\SAvgMechanicalPowerNoMaD / \CEnergyConvert) *
        \SAvgRunTimeNoMaD
        +
        (\SEpisodeTermSpoiledNoMaD * \PMktFood) +
        ((\SEpisodeTermTimeoutRobotNoMaD + \SEpisodeTermPhysicalAssistanceNoMaD) * \PMktRobotDeli)
        +
        \SPedInjuryNoMaD
        +
        (
        (\SPropBollardContactNoMaD     * \PPropBollardDamage) +
        (\SPropMailBoxContactNoMaD    * \PPropMailBoxDamage) +
        (\SPropTrashBinContactNoMaD   * \PPropTrashBinDamage) +
        (\SPropBuildingGlassContactNoMaD * \PPropBuildingGlassDamage)
        )
        +
        (\PRobot / \NRobotLifeRun) *
        \CRepairCalc *
        (\SEpisodeTermPhysicalAssistanceNoMaD / \CPhysicalAssistance)
        )
    }
}

\newcommand{\CMarginNavDP}{
    \fpeval{
        (\PMktRobotDeli * \SEpisodeTermSLANavDP)
        -
        (
        \PElec *
        (\SAvgMechanicalPowerNavDP / \CEnergyConvert) *
        \SAvgRunTimeNavDP
        +
        (\SEpisodeTermSpoiledNavDP * \PMktFood) +
        ((\SEpisodeTermTimeoutRobotNavDP + \SEpisodeTermPhysicalAssistanceNavDP) * \PMktRobotDeli)
        +
        \SPedInjuryNavDP
        +
        (
        (\SPropBollardContactNavDP     * \PPropBollardDamage) +
        (\SPropMailBoxContactNavDP    * \PPropMailBoxDamage) +
        (\SPropTrashBinContactNavDP   * \PPropTrashBinDamage) +
        (\SPropBuildingGlassContactNavDP * \PPropBuildingGlassDamage)
        )
        +
        (\PRobot / \NRobotLifeRun) *
        \CRepairCalc *
        (\SEpisodeTermPhysicalAssistanceNavDP / \CPhysicalAssistance)
        )
    }
}

\newcommand{\CMarginCANVAS}{
    \fpeval{
        (\PMktRobotDeli * \SEpisodeTermSLACANVAS)
        -
        (
        \PElec *
        (\SAvgMechanicalPowerCANVAS / \CEnergyConvert) *
        \SAvgRunTimeCANVAS
        +
        (\SEpisodeTermSpoiledCANVAS * \PMktFood) +
        ((\SEpisodeTermTimeoutRobotCANVAS + \SEpisodeTermPhysicalAssistanceCANVAS) * \PMktRobotDeli)
        +
        \SPedInjuryCANVAS
        +
        (
        (\SPropBollardContactCANVAS     * \PPropBollardDamage) +
        (\SPropMailBoxContactCANVAS    * \PPropMailBoxDamage) +
        (\SPropTrashBinContactCANVAS   * \PPropTrashBinDamage) +
        (\SPropBuildingGlassContactCANVAS * \PPropBuildingGlassDamage)
        )
        +
        (\PRobot / \NRobotLifeRun) *
        \CRepairCalc *
        (\SEpisodeTermPhysicalAssistanceCANVAS / \CPhysicalAssistance)
        )
    }
}

\newcommand{\REvalEpisode}{8}

\newcommand{\RSDurationOne}{147.72}
\newcommand{\RSDurationTwo}{97.20}
\newcommand{\RSDurationThree}{78.07}
\newcommand{\RSDurationFour}{129.54}
\newcommand{\RSDurationFive}{132.58}
\newcommand{\RSDurationSix}{124.55}
\newcommand{\RSDurationSeven}{57.30}
\newcommand{\RSDurationEight}{87.69}

\newcommand{\RSDistanceOne}{176.93}
\newcommand{\RSDistanceTwo}{91.73}
\newcommand{\RSDistanceThree}{95.92}
\newcommand{\RSDistanceFour}{151.93}
\newcommand{\RSDistanceFive}{117.52}
\newcommand{\RSDistanceSix}{141.51}
\newcommand{\RSDistanceSeven}{57.83}
\newcommand{\RSDistanceEight}{96.82}

\newcommand{\RSAvgVelocityOne}{0.9573}
\newcommand{\RSAvgVelocityTwo}{0.9278}
\newcommand{\RSAvgVelocityThree}{0.9511}
\newcommand{\RSAvgVelocityFour}{0.9772}
\newcommand{\RSAvgVelocityFive}{0.8328}
\newcommand{\RSAvgVelocitySix}{0.9852}
\newcommand{\RSAvgVelocitySeven}{0.9525}
\newcommand{\RSAvgVelocityEight}{0.9023}

\newcommand{\RSAvgImpulseThree}{50}
\newcommand{\RSAvgImpulseFive}{50}
\newcommand{\RSAvgImpulseSeven}{100}

\newcommand{\RSAvgDeltaVThree}{1.0}
\newcommand{\RSAvgDeltaVFive}{1.0}
\newcommand{\RSAvgDeltaVSeven}{2.0}

\newcommand{\RAvgVelocityCANVAS}{0.9358}
\newcommand{\RAvgMechanicalPowerCANVAS}{0.1383}   
\newcommand{\RCollisionImpulseCANVAS}{66.67}      
\newcommand{\RCollisionDeltaVCANVAS}{1.3333}      

\newcommand{\REpisodeTermSLACANVAS}{0.625}                 
\newcommand{\REpisodeTermSpoiledCANVAS}{0.125}             
\newcommand{\REpisodeTermTimeoutRobotCANVAS}{0.00}
\newcommand{\REpisodeTermPhysicalAssistanceCANVAS}{0.25}   

\newcommand{\RPedInjuryCANVAS}{0}

\newcommand{\RPropPlantContactCANVAS}{0.25}    
\newcommand{\RPropFenceContactCANVAS}{0.125}   

\newcommand{\RAvgRunTimeCANVAS}{\fpeval{106.83/3600}}      
\newcommand{\RAvgDeliveryDistanceCANVAS}{\fpeval{116.28/1000}} 

\newcommand{\RHardwareCANVAS}{\fpeval{\PRobot + \PLidar}}
\newcommand{\RCAPEXCANVAS}{\fpeval{\PRobot + \PLidar + \CDataCollection}}

\newcommand{\RElectricityRunCANVAS}{
    \fpeval{
        \PElec *
        (\RAvgMechanicalPowerCANVAS / \CEnergyConvert) *
        \RAvgRunTimeCANVAS
    }
}

\newcommand{\RServiceCompRunCANVAS}{
    \fpeval{
        (\REpisodeTermSpoiledCANVAS * \PMktFood) +
        ((\REpisodeTermTimeoutRobotCANVAS + \REpisodeTermPhysicalAssistanceCANVAS) * \PMktRobotDeli)
    }
}

\newcommand{\RPedestrianRunCANVAS}{\fpeval{\RPedInjuryCANVAS}}

\newcommand{\RRevenueRunCANVAS}{\fpeval{\PMktRobotDeli * \REpisodeTermSLACANVAS}}

\newcommand{\RPropertyRunCANVAS}{
    \fpeval{
        (\RPropPlantContactCANVAS * \PPropPlantDamage) +
        (\RPropFenceContactCANVAS * \PPropFenceDamage)
    }
}

\newcommand{\RRepairRunCANVAS}{
    \fpeval{
        (\PRobot / \NRobotLifeRun) *
        \CRepairCalc *
        (\REpisodeTermPhysicalAssistanceCANVAS / \CPhysicalAssistance)
    }
}

\newcommand{\ROPEXRunCANVAS}{
    \fpeval{
        \PElec *
        (\RAvgMechanicalPowerCANVAS / \CEnergyConvert) *
        \RAvgRunTimeCANVAS
        +
        (\REpisodeTermSpoiledCANVAS * \PMktFood) +
        ((\REpisodeTermTimeoutRobotCANVAS + \REpisodeTermPhysicalAssistanceCANVAS) * \PMktRobotDeli)
        +
        \RPedInjuryCANVAS
        +
        (\RPropPlantContactCANVAS * \PPropPlantDamage) +
        (\RPropFenceContactCANVAS * \PPropFenceDamage)
        +
        (\PRobot / \NRobotLifeRun) *
        \CRepairCalc *
        (\REpisodeTermPhysicalAssistanceCANVAS / \CPhysicalAssistance)
    }
}

\newcommand{\RMarginCANVAS}{
    \fpeval{
        (\PMktRobotDeli * \REpisodeTermSLACANVAS) - \ROPEXRunCANVAS
    }
}

\definecolor{markgreen}{RGB}{0, 150, 0}
\definecolor{markyellow}{RGB}{230, 160, 0}
\definecolor{markred}{RGB}{220, 0, 0}
\newcommand{\cmark}{\textcolor{markgreen}{\ding{51}}}           
\newcommand{\xmark}{\textcolor{markred}{\ding{55}}}             
\newcommand{\trianglemark}{\textcolor{markyellow}{\ding{115}}}  

\theoremstyle{plain}

\theoremstyle{definition}

\theoremstyle{remark}

\makeatletter
\if@preprinttype
\newcommand{\authorrowstrut}{\rule{\z@}{14\p@}}
\renewcommand{\@maketitle}{%
  \vbox{%
    \hsize\textwidth
    \linewidth\hsize
    \vskip 0.1in
    \centering
    {\LARGE\bf \@title\par}
    \if@conferencefinal
      \def\And{%
        \end{tabular}\hfil\linebreak[0]\hfil%
        \begin{tabular}[t]{c}\bf\authorrowstrut\ignorespaces%
      }
      \def\AND{%
        \end{tabular}\hfil\linebreak[4]\hfil%
        \begin{tabular}[t]{c}\bf\authorrowstrut\ignorespaces%
      }
      \begin{tabular}[t]{c}\bf\authorrowstrut\@author\end{tabular}%
    \else
      \if@preprinttype
        \def\And{%
          \end{tabular}\hfil\linebreak[0]\hfil%
          \begin{tabular}[t]{c}\bf\authorrowstrut\ignorespaces%
        }
        \def\AND{%
          \end{tabular}\hfil\linebreak[4]\hfil%
          \begin{tabular}[t]{c}\bf\authorrowstrut\ignorespaces%
        }
        \begin{tabular}[t]{c}\bf\authorrowstrut\@author\end{tabular}%
      \else
        \begin{tabular}[t]{c}\bf\authorrowstrut
          Anonymous Author(s) \\
          Affiliation \\
          Address \\
          \texttt{email} \\
        \end{tabular}%
      \fi
    \fi
    \vskip 0.3in \@minus 0.1in
  }
}
\fi
\makeatother

\hypersetup{
  pdftitle={CostNav: A Navigation Benchmark for Real-World Economic-Cost Evaluation of Physical AI Agents},
  pdfauthor={Haebin Seong, Sungmin Kim, Yongjun Cho, Myunchul Joe, Geunwoo Kim, Yubeen Park, Sunhoo Kim, Samwoo Seong, Yoonshik Kim, Suhwan Choi, Jaeyoon Jung, Jiyong Youn, Jinmyung Kwak, Sunghee Ahn, Jaemin Lee, Younggil Do, Seungyeop Yi, Woojin Cheong, Minhyeok Oh, Minchan Kim, Seongjae Kang, Youngjae Yu, Yunsung Lee},
}

\title{\emph{CostNav}: A Navigation Benchmark for Real-World Economic-Cost Evaluation of Physical AI Agents}

\author{
  Haebin Seong$^{1*}$
  \And Sungmin Kim$^{1*}$
  \And Yongjun Cho$^{1*}$
  \And Myunchul Joe$^{1}$
  \And Geunwoo Kim$^{3}$
  \And Yubeen Park$^{1}$
  \And Sunhoo Kim$^{1}$
  \And Samwoo Seong$^{1}$
  \And Yoonshik Kim$^{1}$
  \And Suhwan Choi$^{1}$
  \And Jaeyoon Jung$^{1}$
  \And Jiyong Youn$^{1}$
  \And Jinmyung Kwak$^{2}$
  \And Sunghee Ahn$^{4}$
  \And Jaemin Lee$^{4}$
  \And Younggil Do$^{4}$
  \And Seungyeop Yi$^{4}$
  \And Woojin Cheong$^{4}$
  \And Minhyeok Oh$^{4}$
  \And Minchan Kim$^{1}$
  \And Seongjae Kang$^{2}$
  \And Youngjae Yu$^{4\dagger}$
  \And Yunsung Lee$^{1\dagger}$
  \AND
  \normalfont
  $^{1}$MAUM.AI \quad
  $^{2}$KAIST \quad
  $^{3}$University of California, Irvine \quad
  $^{4}$Seoul National University \\
  \texttt{\{youngjaeyu@snu.ac.kr, sung@maum.ai\}} \\
  {\small $^{*}$Equal contribution \quad $^{\dagger}$Corresponding authors}
}

\newcommand{\headercell}[2]{\parbox{#1}{\centering\linespread{0.9}\selectfont #2}}

\begin{document}

\setcounter{bottomnumber}{4}
\setcounter{totalnumber}{6}
\renewcommand{\bottomfraction}{0.6}
\renewcommand{\textfraction}{0.07}

\begin{table}[!b]
    \vspace{-1em}
    \centering
    \caption{Comparison of Robot Navigation Simulation Benchmarks across Physics Fidelity and Economic Cost Modeling.}
    \label{tab:benchmark_comparison}

    \setlength{\tabcolsep}{3pt}
    \footnotesize

    \resizebox{\textwidth}{!}{%
        \begin{tabular}{l l c c c c c c c c c}
            \toprule
                                                            &                    & \multicolumn{3}{c}{\textbf{Physics \& Dynamics}} & \multicolumn{6}{c}{\textbf{Cost \& Economic Modeling}}                                                                                     \\
            \cmidrule(lr){3-5} \cmidrule(lr){6-11}
            \textbf{Benchmark}                              & \textbf{Sim.}      &
            \headercell{1.2cm}{Delivery\\Robot}             &
            \headercell{1.2cm}{Collision\\Dynamics}         &
            \headercell{1.2cm}{Delivery\\Package\\Dynamics} &
            \headercell{1.0cm}{Energy\\Cost}                &
            \headercell{1.2cm}{Pedestrian\\Safety\\Cost}    &
            \headercell{1.2cm}{Property\\Damage\\Cost}      &
            \headercell{1.2cm}{Economic\\Cost\\Model}       &
            \headercell{1.2cm}{Real-World\\Cost Ref.}       &
            \headercell{1.2cm}{BEP\\Analysis}                                                                                                                                                                                                                                    \\
            \midrule
            UnrealZoo                                       & Unreal             & \trianglemark                                    & \trianglemark                                          & \xmark & \xmark        & \xmark & \xmark & \xmark        & \xmark & \xmark        \\
            OpenBench                                       & Gazebo             & \trianglemark                                    & \trianglemark                                          & \xmark & \xmark        & \xmark & \xmark & \xmark        & \xmark & \xmark        \\
            Arena-Rosnav                                    & Gazebo             & \trianglemark                                    & \trianglemark                                          & \xmark & \trianglemark & \xmark & \xmark & \xmark        & \xmark & \xmark        \\
            Urban-Sim                                       & Isaac Lab          & \cmark                                           & \trianglemark                                          & \xmark & \trianglemark & \xmark & \xmark & \xmark        & \xmark & \xmark        \\
            DeliveryBench                                   & Unreal             & \trianglemark                                    & \trianglemark                                          & \xmark & \trianglemark & \xmark & \xmark & \trianglemark & \xmark & \trianglemark \\
            \textbf{CostNav}                                & \textbf{Isaac Sim} & \cmark                                           & \cmark                                                 & \cmark & \cmark        & \cmark & \cmark & \cmark        & \cmark & \cmark        \\
            \bottomrule
        \end{tabular}%
    }
    \vspace{2pt}
    \par\raggedright\scriptsize\cmark~Fully supported, \trianglemark~Partially supported, \xmark~Not supported.
\end{table}

\maketitle


\begin{abstract}
    Current navigation benchmarks focus on task success but do not capture the economic constraints essential for commercializing autonomous delivery systems.
    We introduce \emph{CostNav}, an Economic Navigation Benchmark that evaluates physical AI agents on a cost-revenue and break-even analysis, pairing Isaac Sim's collision and cargo dynamics with industry-standard data such as Securities and Exchange Commission (SEC) filings and Abbreviated Injury Scale (AIS) injury reports.
    To our knowledge, CostNav is the first physics-grounded economic benchmark to use regulatory and financial data to quantify the gap between navigation metrics and commercial deployment, revealing that high task-success rates alone do not ensure economic viability.
    Evaluating seven baselines (two rule-based and five imitation-learning methods), we find no method economically viable: all yield negative contribution margins. CANVAS, using only an RGB camera and GPS, attains the highest task success and the least-negative margin among methods with non-zero Service-Level Agreement (SLA) compliance ($-$\$\num[round-mode=places, round-precision=2]{\fpeval{-1*\CMarginCANVAS}}/run), outperforming LiDAR-equipped Nav2 w/ GPS ($-$\$\num[round-mode=places, round-precision=2]{\fpeval{-1*\CMarginGPS}}/run).
    A sim-trained policy evaluated on a real delivery robot yields SLA compliance close to its simulation result, indicating that policy performance in CostNav's simulation transfers to real-world deployment.
    We challenge the community to achieve economic viability on CostNav, which scores methods by cost-revenue outcomes.
    \finalonly{All resources are available at \url{https://github.com/worv-ai/CostNav}.}
\end{abstract}

\begin{figure}[t]
\centering
\includegraphics[width=0.8\textwidth]{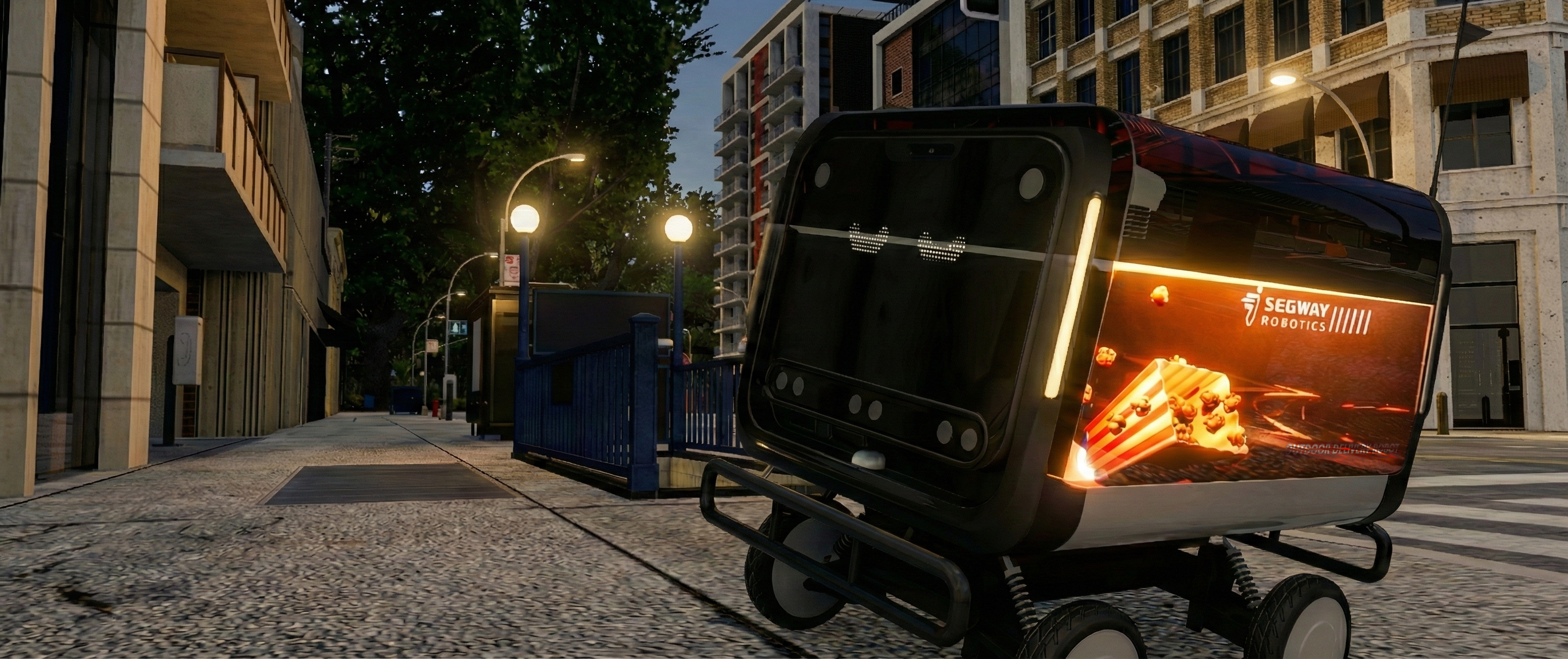}
    \caption{
    \small
    \textbf{A motivational example highlighting the core idea behind the CostNav benchmark.}
    Traditional metrics like success rate or collision rate overlook navigation behaviors that can lead to costly outcomes.
    For instance, overly sharp turning can spill delivery food and cause unnecessary expenses.
    This gap motivates CostNav, which evaluates navigation through an economic lens.
    \textbf{This scene is captured directly from CostNav's simulation environment, with lighting enhancements.}
}
\vspace{-1em}
\label{fig:motivation2}
\end{figure}

\section{Introduction}
\label{sec:intro}

The deployment of autonomous navigation systems has transitioned from research laboratories to real-world commercial applications.
Delivery robots now navigate sidewalks in university campuses and urban areas~\cite{starship-usc2024, starship-dimension, serve-robotics-10k-2023}, autonomous vehicles transport goods in warehouses~\cite{russell2024review}, and mobile robots perform tasks in construction sites~\cite{zeng2024autonomous} and hospitals~\cite{fragapane2020autonomous}.
The adoption of these systems depends on their economic viability; a recent survey finds that decreased operating cost is the principal factor driving the implementation of autonomous delivery robots~\cite{Alverhed2024_ADR_LitReview}.
However, existing academic navigation benchmarks~\cite{unrealzoo2025, openbench2025, arena2025, urbansim2025, deliverybench2025} evaluate embodied physical AI agents primarily on task-oriented metrics such as success rate, collision rate, path length, and navigation time.
While these metrics are valuable, they remain disconnected from the cost-revenue dimensions of real-world deployment, leaving a critical practical question unanswered. \emph{Which navigation approach minimizes cost and maximizes revenue on the path to profitability?}


In this work, we introduce \emph{CostNav}, an \textbf{Economic Navigation Benchmark} designed to evaluate embodied physical AI agents through the lens of economic viability.
CostNav bridges the gap between navigation research and commercial deployment by providing a cost-revenue framework that models the economic lifecycle of autonomous navigation systems.
Our key insight is that real-world navigation performance cannot be captured by task-oriented metrics alone because they ignore the economic cost of operation.
Evaluating navigation for real-world deployment, therefore, requires an economic-cost metric, and we propose \emph{profit per run}, which incorporates both delivery revenue and the cost of operation.
CostNav is initially instantiated in autonomous food delivery on a complex urban sidewalk map using the Segway E1 delivery robot, a commercially deployed last-mile platform, and makes several key contributions:

\noindent\textbf{High-Fidelity Physics Simulation with Sim-to-Real Evidence.}
CostNav grounds its cost metrics in Isaac Sim's high-fidelity physics, deriving them from physical quantities rather than binary collision outcomes.
It maps collision impulse and its variation $\Delta v$ to the Abbreviated Injury Scale (AIS)~\cite{aaam_ais_2015_revision} and crash reports~\cite{wang2022_mais0508_deltav, nhtsa-crash-costs-2019}, and captures jerk-induced food spoilage (\autoref{fig:motivation2}) and pedestrian injury liability, operational costs not typically captured by navigation simulation.
A real-world check supports this grounding, with a sim-trained CANVAS policy evaluated on a physical Segway E1 along an outdoor urban sidewalk route showing SLA compliance close to simulation (\S\ref{sec:sim_to_real}), indicating that policy performance in CostNav's simulation transfers to real-world deployment.

\noindent\textbf{Real-World referenced Cost-Revenue Model with Break-Even Point Analysis.}
We develop an economic model that captures the major cost factors in robot navigation, spanning \emph{pre-run capital expenditure} (CAPEX) and \emph{per-run operating expenditure} (OPEX), and enables Break-Even Point (BEP) analysis that directly answers how long a deployment takes to become profitable.
CostNav integrates revenue modeling based on actual delivery service pricing, modulated by success rate and Service-Level Agreement (SLA) compliance.
Every cost and revenue parameter is drawn from public sources such as SEC filings~\cite{serve-robotics-10k-2023}, AIS injury reports, and commercial delivery pricing~\cite{starship-usc2024}, rather than simulator-defined values.

\noindent\textbf{Open-Source Benchmark.}
We fully open-source our benchmark, including simulation scenarios, real-world referenced cost models, baseline methods, training code and dataset, and evaluation code to facilitate future research at the intersection of physical AI and economic viability.

\section{Related Work}
\label{sec:related}

\autoref{tab:benchmark_comparison} shows a comprehensive comparison of our CostNav benchmark with other benchmarks.

\noindent\textbf{Navigation Benchmarks and Systems.}
Embodied AI simulation has produced indoor benchmarks~\cite{savva2019habitat,kolve2017ai2thor,arena2025}, outdoor robotics simulators~\cite{dosovitskiy2017carla,shah2017airsim,unrealzoo2025}, and recent urban-sidewalk and delivery-oriented suites~\cite{urbansim2025,openbench2025,deliverybench2025} that mostly score agents on task-oriented metrics such as success rate, SPL, and collision-related measures.
Navigation methods relevant to this space span classical navigation/localization stacks~\cite{nav2,amcl} and modern learning-based controllers~\cite{shah2023gnm,shah2023vint,sridhar2023nomad,cai2025navdp,choi2025canvas}, but most benchmarks lack deployment-economics evaluation.
Among these benchmarks, DeliveryBench~\cite{deliverybench2025} scores per-episode VLM-courier profit under simulator-defined cost values, but it does not model autonomous-robot CAPEX/OPEX using industry-referenced operating-cost parameters.
CostNav closes this gap by combining collision dynamics and cargo dynamics with industry-grounded cost references.

\noindent\textbf{Cost-Aware Robotics.}
Existing work optimizes isolated cost components such as energy-aware path planning across ground and aerial platforms~\cite{pmlr-v100-wei20a,maidana2020energy,visca2021conv1d,datsko2024energy}, while domain-specific reviews document deployment economics across warehouse~\cite{russell2024review}, hospital~\cite{fragapane2020autonomous}, last-mile delivery~\cite{Alverhed2024_ADR_LitReview}, and construction~\cite{zeng2024autonomous} robotics.
Industry filings~\cite{starship-commercial-rollout,serve-robotics-10k-2023,serve-robotics-424b4-2024} disclose limited operational figures but keep cost breakdowns proprietary, and none unify these signals into a simulation-driven economic model.

\noindent\textbf{Economic Evaluation of AI.}
Recent digital-agent benchmarks evaluate general interactive capability~\cite{liu2024agentbench,mialon2024gaia,zhou2024webarena} and software-engineering capability~\cite{jimenez2024swebench}, while newer evaluations attach monetary or labor-automation value to software and remote-work tasks~\cite{miserendino2025swe,mazeika2025remote}.
CostNav extends this value-oriented evaluation perspective from digital tasks to embodied navigation, grounding cost--revenue analysis in physics-based robot interactions and real-world business metrics.

\noindent\textbf{Simulation Platforms for Robotics.}
Robotics simulators range from game engines such as Unity~\cite{unity} and Unreal~\cite{unreal2025}, to physics-focused platforms like Gazebo~\cite{gazebo2004}, MuJoCo~\cite{mujoco}, and Isaac Lab~\cite{isaaclab2025}, with recent hybrid approaches combining rendering and physics engines~\cite{embleyriches2025unrealroboticslabhighfidelity}.
CostNav adopts \textbf{NVIDIA Isaac Sim}, which provides unified high-fidelity physics (PhysX 5 and Newton engine) and rendering within a single platform, enabling collision dynamics that directly translate to cost estimation.

\section{CostNav: Cost-Aware Navigation Benchmark for Embodied Agents}
\label{sec:costnav}

\begin{figure}[!ht]
\centering
\includegraphics[width=\textwidth]{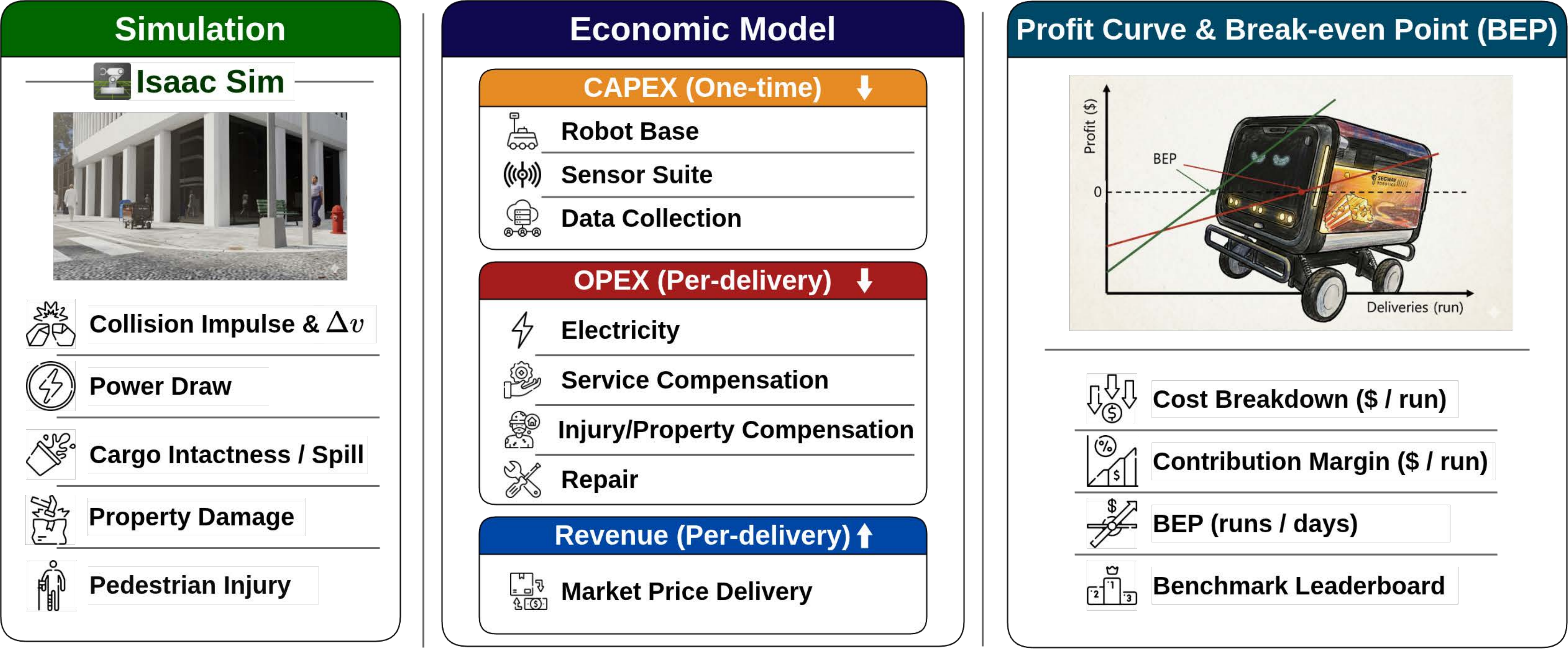}
\caption{
\small
\textbf{End-to-end process of the CostNav benchmark, from simulation environments to break-even analysis.} Simulation logs capture key operational signals that characterize robot behavior in realistic delivery scenarios, which are then integrated with real-world cost and revenue models to compute profit curves and identify each method's break-even point. By translating navigation behavior into economic outcomes, CostNav evaluates embodied agents on financial performance, complementing conventional task-centric metrics.
}
\vspace{-1em}
\label{fig:pipeline2}
\end{figure}

We introduce CostNav, a comprehensive benchmark for evaluating embodied navigation systems through the lens of \emph{economic viability}.
While prior navigation benchmarks primarily focus on geometric efficiency or task success, CostNav models the cost--revenue lifecycle of a delivery robot deployment.
This enables principled comparison of navigation methods based on their real-world business impact.
Figure~\ref{fig:pipeline2} illustrates the end-to-end process of CostNav.
This section presents our economic model (\S\ref{sec:cost-model}) and simulation environment (\S\ref{sec:environment}).

\subsection{Economic Model}
\label{sec:cost-model}

CostNav models the complete economic lifecycle of an autonomous delivery robot, spanning capital expenditure, operational expenditure, and delivery revenue. Our framework explicitly connects simulator-measured quantities to real-world monetary outcomes, enabling direct comparison of navigation methods based on their economic viability. Further details on cost parameters, simulation measurements, and cost-formula derivations are provided in Appendices~\ref{sec:cost_parameters}, \ref{sec:simulation_evaluation}, and \ref{sec:cost_formula}.

\subsubsection{Cost Framework Overview}

We define the total profit $\mathbb{P}$ of a navigation method as:
\begin{equation}
    \mathbb{P} = \mathbb{R} - (C_{\text{CAPEX}} + C_{\text{OPEX}} \times N),
\end{equation}
where $\mathbb{R}$ denotes cumulative revenue, CAPEX represents capital expenditure incurred before deployment, OPEX captures per-delivery operational costs, and $N$ is the number of completed deliveries.

Equivalently, the contribution margin per delivery is:
\begin{equation}
    C_{\text{ContributionMargin}} = R - C_{\text{OPEX}},
\end{equation}
where $R = P_{\text{MktRobotDeli}} \times S_{\text{EpsTermSLA}}$ is the revenue per delivery (base delivery fee $\times$ SLA compliance; zero if timeout), and $C_{\text{OPEX}}$ is the total operational cost per run.

The BEP represents the number of deliveries required to recover capital expenditure:
\begin{equation}
    \text{BEP} = \frac{C_{\text{CAPEX}}}{R - C_{\text{OPEX}}},
\end{equation}
where $R - C_{\text{OPEX}}$ is the Contribution Margin (profit per delivery).
When Contribution Margin is negative ($R < C_{\text{OPEX}}$), the BEP is undefined and the navigation system is not economically viable---each delivery increases cumulative losses.
Lower BEP indicates faster path to profitability, enabling direct comparison of navigation approaches: \emph{Which method recovers its upfront investment fastest?}

\subsubsection{Capital Expenditure (CAPEX)}

Capital expenditure represents fixed upfront investments incurred prior to deployment.

\noindent\textbf{Hardware Cost.}
The upfront hardware investment is:
\begin{equation}
    C_{\text{Hardware}} = P_{\text{Robot}} + (P_{\text{LiDAR}} + P_{\text{GPS}})
\end{equation}
where $P_{\text{Robot}}$ is the robot hardware cost, including onboard sensors, compute hardware, chassis, battery, and supporting electronics.

\noindent\textbf{Data Collection Cost.}
For learning-based navigation, we model the cost of collecting training data:
\begin{equation}
    C_{\text{DataCollection}} = P_{\text{DataCollector}} \times S_{\text{DataCollectionTime}}
\end{equation}
where $P_{\text{DataCollector}}$ is the hourly wage for data collection personnel and $S_{\text{DataCollectionTime}}$ is the total data collection time.
Rule-based methods require occupancy maps and environment surveys; these costs are left for future extensions.

\subsubsection{Operational Expenditure (OPEX)}
\label{sec:opex}

Operational costs are variable expenses incurred on a per-delivery basis.
These costs are directly influenced by navigation behavior, making them a critical link between low-level control decisions and economic outcomes.
We organize OPEX into three categories: direct costs (electricity, repair), service costs (customer compensation), and liability costs (pedestrian and property damage).

\noindent\textbf{Electricity Cost.}
Energy consumption is computed from simulation and converted to monetary cost:
\begin{equation}
    C_{\text{Elec}} = P_{\text{Elec}} \times \frac{S_{\text{AvgPower}}}{C_{\text{ElectroMechanicalEff}}} \times S_{\text{AvgRunTime}}
\end{equation}
where $P_{\text{Elec}}$ is the commercial electricity rate, $S_{\text{AvgPower}}$ is the average power draw measured from simulation, $C_{\text{ElectroMechanicalEff}}$ is the grid-to-wheel efficiency factor, and $S_{\text{AvgRunTime}}$ is the average run time in simulation.

\noindent\textbf{Repair Cost.}
Robot repairs arise from physical damage during operation, particularly \emph{physical-assistance events}---collisions that tip the robot over or accumulate impulse beyond a recoverable threshold, requiring a human operator to physically recover it:
\begin{equation}
    C_{\text{RepairRun}} = \frac{P_{\text{Robot}}}{N_{\text{RobotLifeRun}}} \times C_{\text{Repair}} \times \frac{S_{\text{EpsTermPhys}}}{C_{\text{PhysicalAssistance}}}
\end{equation}
where $P_{\text{Robot}}$ is the robot hardware cost, $N_{\text{RobotLifeRun}}$ is the expected lifetime deliveries per robot, $C_{\text{Repair}}$ is the annual repair rate as a fraction of hardware cost, and $S_{\text{EpsTermPhys}}/C_{\text{PhysicalAssistance}}$ represents the ratio of simulated physical assistance events to the real-world baseline rate.

\noindent\textbf{Service Compensation Cost.}
Failed deliveries incur compensation costs for customer refunds:
\begin{equation}
    C_{\text{ServiceCompRun}} = S_{\text{EpsTermSpoiled}} \times P_{\text{MktFood}} + (S_{\text{EpsTermTimeout}} + S_{\text{EpsTermPhys}}) \times P_{\text{MktRobotDeli}}
\end{equation}
where $S_{\text{EpsTermSpoiled}}$ is the food spoilage rate (delivery arrived but cargo damaged), $P_{\text{MktFood}}$ is the average food cost requiring refund, $S_{\text{EpsTermTimeout}}$ is the timeout rate, $S_{\text{EpsTermPhys}}$ is the rate of physical-assistance events, and $P_{\text{MktRobotDeli}}$ is the delivery fee refunded for timeout failures.
This cost reflects real-world refund policies where late or damaged deliveries receive full or partial compensation.

\noindent\textbf{Pedestrian Safety Cost.}
\label{sec:pedestrian_cost}
We model pedestrian injury costs using the AIS levels~\cite{aaam_ais_2015_revision}, measured from $\Delta v$ using $P(\text{AIS} \mid \Delta v)$ from~\cite{wang2022_mais0508_deltav}, with economic costs per AIS severity from~\cite{nhtsa-crash-costs-2019}:
\begin{equation}
    C_{\text{PedestrianRun}} = K_{\text{InjuryAdjustment}} \sum_{\text{AIS}} P(\text{AIS} \mid \Delta v) \cdot P_{\text{PedDamageAIS}}
\end{equation}
where $P(\text{AIS} \mid \Delta v)$ is the probability of AIS severity given impact speed, $P_{\text{PedDamageAIS}}$ is the economic cost per incident AIS severity level, and $K_{\text{InjuryAdjustment}}$ is an adjustment factor to adjust for differences in vehicle weight between the delivery robot and the reference vehicle in the crash report. See Appendix~\ref{app:pedestrian_proxy} for the scope and limitations of this proxy.

\noindent\textbf{Property Damage Cost.}
Collisions with urban infrastructure incur repair or replacement costs:
\begin{equation}
    C_{\text{PropertyRun}} = \sum_{j} S_{\text{PropContact}_j} \times P_{\text{PropDamage}_j}
\end{equation}
where $S_{\text{PropContact}_j}$ is a count for contact with property type $j$, and $P_{\text{PropDamage}_j}$ is the associated repair or replacement cost. The property types considered are $j \in \{\text{bollard, mailbox, trash bin, building glass}\}$, selected for their prevalence on urban sidewalks and vulnerability to robot collisions, with costs derived from municipal replacement rates.

\subsection{Simulation Environment}
\label{sec:environment}

To evaluate the economic performance of each navigation method, CostNav simulates a Segway E1 delivery robot on complex urban sidewalks in NVIDIA Isaac Sim, chosen for its high-fidelity physics.
The map is a 200\,m $\times$ 200\,m sidewalk network spanning distinct zones such as construction sites, work areas, and standard sidewalks, providing diverse and complex environments and scenarios.
It is populated with pedestrians at configurable densities and with the urban infrastructure obstacles our cost model prices (Appendix~\ref{sec:cost_parameters}), so pedestrian and property contacts are produced as the agent navigates, grounding the pedestrian-injury and property-damage costs of \S\ref{sec:opex} in its actual behavior.

We evaluate every navigation method on a single Segway E1 platform, whose sensor options span the configurations these methods require.
We adopt it because it is a commercially deployed last-mile delivery robot whose hardware specifications and itemized component pricing are both publicly documented~\cite{segway-e1-specs, segway-price}.
The published specifications enable us to reproduce its physical and sensing characteristics in simulation, while the listed prices for the base robot and its optional LiDAR and GPS modules ground per-configuration capital expenditure in real market values.

\section{Results}
\label{sec:results}

\subsection{Evaluation Details}

We evaluate each method over \SEvalEpisode\ popcorn delivery episodes on the urban sidewalk environment described in \S\ref{sec:environment} (Fig.~\ref{fig:motivation2}), comparing seven baselines in two categories. The rule-based baselines run Nav2~\cite{nav2} under two localization configurations, AMCL~\cite{amcl} and GPS-only. The learning-based baselines are five end-to-end policies: GNM~\cite{shah2023gnm}, ViNT~\cite{shah2023vint}, NoMaD~\cite{sridhar2023nomad}, NavDP~\cite{cai2025navdp}, and CANVAS~\cite{choi2025canvas}, all finetuned on our \SDataCollectionTime\ hr CostNav teleoperation dataset. Appendix~\ref{sec:baseline_details} summarizes the per-baseline configurations in Table~\ref{tab:baselines} and provides their full details. To accommodate more complex avoidance maneuvers, we modify CANVAS to predict command velocities end-to-end through the FAST tokenizer~\cite{pertsch2025fast}, replacing its original waypoint tokens and the separate PD controller.

\subsection{Simulation Results}

We report the simulation outcomes in two tables. Table~\ref{tab:simulation-eval} reports the raw measurements aggregated across all evaluation episodes, covering the physical, safety, and outcome quantities defined in Section~\ref{sec:cost-model}; full details are in Appendix~\ref{sec:simulation_evaluation}. Table~\ref{tab:overall} converts these measurements into the economic metrics of CAPEX, OPEX components, revenue, and contribution margin, using the real-world cost parameters listed in Appendix~\ref{sec:cost_parameters}; full per-method derivations are in Appendix~\ref{sec:cost_formula}.


\noindent\textbf{Economic Viability.} All seven methods yield negative contribution margins, so none reach break-even. Under our current cost model, the largest modeled OPEX component is pedestrian safety cost, which ranges from \$\num[round-mode=places, round-precision=2]{\CPedestrianRunNavDP}/run (NavDP) to \$\num[round-mode=places, round-precision=2]{\CPedestrianRunViNT}/run (ViNT) across baselines, with CANVAS at \$\num[round-mode=places, round-precision=2]{\CPedestrianRunCANVAS}/run. In terms of overall contribution margin, ViNT performs worst at $-$\$\num[round-mode=places, round-precision=2]{\fpeval{-1*\CMarginViNT}}/run. NavDP's least negative margin at $-$\$\num[round-mode=places, round-precision=2]{\fpeval{-1*\CMarginNavDP}}/run does not indicate profitable navigation: the policy barely moves, producing \num[round-mode=places, round-precision=0]{\fpeval{\SEpisodeTermTimeoutRobotNavDP*100}}\% timeout and \num[round-mode=places, round-precision=0]{\fpeval{\SEpisodeTermSLANavDP*100}}\% SLA compliance, so both revenue and OPEX stay near zero, and the resulting margin is slightly negative.

\noindent\textbf{Observations on Nav2 and CANVAS.} We analyze each of the three highest-performing baselines individually: Nav2 w/ AMCL (\num[round-mode=places, round-precision=0]{\fpeval{\SEpisodeTermSLAAMCL*100}}\% SLA), Nav2 w/ GPS-only (\num[round-mode=places, round-precision=0]{\fpeval{\SEpisodeTermSLAGPS*100}}\%), and CANVAS (\num[round-mode=places, round-precision=0]{\fpeval{\SEpisodeTermSLACANVAS*100}}\%). Nav2 pairs a prebuilt occupancy map with a classical planner and a LiDAR-based costmap for obstacle avoidance, which is robust on the static sidewalk geometry. CANVAS instead learns an end-to-end policy that produces velocity commands directly from RGB observations and a sketch goal. Training on human teleoperation demonstrations collected under the explicit instruction to drive smoothly so that the food being delivered does not spill has two effects: the policy learns the velocity required to reach the destination, attaining a timeout rate of \num[round-mode=places, round-precision=0]{\fpeval{\SEpisodeTermTimeoutRobotCANVAS*100}}\%, the lowest among all baselines; and at that same velocity, it learns to drive without spilling the food. Both Nav2 configurations use a DWB local planner whose critic weights are tuned to maintain forward velocity in order to avoid timeouts, so encountering a pedestrian causes the planner to sample trajectories that hold forward speed and avoid laterally instead of decelerating; the resulting abrupt evasive maneuvers raise the food-spoilage rate to \num[round-mode=places, round-precision=0]{\fpeval{\SEpisodeTermSpoiledGPS*100}}\% in the GPS-only variant.

\noindent\textbf{Failure Analysis.} GNM, NoMaD, and NavDP achieve near-zero SLA compliance (\num[round-mode=places, round-precision=0]{\fpeval{\SEpisodeTermSLAGNM*100}}\%, \num[round-mode=places, round-precision=0]{\fpeval{\SEpisodeTermSLANoMaD*100}}\%, \num[round-mode=places, round-precision=0]{\fpeval{\SEpisodeTermSLANavDP*100}}\%) with timeout rates of \num[round-mode=places, round-precision=0]{\fpeval{\SEpisodeTermTimeoutRobotGNM*100}}\%, \num[round-mode=places, round-precision=0]{\fpeval{\SEpisodeTermTimeoutRobotNoMaD*100}}\%, and \num[round-mode=places, round-precision=0]{\fpeval{\SEpisodeTermTimeoutRobotNavDP*100}}\%, indicating that these policies fail to reach the goal, getting stuck on or colliding with obstacles. ViNT, by contrast, attains \num[round-mode=places, round-precision=0]{\fpeval{\SEpisodeTermSLAViNT*100}}\% SLA compliance, but its \num[round-mode=places, round-precision=0]{\fpeval{\SEpisodeTermPhysicalAssistanceViNT*100}}\% physical-assistance rate indicates frequent collisions requiring operator intervention. GNM, ViNT, and NoMaD condition on topological sub-goal images sampled along the demonstrated path. When the robot deviates from this path to avoid a crossing pedestrian or to make a sharp turn, the current observation falls out of distribution and image-based localization fails, leaving the policy unable to recover. NavDP, separately, conditions on a single image goal plus a point goal; this conditioning provides no intermediate guidance between the current pose and the destination and therefore fails to scale to long-horizon routes.

\begin{table}[t]
    \centering
    \small
    \caption{\textbf{Simulation Evaluation Results} from \SEvalEpisode\ evaluation episodes on urban sidewalk scenarios. Full details in Appendix~\ref{sec:simulation_evaluation}.}
    \label{tab:simulation-eval}
    \resizebox{\textwidth}{!}{%
        \begin{tabular}{@{}lccccccc@{}}
            \toprule
                                              & \multicolumn{2}{c}{\textbf{Rule-Based}}                             & \multicolumn{5}{c}{\textbf{Learning-Based}}                                                                                                                                                                                                                                                                                                                                                                                         \\
            \cmidrule(lr){2-3} \cmidrule(lr){4-8}
            \textbf{Metric}                   & \textbf{Nav2 w/ AMCL}                                               & \textbf{Nav2 w/ GPS}                                               & \textbf{GNM}                                                       & \textbf{ViNT}                                                       & \textbf{NoMaD}                                                       & \textbf{NavDP}                                                       & \textbf{CANVAS}                                                       \\
            \midrule
            \multicolumn{8}{l}{\emph{Episode Outcomes (/run)}}                                                                                                                                                                                                                                                                                                                                                                                                                                                                                            \\
            \textbf{SLA Compliance}           & \textbf{\SEpisodeTermSLAAMCL}                                       & \textbf{\SEpisodeTermSLAGPS}                                       & \textbf{\SEpisodeTermSLAGNM}                                       & \textbf{\SEpisodeTermSLAViNT}                                       & \textbf{\SEpisodeTermSLANoMaD}                                       & \textbf{\SEpisodeTermSLANavDP}                                       & \textbf{\SEpisodeTermSLACANVAS}                                       \\
            Spoiled                           & \SEpisodeTermSpoiledAMCL                                            & \SEpisodeTermSpoiledGPS                                            & \SEpisodeTermSpoiledGNM                                            & \SEpisodeTermSpoiledViNT                                            & \SEpisodeTermSpoiledNoMaD                                            & \SEpisodeTermSpoiledNavDP                                            & \SEpisodeTermSpoiledCANVAS                                            \\
            Timeout                           & \SEpisodeTermTimeoutRobotAMCL                                       & \SEpisodeTermTimeoutRobotGPS                                       & \SEpisodeTermTimeoutRobotGNM                                       & \SEpisodeTermTimeoutRobotViNT                                       & \SEpisodeTermTimeoutRobotNoMaD                                       & \SEpisodeTermTimeoutRobotNavDP                                       & \SEpisodeTermTimeoutRobotCANVAS                                       \\
            Physical Assistance               & \SEpisodeTermPhysicalAssistanceAMCL                                 & \SEpisodeTermPhysicalAssistanceGPS                                 & \SEpisodeTermPhysicalAssistanceGNM                                 & \SEpisodeTermPhysicalAssistanceViNT                                 & \SEpisodeTermPhysicalAssistanceNoMaD                                 & \SEpisodeTermPhysicalAssistanceNavDP                                 & \SEpisodeTermPhysicalAssistanceCANVAS                                 \\
            \midrule
            \multicolumn{8}{l}{\emph{Physical Metrics}}                                                                                                                                                                                                                                                                                                                                                                                                                                                                                                   \\
            Avg Velocity (m/s)                & \num[round-mode=places, round-precision=4]\SAvgVelocityAMCL         & \num[round-mode=places, round-precision=4]\SAvgVelocityGPS         & \num[round-mode=places, round-precision=4]\SAvgVelocityGNM         & \num[round-mode=places, round-precision=4]\SAvgVelocityViNT         & \num[round-mode=places, round-precision=4]\SAvgVelocityNoMaD         & \num[round-mode=places, round-precision=4]\SAvgVelocityNavDP         & \num[round-mode=places, round-precision=4]\SAvgVelocityCANVAS         \\
            Avg Mech.\ Power (kW)             & \num[round-mode=places, round-precision=4]\SAvgMechanicalPowerAMCL  & \num[round-mode=places, round-precision=4]\SAvgMechanicalPowerGPS  & \num[round-mode=places, round-precision=4]\SAvgMechanicalPowerGNM  & \num[round-mode=places, round-precision=4]\SAvgMechanicalPowerViNT  & \num[round-mode=places, round-precision=4]\SAvgMechanicalPowerNoMaD  & \num[round-mode=places, round-precision=4]\SAvgMechanicalPowerNavDP  & \num[round-mode=places, round-precision=4]\SAvgMechanicalPowerCANVAS  \\
            Avg Collision Impulse (N$\cdot$s) & \num[round-mode=places, round-precision=2]\SCollisionImpulseAMCL    & \num[round-mode=places, round-precision=2]\SCollisionImpulseGPS    & \num[round-mode=places, round-precision=2]\SCollisionImpulseGNM    & \num[round-mode=places, round-precision=2]\SCollisionImpulseViNT    & \num[round-mode=places, round-precision=2]\SCollisionImpulseNoMaD    & \num[round-mode=places, round-precision=2]\SCollisionImpulseNavDP    & \num[round-mode=places, round-precision=2]\SCollisionImpulseCANVAS    \\
            Avg Collision $\Delta v$ (m/s)    & \num[round-mode=places, round-precision=4]\SCollisionDeltaVAMCL     & \num[round-mode=places, round-precision=4]\SCollisionDeltaVGPS     & \num[round-mode=places, round-precision=4]\SCollisionDeltaVGNM     & \num[round-mode=places, round-precision=4]\SCollisionDeltaVViNT     & \num[round-mode=places, round-precision=4]\SCollisionDeltaVNoMaD     & \num[round-mode=places, round-precision=4]\SCollisionDeltaVNavDP     & \num[round-mode=places, round-precision=4]\SCollisionDeltaVCANVAS     \\
            \midrule
            \multicolumn{8}{l}{\emph{Safety Events (/run)}}                                                                                                                                                                                                                                                                                                                                                                                                                                                                                               \\
            Pedestrian Injury Cost (\$)       & \SPedInjuryAMCL                                                     & \SPedInjuryGPS                                                     & \SPedInjuryGNM                                                     & \SPedInjuryViNT                                                     & \SPedInjuryNoMaD                                                     & \SPedInjuryNavDP                                                     & \SPedInjuryCANVAS                                                     \\
            Mailbox Contact                   & \SPropMailBoxContactAMCL                                            & \SPropMailBoxContactGPS                                            & \SPropMailBoxContactGNM                                            & \SPropMailBoxContactViNT                                            & \SPropMailBoxContactNoMaD                                            & \SPropMailBoxContactNavDP                                            & \SPropMailBoxContactCANVAS                                            \\
            Trash Bin Contact                 & \SPropTrashBinContactAMCL                                           & \SPropTrashBinContactGPS                                           & \SPropTrashBinContactGNM                                           & \SPropTrashBinContactViNT                                           & \SPropTrashBinContactNoMaD                                           & \SPropTrashBinContactNavDP                                           & \SPropTrashBinContactCANVAS                                           \\
            Building Glass Contact            & \SPropBuildingGlassContactAMCL                                      & \SPropBuildingGlassContactGPS                                      & \SPropBuildingGlassContactGNM                                      & \SPropBuildingGlassContactViNT                                      & \SPropBuildingGlassContactNoMaD                                      & \SPropBuildingGlassContactNavDP                                      & \SPropBuildingGlassContactCANVAS                                      \\
            Bollard Contact                   & \SPropBollardContactAMCL                                            & \SPropBollardContactGPS                                            & \SPropBollardContactGNM                                            & \SPropBollardContactViNT                                            & \SPropBollardContactNoMaD                                            & \SPropBollardContactNavDP                                            & \SPropBollardContactCANVAS                                            \\
            \midrule
            \multicolumn{8}{l}{\emph{Performance}}                                                                                                                                                                                                                                                                                                                                                                                                                                                                                                        \\
            Avg Runtime (hr/run)              & \num[round-mode=places, round-precision=4]\SAvgRunTimeAMCL          & \num[round-mode=places, round-precision=4]\SAvgRunTimeGPS          & \num[round-mode=places, round-precision=4]\SAvgRunTimeGNM          & \num[round-mode=places, round-precision=4]\SAvgRunTimeViNT          & \num[round-mode=places, round-precision=4]\SAvgRunTimeNoMaD          & \num[round-mode=places, round-precision=4]\SAvgRunTimeNavDP          & \num[round-mode=places, round-precision=4]\SAvgRunTimeCANVAS          \\
            Avg Distance (km/run)             & \num[round-mode=places, round-precision=4]\SAvgDeliveryDistanceAMCL & \num[round-mode=places, round-precision=4]\SAvgDeliveryDistanceGPS & \num[round-mode=places, round-precision=4]\SAvgDeliveryDistanceGNM & \num[round-mode=places, round-precision=4]\SAvgDeliveryDistanceViNT & \num[round-mode=places, round-precision=4]\SAvgDeliveryDistanceNoMaD & \num[round-mode=places, round-precision=4]\SAvgDeliveryDistanceNavDP & \num[round-mode=places, round-precision=4]\SAvgDeliveryDistanceCANVAS \\
            \bottomrule
        \end{tabular}%
    }
\end{table}

\begin{table}[t]
    \centering
    \small
    \caption{\textbf{Overall Economic Performance} on urban sidewalk scenarios. Full details in Appendix~\ref{sec:cost_formula}.}
    \label{tab:overall}
    \resizebox{\textwidth}{!}{%
        \begin{tabular}{@{}lccccccc@{}}
            \toprule
                                                  & \multicolumn{2}{c}{\textbf{Rule-Based}}                               & \multicolumn{5}{c}{\textbf{Learning-Based}}                                                                                                                                                                                                                                                                                                                                                                                                     \\
            \cmidrule(lr){2-3} \cmidrule(lr){4-8}
            \textbf{Metric}                       & \textbf{Nav2 w/ AMCL}                                                 & \textbf{Nav2 w/ GPS}                                                 & \textbf{GNM}                                                         & \textbf{ViNT}                                                         & \textbf{NoMaD}                                                         & \textbf{NavDP}                                                         & \textbf{CANVAS}                                                         \\
            \midrule
            \multicolumn{8}{l}{\emph{CAPEX (\$)}}                                                                                                                                                                                                                                                                                                                                                                                                                                                                                                                           \\
            Hardware                              & \CHardware                                                            & \CHardwareGPS                                                        & \CHardwareGNM                                                        & \CHardwareViNT                                                        & \CHardwareNoMaD                                                        & \CHardwareNavDP                                                        & \CHardwareCANVAS                                                        \\
            Data Collection                       & 0                                                                     & 0                                                                    & \CDataCollection                                                     & \CDataCollection                                                      & \CDataCollection                                                       & \CDataCollection                                                       & \CDataCollection                                                        \\
            \hdashline[0.5pt/2pt]
            \noalign{\vskip 2pt}
            \textbf{Total CAPEX Cost}             & \CCAPEXAMCL                                                           & \CCAPEXGPS                                                           & \CCAPEXGNM                                                           & \CCAPEXViNT                                                           & \CCAPEXNoMaD                                                           & \CCAPEXNavDP                                                           & \CCAPEXCANVAS                                                           \\
            \midrule
            \multicolumn{8}{l}{\emph{OPEX (\$/run)}}                                                                                                                                                                                                                                                                                                                                                                                                                                                                                                                        \\
            Electricity                           & \num[round-mode=places, round-precision=4]{\CElectricityRunAMCL}      & \num[round-mode=places, round-precision=4]{\CElectricityRunGPS}      & \num[round-mode=places, round-precision=4]{\CElectricityRunGNM}      & \num[round-mode=places, round-precision=4]{\CElectricityRunViNT}      & \num[round-mode=places, round-precision=4]{\CElectricityRunNoMaD}      & \num[round-mode=places, round-precision=4]{\CElectricityRunNavDP}      & \num[round-mode=places, round-precision=4]{\CElectricityRunCANVAS}      \\
            Service Compensation                  & \num[round-mode=places, round-precision=4]{\CServiceCompRunAMCL}      & \num[round-mode=places, round-precision=4]{\CServiceCompRunGPS}      & \num[round-mode=places, round-precision=4]{\CServiceCompRunGNM}      & \num[round-mode=places, round-precision=4]{\CServiceCompRunViNT}      & \num[round-mode=places, round-precision=4]{\CServiceCompRunNoMaD}      & \num[round-mode=places, round-precision=4]{\CServiceCompRunNavDP}      & \num[round-mode=places, round-precision=4]{\CServiceCompRunCANVAS}      \\
            Pedestrian Safety                     & \num[round-mode=places, round-precision=4]{\CPedestrianRunAMCL}       & \num[round-mode=places, round-precision=4]{\CPedestrianRunGPS}       & \num[round-mode=places, round-precision=4]{\CPedestrianRunGNM}       & \num[round-mode=places, round-precision=4]{\CPedestrianRunViNT}       & \num[round-mode=places, round-precision=4]{\CPedestrianRunNoMaD}       & \num[round-mode=places, round-precision=4]{\CPedestrianRunNavDP}       & \num[round-mode=places, round-precision=4]{\CPedestrianRunCANVAS}       \\
            Property Damage                       & \num[round-mode=places, round-precision=4]{\CPropertyRunAMCL}         & \num[round-mode=places, round-precision=4]{\CPropertyRunGPS}         & \num[round-mode=places, round-precision=4]{\CPropertyRunGNM}         & \num[round-mode=places, round-precision=4]{\CPropertyRunViNT}         & \num[round-mode=places, round-precision=4]{\CPropertyRunNoMaD}         & \num[round-mode=places, round-precision=4]{\CPropertyRunNavDP}         & \num[round-mode=places, round-precision=4]{\CPropertyRunCANVAS}         \\
            Repair                                & \num[round-mode=places, round-precision=4]{\CRepairRunAMCL}           & \num[round-mode=places, round-precision=4]{\CRepairRunGPS}           & \num[round-mode=places, round-precision=4]{\CRepairRunGNM}           & \num[round-mode=places, round-precision=4]{\CRepairRunViNT}           & \num[round-mode=places, round-precision=4]{\CRepairRunNoMaD}           & \num[round-mode=places, round-precision=4]{\CRepairRunNavDP}           & \num[round-mode=places, round-precision=4]{\CRepairRunCANVAS}           \\
            \hdashline[0.5pt/2pt]
            \noalign{\vskip 2pt}
            \textbf{Total OPEX Cost}              & \textbf{\num[round-mode=places, round-precision=4]{\COPEXRunAMCL}}    & \textbf{\num[round-mode=places, round-precision=4]{\COPEXRunGPS}}    & \textbf{\num[round-mode=places, round-precision=4]{\COPEXRunGNM}}    & \textbf{\num[round-mode=places, round-precision=4]{\COPEXRunViNT}}    & \textbf{\num[round-mode=places, round-precision=4]{\COPEXRunNoMaD}}    & \textbf{\num[round-mode=places, round-precision=4]{\COPEXRunNavDP}}    & \textbf{\num[round-mode=places, round-precision=4]{\COPEXRunCANVAS}}    \\
            \midrule
            \multicolumn{8}{l}{\emph{Profitability}}                                                                                                                                                                                                                                                                                                                                                                                                                                                                                                                        \\
            \textbf{Revenue (\$/run)}             & \textbf{\num[round-mode=places, round-precision=4]{\CRevenueRunAMCL}} & \textbf{\num[round-mode=places, round-precision=4]{\CRevenueRunGPS}} & \textbf{\num[round-mode=places, round-precision=4]{\CRevenueRunGNM}} & \textbf{\num[round-mode=places, round-precision=4]{\CRevenueRunViNT}} & \textbf{\num[round-mode=places, round-precision=4]{\CRevenueRunNoMaD}} & \textbf{\num[round-mode=places, round-precision=4]{\CRevenueRunNavDP}} & \textbf{\num[round-mode=places, round-precision=4]{\CRevenueRunCANVAS}} \\
            \textbf{Contribution Margin (\$/run)} & \textbf{\num[round-mode=places, round-precision=2]{\CMarginAMCL}}     & \textbf{\num[round-mode=places, round-precision=2]{\CMarginGPS}}     & \textbf{\num[round-mode=places, round-precision=2]{\CMarginGNM}}     & \textbf{\num[round-mode=places, round-precision=2]{\CMarginViNT}}     & \textbf{\num[round-mode=places, round-precision=2]{\CMarginNoMaD}}     & \textbf{\num[round-mode=places, round-precision=2]{\CMarginNavDP}}     & \textbf{\num[round-mode=places, round-precision=2]{\CMarginCANVAS}}     \\
            \textbf{BEP (runs)}                   & \textbf{--}                                                           & \textbf{--}                                                          & \textbf{--}                                                          & \textbf{--}                                                           & \textbf{--}                                                            & \textbf{--}                                                            & \textbf{--}                                                             \\
            \bottomrule
        \end{tabular}%
    }
\end{table}

\subsection{Sim-to-Real Performance Experiment}
\label{sec:sim_to_real}

To assess whether CostNav-trained policies transfer to the real world, we deployed the same CANVAS checkpoint used in our simulation evaluation on a Segway E1 platform along an outdoor sidewalk route in an urban area, using onboard LiDAR for localization (Table~\ref{tab:sim_to_real}).
Across \REvalEpisode\ delivery scenarios, the sim-trained policy achieved \num[round-mode=places, round-precision=0]{\fpeval{\REpisodeTermSLACANVAS*100}}\% SLA compliance, similar to the \num[round-mode=places, round-precision=0]{\fpeval{\SEpisodeTermSLACANVAS*100}}\% rate observed in simulation.
Failures involved contact with vegetation or a fence; one vegetation contact also spoiled cargo.
Per-scenario raw values are reported in Appendix~\ref{sec:sim_to_real_appendix}.
The close match between simulated and real-world SLA compliance shows that CostNav's physics is adequate to reproduce benchmark performance on physical hardware, supporting the simulation benchmark as a proxy for real-world deployment: economic metrics computed in simulation reflect behavior observable on physical hardware.

\begin{table}[t]
    \centering
    \caption{\textbf{Sim-to-Real Performance Comparison} on the CANVAS policy. See Appendix~\ref{sec:sim_to_real_appendix} for the breakdown by scenario.}
    \label{tab:sim_to_real}
    \renewcommand{\arraystretch}{0.88}%
    \setlength{\tabcolsep}{4pt}%

    \begin{minipage}[t]{0.42\textwidth}
        \centering
        \footnotesize
        \subcaption{Evaluation Results}
        \label{tab:sim_to_real_eval}

        \resizebox{0.7\linewidth}{!}{%
        \begin{tabular}{@{}lcc@{}}
            \toprule
            \textbf{Metric}                       & \textbf{Sim}                          & \textbf{Real}                          \\
            \midrule
            \multicolumn{3}{l}{\emph{Episode Outcomes (/run)}}                                                                          \\
            \textbf{SLA Compliance}               & \textbf{\SEpisodeTermSLACANVAS}       & \textbf{\REpisodeTermSLACANVAS}        \\
            Spoiled                               & \SEpisodeTermSpoiledCANVAS            & \REpisodeTermSpoiledCANVAS             \\
            Timeout                               & \SEpisodeTermTimeoutRobotCANVAS       & \REpisodeTermTimeoutRobotCANVAS        \\
            Physical Assistance                   & \SEpisodeTermPhysicalAssistanceCANVAS & \REpisodeTermPhysicalAssistanceCANVAS  \\
            \midrule
            \multicolumn{3}{l}{\emph{Physical Metrics}}                                                                                 \\
            Avg Velocity (m/s)                    & \SAvgVelocityCANVAS                   & \RAvgVelocityCANVAS                    \\
            Avg Mech.\ Power (kW)                 & \SAvgMechanicalPowerCANVAS            & \RAvgMechanicalPowerCANVAS             \\
            Avg Coll.\ Impulse (N$\cdot$s)        & \SCollisionImpulseCANVAS              & \RCollisionImpulseCANVAS               \\
            Avg Coll.\ $\Delta v$ (m/s)           & \SCollisionDeltaVCANVAS               & \RCollisionDeltaVCANVAS                \\
            \midrule
            \multicolumn{3}{l}{\emph{Safety Events (/run)}}                                                                             \\
            Pedestrian Injury (\$)                & \SPedInjuryCANVAS                     & \RPedInjuryCANVAS                      \\
            Static obstacles$^{\dagger}$          & \num[round-mode=places, round-precision=2]{\fpeval{\SPropMailBoxContactCANVAS + \SPropTrashBinContactCANVAS + \SPropBuildingGlassContactCANVAS + \SPropBollardContactCANVAS}} & 0 \\
            Plant Contact                         & --                                    & \RPropPlantContactCANVAS               \\
            Fence Contact                         & --                                    & \RPropFenceContactCANVAS               \\
            \midrule
            \multicolumn{3}{l}{\emph{Performance}}                                                                                      \\
            Avg Runtime (hr/run)                  & \num[round-mode=places, round-precision=4]\SAvgRunTimeCANVAS  & \num[round-mode=places, round-precision=4]\RAvgRunTimeCANVAS  \\
            Avg Distance (km/run)                 & \num[round-mode=places, round-precision=4]\SAvgDeliveryDistanceCANVAS & \num[round-mode=places, round-precision=4]\RAvgDeliveryDistanceCANVAS \\
            \bottomrule
        \end{tabular}%
        }

        \par\vspace{1pt}\raggedright\scriptsize $^{\dagger}$Mailbox/Trash/Glass/Bollard sum.
    \end{minipage}\hfill
    \begin{minipage}[t]{0.46\textwidth}
        \centering
        \footnotesize
        \subcaption{Economic Performance}
        \label{tab:sim_to_real_overall}

        \resizebox{0.98\linewidth}{!}{%
        \begin{tabular}{@{}lcc@{}}
            \toprule
            \textbf{Metric}                       & \textbf{Sim}                                                          & \textbf{Real}                                                         \\
            \midrule
            \multicolumn{3}{l}{\emph{CAPEX (\$)}}                                                                                                                                                \\
            Hardware                              & \CHardwareCANVAS                                                      & \RHardwareCANVAS                                                      \\
            Data Collection                       & \CDataCollection                                                      & \CDataCollection                                                      \\
            \hdashline[0.5pt/2pt]
            \noalign{\vskip 2pt}
            \textbf{Total CAPEX}                  & \textbf{\CCAPEXCANVAS}                                                & \textbf{\RCAPEXCANVAS}                                                \\
            \midrule
            \multicolumn{3}{l}{\emph{OPEX (\$/run)}}                                                                                                                                             \\
            Electricity                           & \num[round-mode=places, round-precision=4]{\CElectricityRunCANVAS}    & \num[round-mode=places, round-precision=4]{\RElectricityRunCANVAS}    \\
            Service Compensation                  & \num[round-mode=places, round-precision=4]{\CServiceCompRunCANVAS}    & \num[round-mode=places, round-precision=4]{\RServiceCompRunCANVAS}    \\
            Pedestrian Safety                     & \num[round-mode=places, round-precision=4]{\CPedestrianRunCANVAS}     & \RPedestrianRunCANVAS                                                 \\
            Property Damage                       & \num[round-mode=places, round-precision=4]{\CPropertyRunCANVAS}       & \num[round-mode=places, round-precision=2]{\RPropertyRunCANVAS}       \\
            Repair                                & \num[round-mode=places, round-precision=4]{\CRepairRunCANVAS}         & \num[round-mode=places, round-precision=2]{\RRepairRunCANVAS}         \\
            \hdashline[0.5pt/2pt]
            \noalign{\vskip 2pt}
            \textbf{Total OPEX}                   & \textbf{\num[round-mode=places, round-precision=4]{\COPEXRunCANVAS}}  & \textbf{\num[round-mode=places, round-precision=2]{\ROPEXRunCANVAS}}  \\
            \midrule
            \multicolumn{3}{l}{\emph{Profitability}}                                                                                                                                             \\
            \textbf{Revenue (\$/run)}             & \textbf{\num[round-mode=places, round-precision=4]{\CRevenueRunCANVAS}} & \textbf{\num[round-mode=places, round-precision=4]{\RRevenueRunCANVAS}} \\
            \textbf{Contribution Margin (\$/run)} & \textbf{\num[round-mode=places, round-precision=2]{\CMarginCANVAS}}   & \textbf{\num[round-mode=places, round-precision=2]{\RMarginCANVAS}}   \\
            \textbf{BEP (runs)}                   & --                                                                    & --                                                                    \\
            \bottomrule
        \end{tabular}%
        }
    \end{minipage}
\end{table}

\section{Conclusion}
\label{sec:conclusion}

We introduced CostNav, an \textbf{Economic Navigation Benchmark} that evaluates physical AI agents through cost-revenue outcomes. CostNav connects simulator-measured behavior to real-world referenced costs and revenue, enabling contribution-margin and break-even analysis for autonomous delivery. Across seven rule-based and learning-based navigation baselines, CostNav shows that high task success does not imply economic viability: every evaluated method yields a negative contribution margin, even the method with the highest task success. CostNav thereby serves as a benchmark for navigation systems that optimize economic outcomes rather than conventional metrics.

\section{Limitations}
\label{sec:limitations}

\noindent\textbf{Simulation and cost transfer.}
Although CostNav uses real-world referenced cost parameters, its event counts per episode are computed from Isaac Sim and only partially validated against real-world deployment (\S\ref{sec:sim_to_real}), so methods with similar totals can differ by only a few events, and close rankings should not be over-interpreted. For pedestrian safety, we approximate the price by rescaling passenger vehicle injury costs by mass ratio, with robot-scale crash test evidence providing partial context for this transfer in Appendix~\ref{app:pedestrian_proxy}. Even setting this term to zero leaves every evaluated method with a negative margin, so our viability conclusion is robust to it.

\noindent\textbf{Economic model and failure modes.}
Our economic model captures the dominant cost dynamics of delivery robot operations, but several components remain outside its current scope. CAPEX does not yet include training compute or rule-based mapping costs, revenue does not account for demand or dynamic pricing, and episode outcomes do not yet cover failure modes such as battery depletion.

\noindent\textbf{A single controlled evaluation setup.}
We evaluate only the Segway E1 platform, popcorn cargo, an urban sidewalk map, and daytime conditions. A subset of learning baselines reach 10\% SLA or below. We used a common integration and training setup, but further method-specific optimization may improve performance. Our results should therefore be interpreted as performance under a common CostNav setup, not as each method's absolute ceiling.

\noindent These limitations can be addressed by collecting more real-world deployment data to broaden sim-to-real validation and refine the pedestrian-injury cost, and by expanding the economic model to include omitted components such as training compute and demand effects. Stronger baselines and method-specific optimization can further improve evaluated performance.


\acknowledgments{We thank Yoonseok Kang, Yumin Jung, and Jaehyun Kim for their assistance with the literature review on cost functions, simulation environment setup, and data collection, respectively. We are also grateful to all those who supported this work. This work was supported by Institute of Information \& communications Technology Planning \& Evaluation (IITP) grant funded by the Korea government (MSIT) (No.~RS-2026-25522885, Development of a World Foundation Model for Training and Deployment of Physical AI Systems).}


\bibliography{main}

\clearpage
\appendix
\section{Cost Parameter Details}
\label{sec:cost_parameters}

Detailed breakdown of the cost parameters used in the CostNav benchmark.

{
    \small
    \setlength{\tabcolsep}{3pt}
    \begin{longtable}{l p{0.17\textwidth} l r l p{0.23\textwidth}}
        \caption{Detailed Cost Parameters.} \label{tab:cost_params} \\
        \toprule
        \textbf{Category} & \textbf{Variable} & \textbf{Symbol} & \textbf{Value} & \textbf{Unit} & \textbf{Reference/Rationale} \\
        \midrule
        \endfirsthead
    
        \multicolumn{6}{c}%
        {{\bfseries \tablename\ \thetable{} -- continued from previous page}} \\
        \toprule
        \textbf{Category} & \textbf{Variable} & \textbf{Symbol} & \textbf{Value} & \textbf{Unit} & \textbf{Rationale/Source} \\
        \midrule
        \endhead
    
        \midrule
        \multicolumn{6}{r}{{Continued on next page}} \\
        \bottomrule
        \endfoot
    
        \bottomrule
        \endlastfoot
    
        CAPEX & Robot Cost & $P_{\text{Robot}}$ & \PRobot & \$/robot &
        Retail price of commercial sidewalk delivery robots~\cite{segway-price} (Figure~\ref{fig:p_robot_source}). \\

         CAPEX & LiDAR Cost & $P_{\text{LiDAR}}$ & \PLidar & \$/robot &
       Retail price of LiDAR sensors for commercial sidewalk delivery robots~\cite{segway-e1-specs} (Figure~\ref{fig:p_lidar_source}). \\

      CAPEX & GPS Cost & $P_{\text{GPS}}$ & \PGPS & \$/robot &
      Retail price of GPS/GNSS positioning hardware for commercial sidewalk delivery robots~\cite{segway-e1-specs} (Figure~\ref{fig:p_gps_source}). \\
    
        CAPEX & Useful Life of Robots & $T_{\text{RobotLife}}$ & \TRobotLife & years &
        Expected operational lifespan per SEC filings~\cite{serve-robotics-10q} (Figure~\ref{fig:t_robotlife_source}). \\
    
        CAPEX & Useful Delivery Run Count & $N_{\text{RobotLifeRun}}$ & \NRobotLifeRun & run/robot &
        Derived: $T_{\text{RobotLife}} \times 365 \times N_{\text{Delivery}}$. \\
    
        CAPEX & Data Collector Wage & $P_{\text{DataCollector}}$ & \PDataCollector & \$/hour &
        U.S. median hourly wage for data collection personnel~\cite{salary-datacollector-2025} (Figure~\ref{fig:p_datacollector_source}). \\
    
        OPEX & Market Robot Delivery Price & $P_{\text{MktRobotDeli}}$ & \PMktRobotDeli & \$/run &
        Consumer delivery fee for campus robot services~\cite{starship-usc2024} (Figure~\ref{fig:p_mktrobotdeli_source}). \\
    
        OPEX & Avg Market Food Price & $P_{\text{MktFood}}$ & \PMktFood & \$/run &
        Derived: $P_{\text{Refund}} - P_{\text{MktRobotDeli}}$. \\
    
        OPEX & Avg Refund Price & $P_{\text{Refund}}$ & \PRefund & \$/run &
        Industry average order value for delivery failures~\cite{restauranthq-delivery-stats} (Figure~\ref{fig:p_refund_source}). \\
    
        OPEX & Operator:Robot Ratio & $N_{\text{Fleet}}$ & \NFleet & robot/operator &
        Level 4 autonomy supervision ratio per SEC filings~\cite{serve-robotics-10k-2023} (Figure~\ref{fig:n_fleet_source}). \\
    
        OPEX & Delivery Runs per day & $N_{\text{Delivery}}$ & \NDelivery & runs/day &
        Reported daily throughput per robot~\cite{serve-robotics-10k-2023} (Figure~\ref{fig:n_delivery_source}). \\
    
        OPEX & Delivery Failure Rate & $C_{\text{DeliveryFailure}}$ & \CDeliveryFailure & \%/run &
        Operational failure rate per SEC filings~\cite{serve-robotics-10k-2023} (Figure~\ref{fig:c_deliveryfailure_source}). \\
    
        OPEX & Remote Assistance Rate & $C_{\text{RemoteAssistance}}$ & \CRemoteAssistance & \%/run &
        Proportion of deliveries requiring teleoperator intervention~\cite{serve-robotics-424b4-2024} (Figure~\ref{fig:c_remoteassistance_source}). \\
    
        OPEX & Physical Assistance Rate & $C_{\text{PhysicalAssistance}}$ & \CPhysicalAssistance & \%/run &
        Proportion of deliveries requiring on-site intervention~\cite{serve-robotics-424b4-2024} (Figure~\ref{fig:c_physicalassistance_source}). \\
            
        OPEX & Repair Rate & $C_{\text{Repair}}$ & \CRepair & \%/year &
        Annual maintenance cost as percentage of $P_{\text{Robot}}$ based on industry data~\cite{standardbots-maintenance,repair-roi-2020,repair-downtime-repair-costs} (Figures~\ref{fig:c_repair_1}--\ref{fig:c_repair_3}). \\
    
        OPEX & Robot Operator Wage & $P_{\text{Operator}}$ & \POperator & \$/hour &
        U.S. median hourly wage for robot operators~\cite{salary-operator-2025} (Figure~\ref{fig:p_operator_source}). \\
    
        OPEX & Electricity Rate (US) & $P_{\text{Elec}}$ & \PElec & \$/kWh &
        U.S. average retail electricity price~\cite{eia-electricity-2025} (Figure~\ref{fig:p_elec_source}). \\
    
        OPEX & Avg Pedestrian Damage (AIS0) & $P_{\text{PedDamageAIS0}}$ & \PPedDamageAISZero & \$/event &
        Economic cost per MAIS pedestrian incident~\cite{nhtsa-crash-costs-2019,aaam_ais_2015_revision} (Figures~\ref{fig:p_peddamage_1}--\ref{fig:p_peddamage_2}). \\
    
        OPEX & Avg Pedestrian Damage (AIS1) & $P_{\text{PedDamageAIS1}}$ & \PPedDamageAISOne & \$/event &
        Same as above. \\
    
        OPEX & Avg Pedestrian Damage (AIS2) & $P_{\text{PedDamageAIS2}}$ & \PPedDamageAISTwo & \$/event &
        Same as above. \\
    
        OPEX & Avg Pedestrian Damage (AIS3) & $P_{\text{PedDamageAIS3}}$ & \PPedDamageAISThree & \$/event &
        Same as above. \\

        OPEX & MAIS Injury Report Passenger Vehicle Weight & $P_{\text{MAISReportVehicleWeight}}$ &  \num[round-mode=places, round-precision=0]\PMAISReportVehicleWeight & kg  &Derived: $10,000~\text{lbs} = 4536~\text{kg}$. Passenger Vehicle Definition in MAIS Injury Report
       ~\cite{wang2022_mais0508_deltav} (Figure~\ref{fig:p_maisreportvehicleweight_source_1}). \\

        OPEX & Mailbox Damage Cost & $P_{\text{PropMailboxDamage}}$ & \PPropMailBoxDamage & \$/event & Municipal replacement cost for mail box damage~\cite{wrentham-2026_mailbox} (Figure~\ref{fig:p_propmailboxdamage_source}). \\

        OPEX & Trash Bin Damage Cost & $P_{\text{PropTrashBinDamage}}$ & \PPropTrashBinDamage & \$/event & Municipal replacement cost for trash bin damage ~\cite{cityhotspring-2026_trashbin} (Figure~\ref{fig:p_proptrashbindamage_source}). \\
    
        OPEX & Building Glass Damage Cost & $P_{\text{PropBuildingGlassDamage}}$ & \PPropBuildingGlassDamage & \$/event &
        Estimated repair cost for building glass damage~\cite{wittenberg-damage-costs-2024-2025} (Figure~\ref{fig:p_propbuildingglassdamage_source}). \\

        OPEX & Bollard Damage Cost & $P_{\text{PropBollardDamage}}$ & \PPropBollardDamage & \$/event &
        Estimated replacement cost per bollard~\cite{wearebollards-bollard-install-cost} (Figure~\ref{fig:p_propbollarddamage_source}). \\

        OPEX & Plant Damage Cost & $P_{\text{PropPlantDamage}}$ & \PPropPlantDamage & \$/event &
        Lower end of the \$5--\$20 per-bush shrub/bush trimming cost~\cite{homeguide-shrub-trimming-2026} (Figure~\ref{fig:p_propplantdamage_source}). \\

        OPEX & Fence Damage Cost & $P_{\text{PropFenceDamage}}$ & \PPropFenceDamage & \$/event &
        Repair cost per linear foot for aluminum fencing at the upper end of the \$20--\$50/LF range~\cite{homeguide-fence-repair-2026} (Figure~\ref{fig:p_propfencedamage_source}). \\

        ETC & Robot Speed & $C_{\text{RobotSpeed}}$ & \CRobotSpeed & km/h &
        Maximum operational velocity per manufacturer specifications~\cite{starship-speed} (Figure~\ref{fig:c_robotspeed_source}). \\
        
        ETC & Robot Max Speed & $C_{\text{RobotMaxSpeed}}$ & \CRobotMaxSpeed & km/h &
        Maximum nominal speed of Segway E1 Robot~\cite{segway-e1-specs} (Figure~\ref{fig:c_robotmaxspeed_source}). \\
    
        ETC & Rolling Resistance Coefficient & $C_{\text{RollingResistanceCoeff}}$ & \CRollingResistanceCoeff & unitless &
        Coefficient of rolling resistance~\cite{rolling-resistance-2015} (Figure~\ref{fig:c_rollingresistancecoeff_source}). \\
        
        ETC & Rolling Resistance Force & $C_{\text{RollingResistanceForce}}$ & \CRollingResistanceForce & N &
        Derived: $C_{\text{RobotWeight}}$  $\times$  $9.8\,\mathrm{m/s^2}$  $\times$  $C_{\text{RollingResistanceCoeff}}$~\cite{rolling-resistance-2015} (Figure~\ref{fig:c_rollingresistanceforce_source}). \\

        ETC & Robot Weight & $C_{\text{RobotWeight}}$ & \CRobotWeight & kg &
        Maximum Off-load Weight of Segway E1 Robot~\cite{segway-e1-specs} (Figure~\ref{fig:c_robotweight_source}). \\
        
        ETC & Avg Delivery Time & $T_{\text{AvgDeliveryTime}}$ & \num[round-mode=places, round-precision=4]\TAvgDeliveryTime & hr/run &
        Mean delivery duration from operational deployments~\cite{starship-finland-2024} (Figure~\ref{fig:t_avgdeliverytime_source}). \\

        ETC & Max Delivery Time & $T_{\text{MaxDeliveryTime}}$ & \TMaxDeliveryTime & hr/run &
        Maximum guaranteed delivery time~\cite{starship-commercial-rollout,agv-network-starship-faq} (Figures~\ref{fig:t_maxdeliverytime_1_source}--\ref{fig:t_maxdeliverytime_2_source}). \\
    
        ETC & Avg Delivery Distance & $T_{\text{AvgDeliveryDistance}}$ & \TAvgDeliveryDistance & km/run &
        Mean delivery distance per SEC filings~\cite{serve-robotics-10k-2023} (Figure~\ref{fig:t_avgdeliverydistance_source}). \\
    
        ETC & Max Delivery Distance & $C_{\text{MaxDeliveryDistance}}$ & \CAvgDeliveryDistance & km/run &
        Maximum operational delivery radius~\cite{starship-dimension} (Figure~\ref{fig:c_avgdeliverydistance_source}). \\

        ETC & Charge Efficiency & $\eta_{\text{charge}}$ & \EtaCharge & unitless &
        Estimated Charge Efficiency~\cite{st-gan-charger} (Figure~\ref{fig:c_energyconvert_source_1}). \\

        ETC & Battery Roundtrip Efficiency & $\eta_{\text{batteryroundtrip}}$ & \EtaBatteryRoundtrip & unitless &
        Estimated Battery Roundtrip Efficiency~\cite{battery-efficiency} (Figure~\ref{fig:c_energyconvert_source_2}). \\

        ETC & Inverter Efficiency & $\eta_{\text{inverter}}$ & \EtaInverter & unitless &
        Estimated Inverter Efficiency~\cite{ti-motor-driver} (Figure~\ref{fig:c_energyconvert_source_3}). \\

        ETC & Motor Efficiency & $\eta_{\text{motor}}$ & \EtaMotor & unitless &
        Estimated Motor Efficiency~\cite{zau-motor} (Figure~\ref{fig:c_energyconvert_source_4}). \\
    
        ETC & Electro-Mechanical Efficiency & $C_{\text{ElectroMechanicalEff}}$ &  \num[round-mode=places, round-precision=2]\CEnergyConvert & unitless &
        Derived: $\eta_{\text{batteryroundtrip}}$ $\times$ $\eta_{\text{motor}}$ $\times$ $\eta_{\text{inverter}}$$\times$ $\eta_{\text{charge}}$. \\
\end{longtable}

}

\paragraph{Pedestrian injury-cost proxy.}
\label{app:pedestrian_proxy}
For each simulated contact, the pedestrian safety term computes an expected injury cost from the measured $\Delta v$ and accumulates it across the run; the simulated contact rate itself is not calibrated against real-world deployment data.
AIS/NHTSA injury-cost tables provide the monetary reference, while robot-scale crash-test evidence bounds how far those numbers transfer to sidewalk robots.
Crash tests with mobile service robots and personal mobility devices weighing $60$ to $133$\,kg report low injury risk at $1.0$ to $1.5$\,m/s, with higher risk at larger impact speeds and for vulnerable pedestrians~\cite{paezgranados2022crashtest}.

\clearpage
\raggedbottom
\subsection{Source Data Figures}

\begin{figure}[htbp!]
    \centering
    \includegraphics[width=0.8\textwidth]{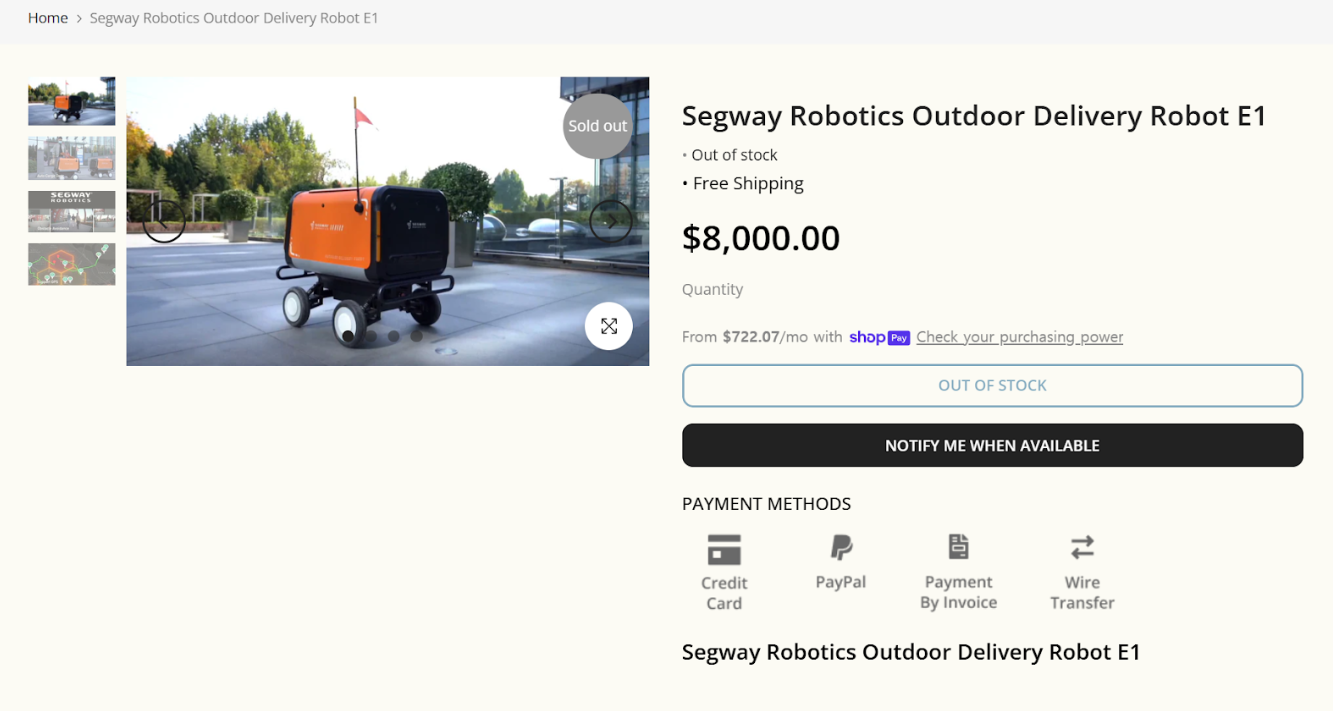}
    \caption{Source data for Robot Cost ($P_{\text{Robot}}$)}
    \label{fig:p_robot_source}
\end{figure}

\begin{figure}[htbp!]
    \centering
    \includegraphics[width=0.8\textwidth]{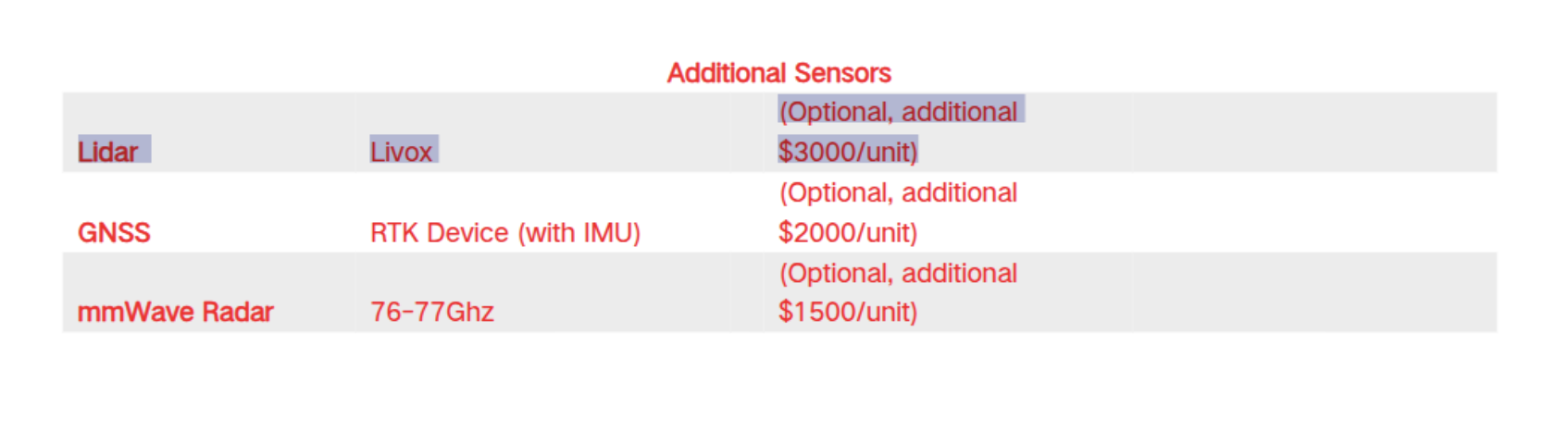}
    \caption{Source data for LiDAR Cost ($P_{\text{LiDAR}}$)}
    \label{fig:p_lidar_source}
\end{figure}

\begin{figure}[htbp!]
    \centering
    \includegraphics[width=0.8\textwidth]{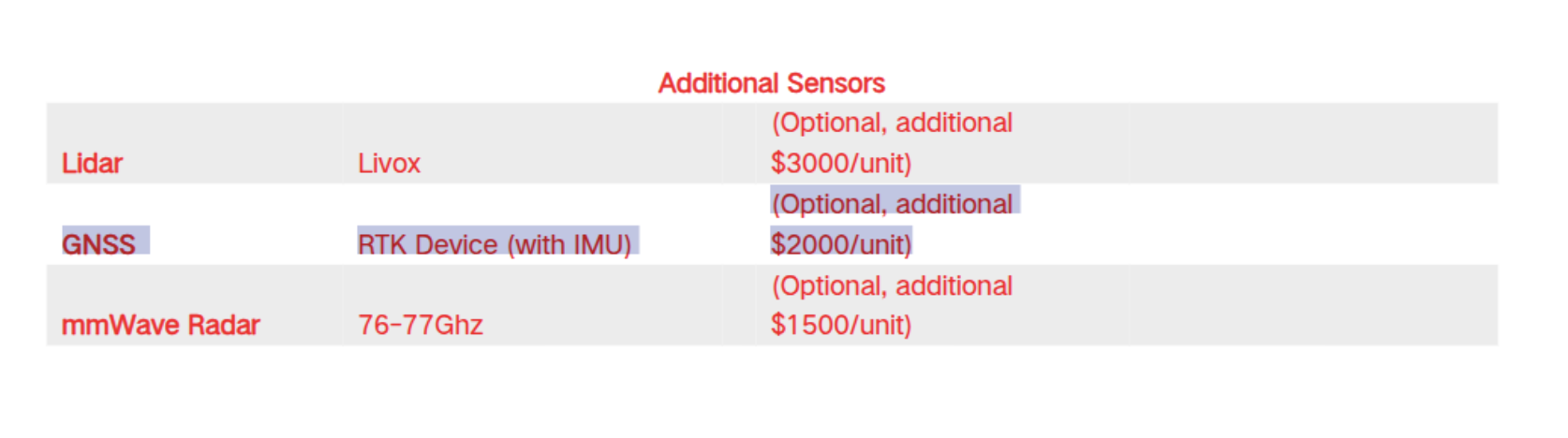}
    \caption{Source data for GPS Cost ($P_{\text{GPS}}$)}
    \label{fig:p_gps_source}
\end{figure}

\begin{figure}[htbp!]
    \centering
    \includegraphics[width=0.8\textwidth]{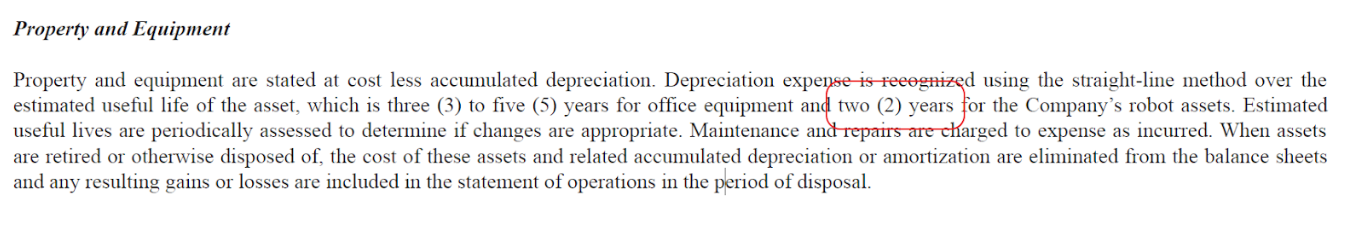}
    \caption{Source data for Useful Life of Robots ($T_{\text{RobotLife}}$)}
    \label{fig:t_robotlife_source}
\end{figure}

\begin{figure}[htbp!]
    \centering
    \includegraphics[width=0.8\textwidth]{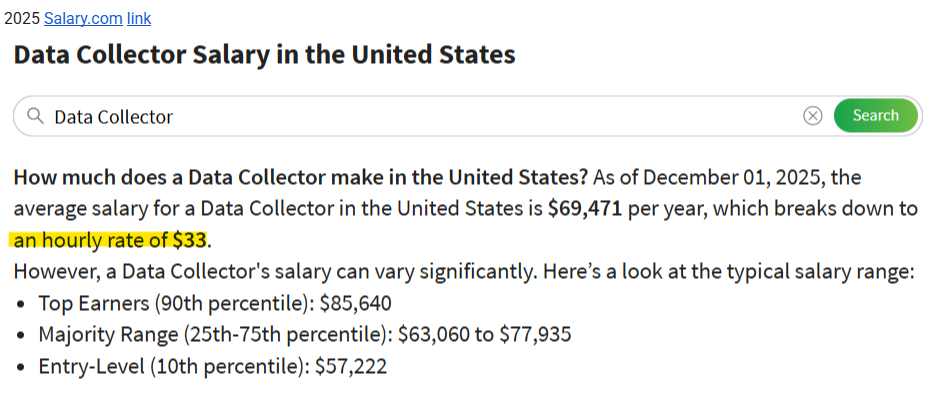}
    \caption{Source data for Data Collector Wage ($P_{\text{DataCollector}}$)}
    \label{fig:p_datacollector_source}
\end{figure}

\begin{figure}[htbp!]
    \centering
    \includegraphics[width=0.8\textwidth]{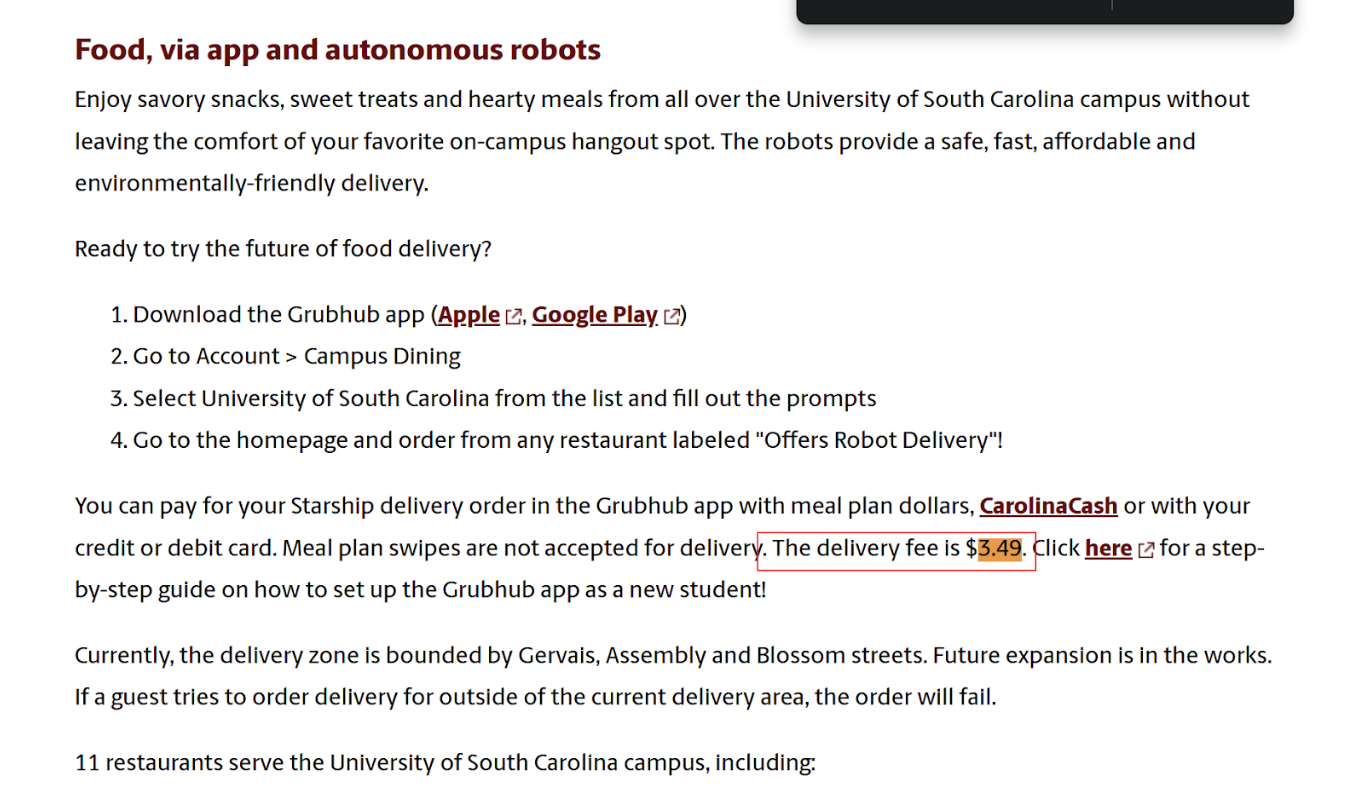}
    \caption{Source data for Price for Market Robot Delivery ($P_{\text{MktRobotDeli}}$)}
    \label{fig:p_mktrobotdeli_source}
\end{figure}

\begin{figure}[htbp!]
    \centering
    \includegraphics[width=0.8\textwidth]{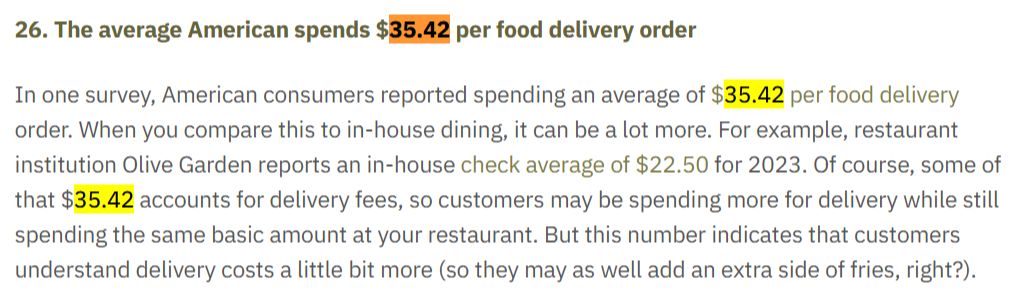}
    \caption{Source data for Avg Refund Price ($P_{\text{Refund}}$)}
    \label{fig:p_refund_source}
\end{figure}

\begin{figure}[htbp!]
    \centering
    \includegraphics[width=0.8\textwidth]{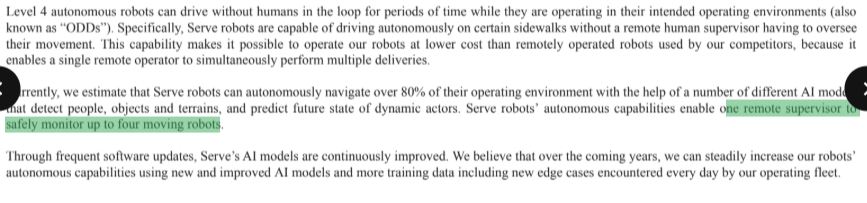}
    \caption{Source data for Operator:Robot Ratio ($N_{\text{Fleet}}$)}
    \label{fig:n_fleet_source}
\end{figure}

\begin{figure}[htbp!]
    \centering
    \includegraphics[width=0.8\textwidth]{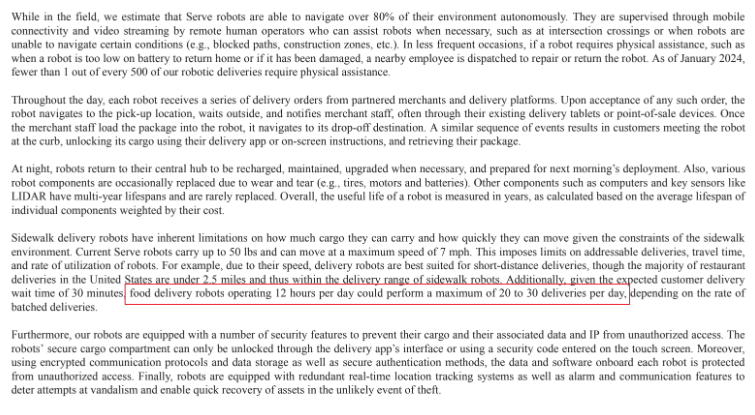}
    \caption{Source data for Delivery Runs per day ($N_{\text{Delivery}}$)}
    \label{fig:n_delivery_source}
\end{figure}

\begin{figure}[htbp!]
    \centering
    \includegraphics[width=0.8\textwidth]{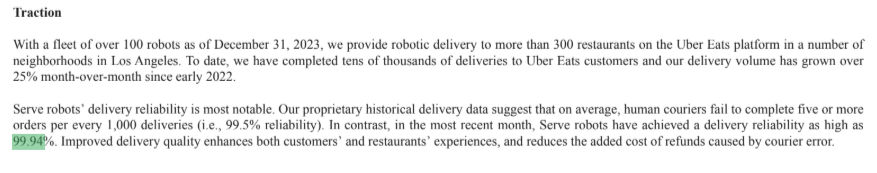}
    \caption{Source data for Delivery Failure Rate ($C_{\text{DeliveryFailure}}$)}
    \label{fig:c_deliveryfailure_source}
\end{figure}

\begin{figure}[htbp!]
    \centering
    \includegraphics[width=0.8\textwidth]{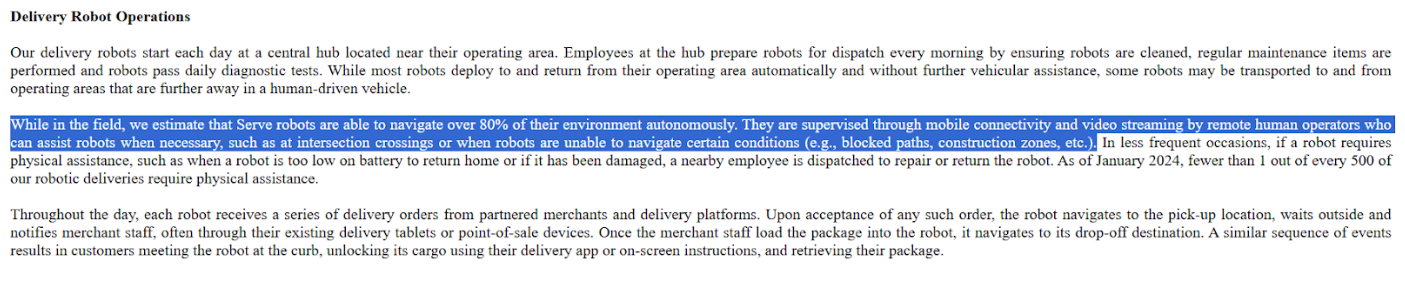}
    \caption{Source data for Remote Assistance Rate ($C_{\text{RemoteAssistance}}$)}
    \label{fig:c_remoteassistance_source}
\end{figure}

\begin{figure}[htbp!]
    \centering
    \includegraphics[width=0.8\textwidth]{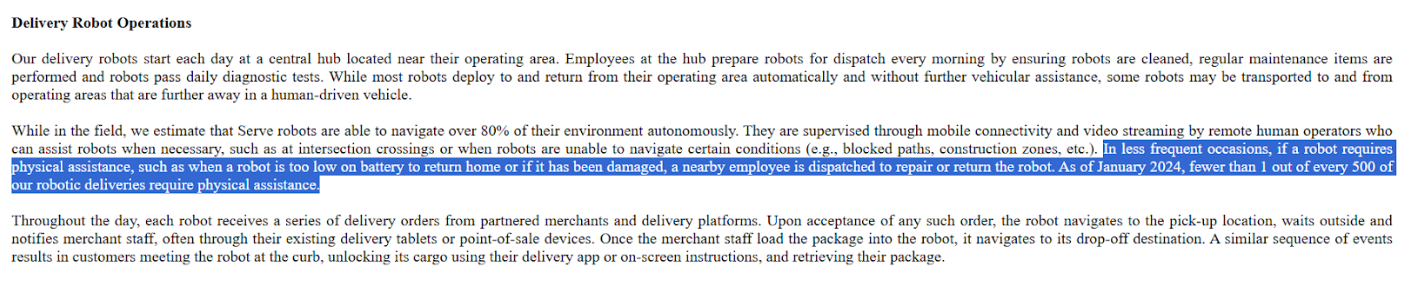}
    \caption{Source data for Physical Assistance Rate ($C_{\text{PhysicalAssistance}}$)}
    \label{fig:c_physicalassistance_source}
\end{figure}

\begin{figure}[htbp!]
    \centering
    \includegraphics[width=0.8\textwidth]{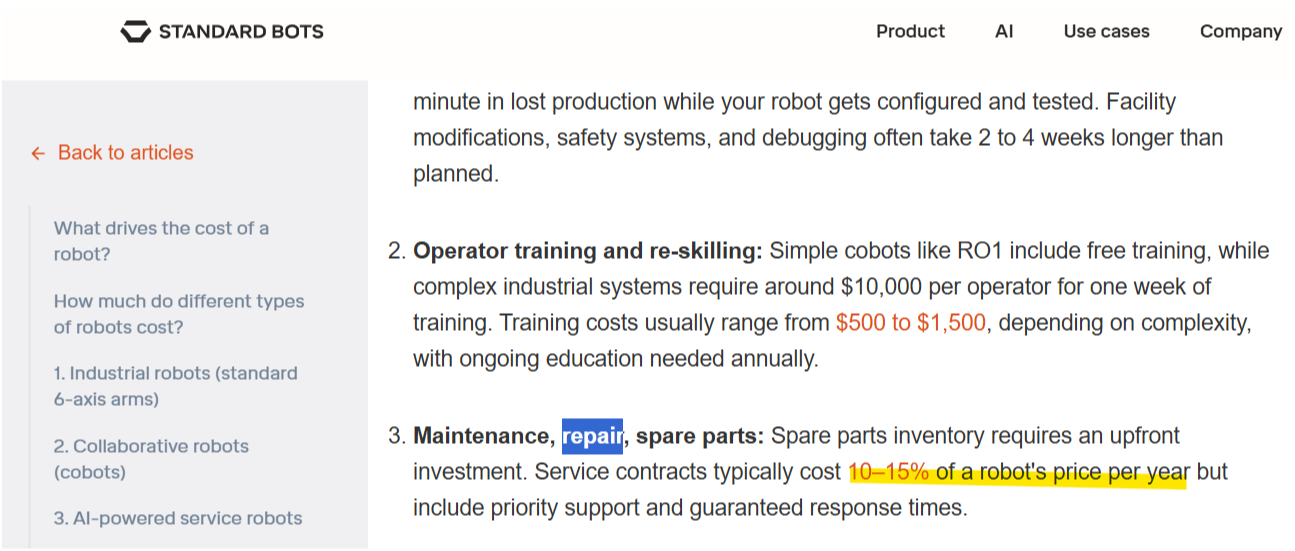}
    \caption{Source data for Repair Rate 1 (Standard Bots)}
    \label{fig:c_repair_1}
\end{figure}

\begin{figure}[htbp!]
    \centering
    \includegraphics[width=0.8\textwidth]{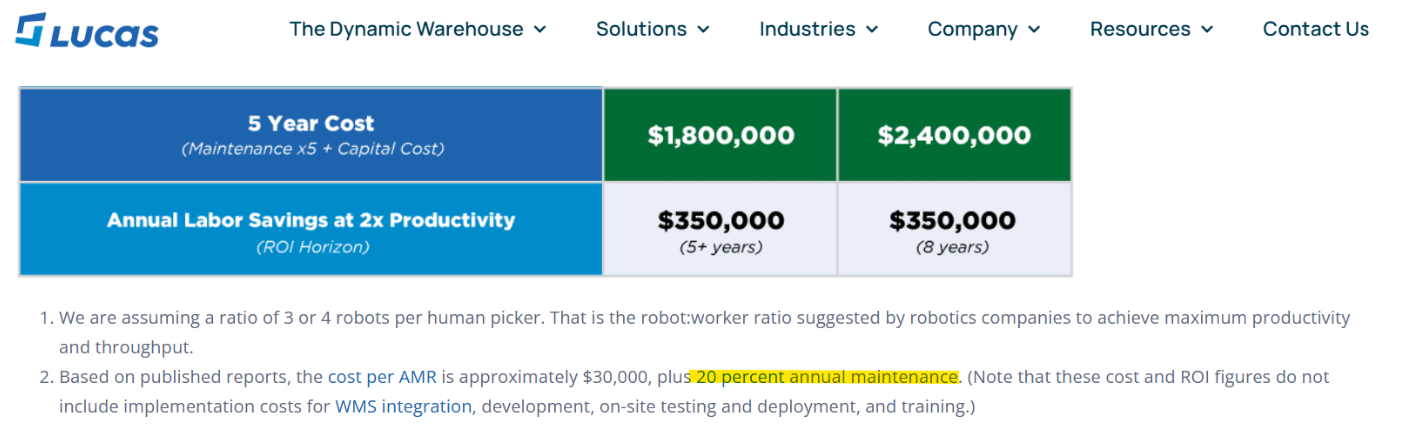}
    \caption{Source data for Repair Rate 2 (Lucas Warehouse)}
    \label{fig:c_repair_2}
\end{figure}

\begin{figure}[htbp!]
    \centering
    \includegraphics[width=0.8\textwidth]{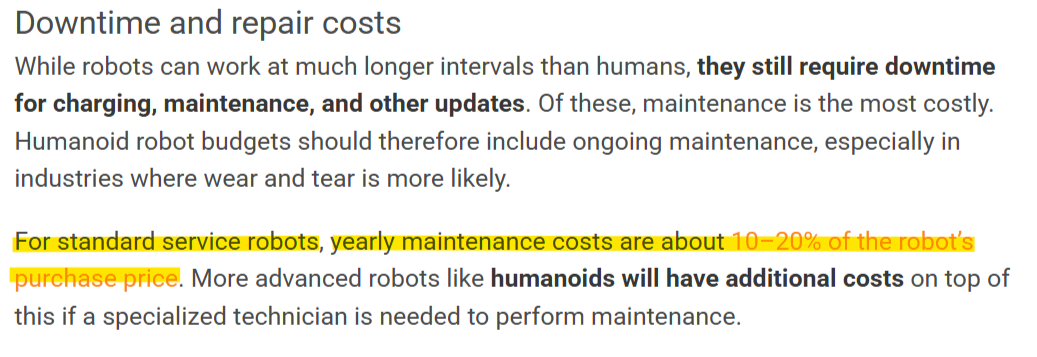}
    \caption{Source data for Repair Rate 3 (Ignus)}
    \label{fig:c_repair_3}
\end{figure}

\begin{figure}[htbp!]
    \centering
    \includegraphics[width=0.8\textwidth]{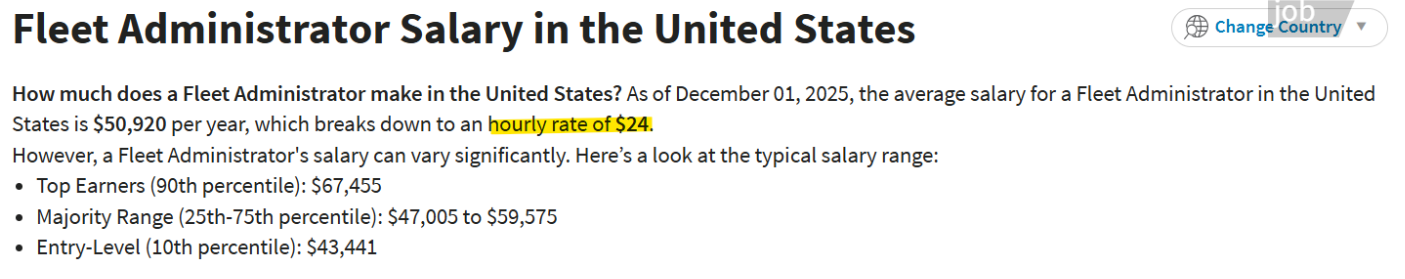}
    \caption{Source data for Robot Operator Wage ($P_{\text{Operator}}$)}
    \label{fig:p_operator_source}
\end{figure}

\begin{figure}[htbp!]
    \centering
    \includegraphics[width=0.8\textwidth]{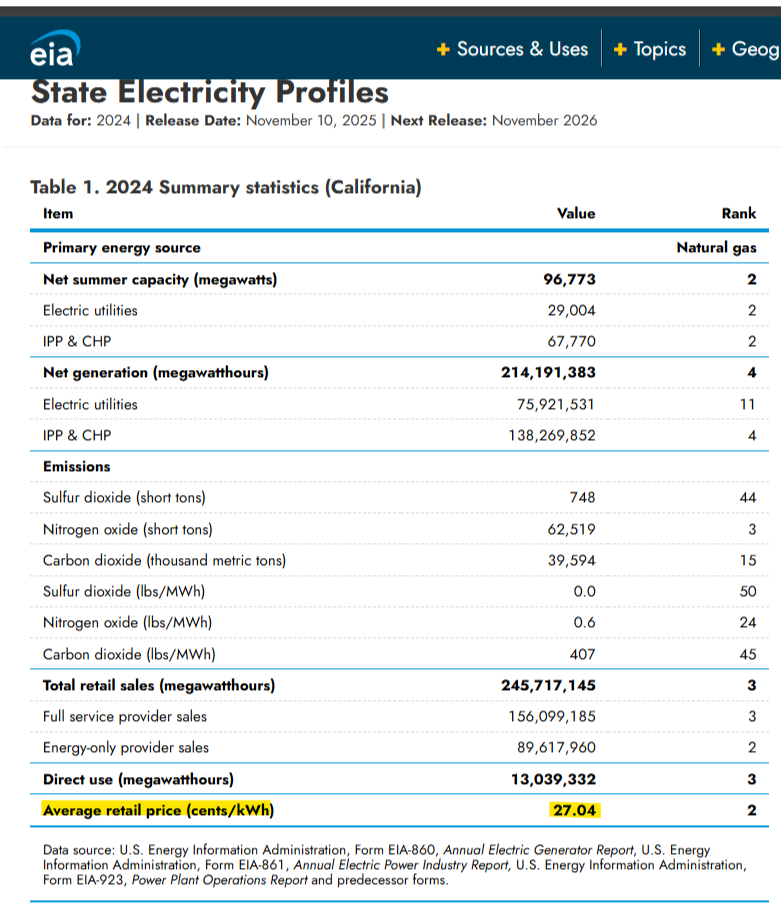}
    \caption{Source data for Electricity Rate (US) ($P_{\text{Elec}}$)}
    \label{fig:p_elec_source}
\end{figure}

\begin{figure}[htbp!]
    \centering
    \includegraphics[width=0.8\textwidth]{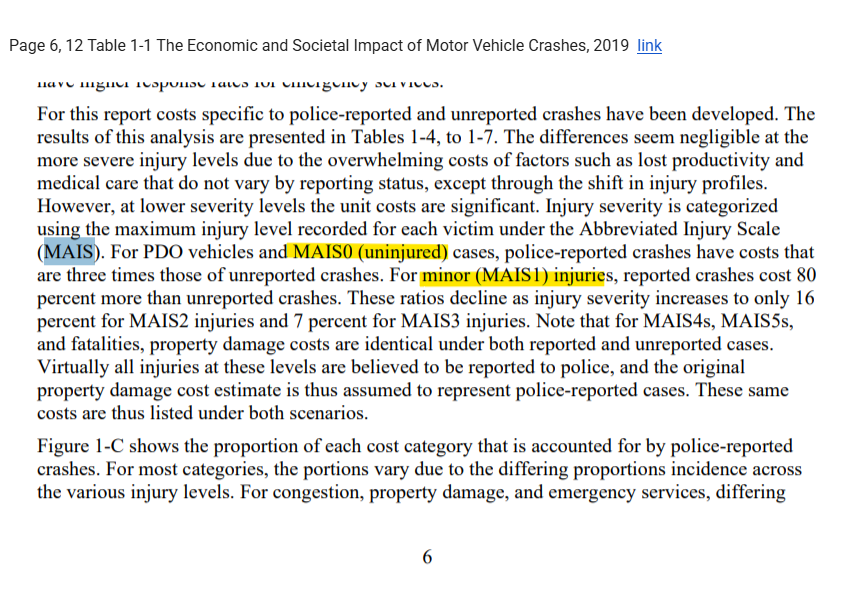}
    \caption{Source data for Avg Pedestrian Damage Cost 1 ($P_{\text{PedDamage}}$)}
    \label{fig:p_peddamage_1}
\end{figure}

\begin{figure}[htbp!]
    \centering
    \includegraphics[width=0.8\textwidth]{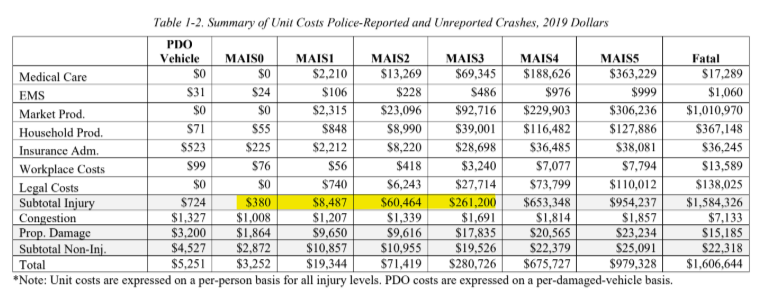}
    \caption{Source data for Avg Pedestrian Damage Cost 2 ($P_{\text{PedDamage}}$)}
    \label{fig:p_peddamage_2}
\end{figure}

\begin{figure}[htbp!]
    \centering
    \includegraphics[width=0.8\textwidth]{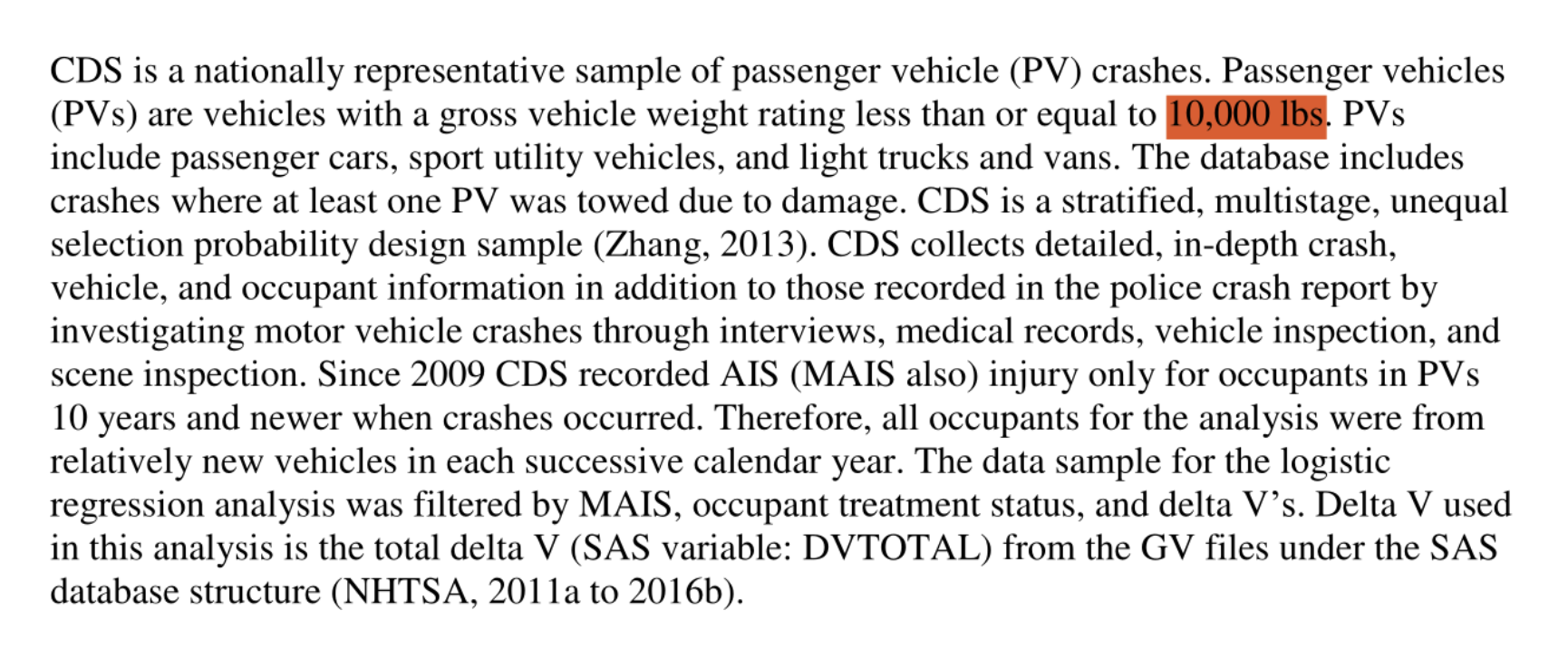}
    \caption{Source data for Vehicle Weight ($P_{\text{MAISReportVehicleWeight}}$)}
    \label{fig:p_maisreportvehicleweight_source_1}
\end{figure}

\begin{figure}[htbp!]
    \centering
    \includegraphics[width=0.8\textwidth]{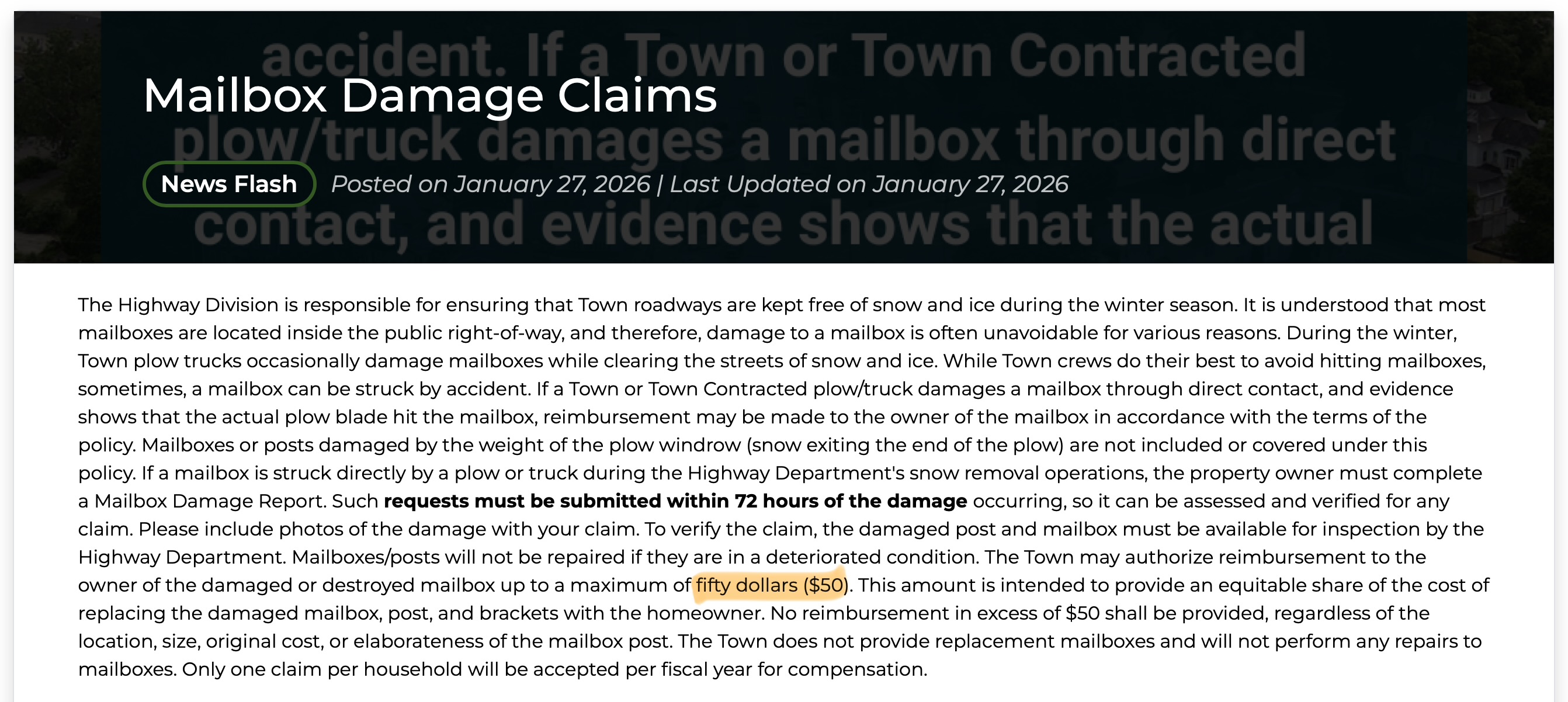}
    \caption{Source data for Mailbox Damage Cost ($P_{\text{PropMailboxDamage}}$)}
    \label{fig:p_propmailboxdamage_source}
\end{figure}

\begin{figure}[htbp!]
    \centering
    \includegraphics[width=0.8\textwidth]{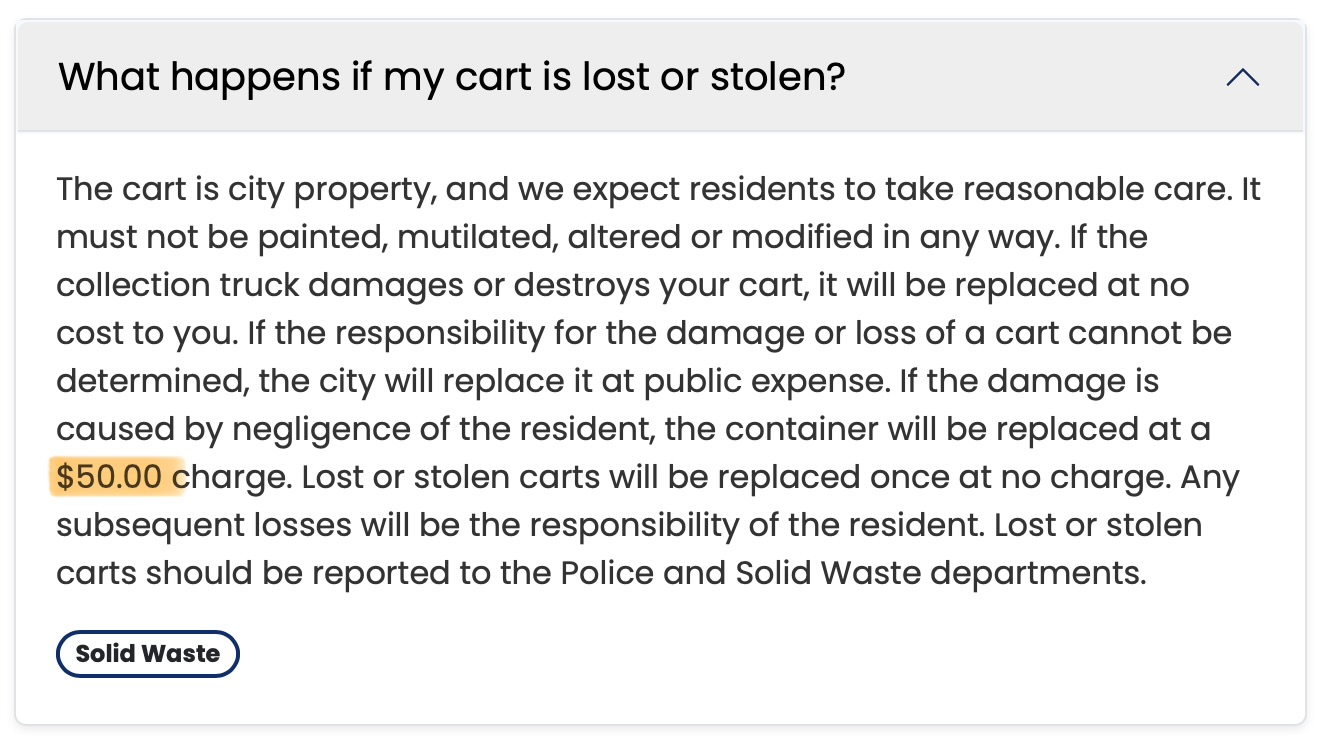}
    \caption{Source data for Trashbin Damage Cost ($P_{\text{PropTrashbinDamage}}$)}
    \label{fig:p_proptrashbindamage_source}
\end{figure}

\begin{figure}[htbp!]
    \centering
    \includegraphics[width=0.8\textwidth]{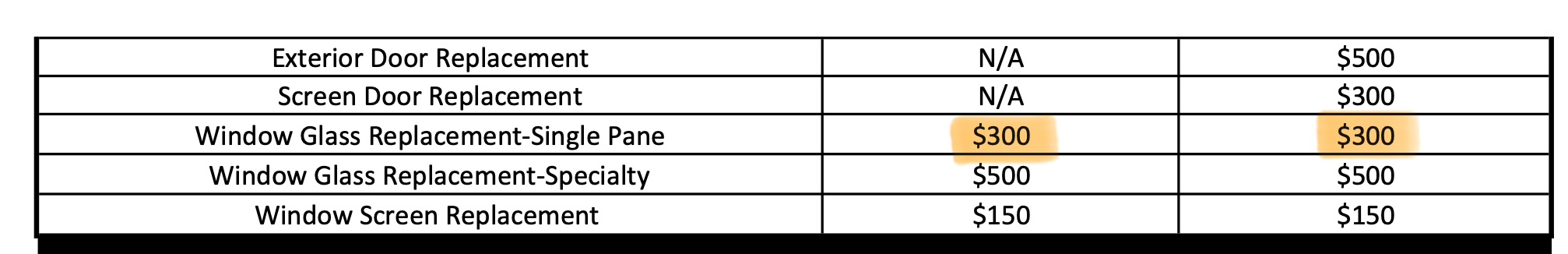}
    \caption{Source data for Building Glass Damage Cost ($P_{\text{PropBuildingGlassDamage}}$)}
    \label{fig:p_propbuildingglassdamage_source}
\end{figure}

\begin{figure}[htbp!]
    \centering
    \includegraphics[width=0.8\textwidth]{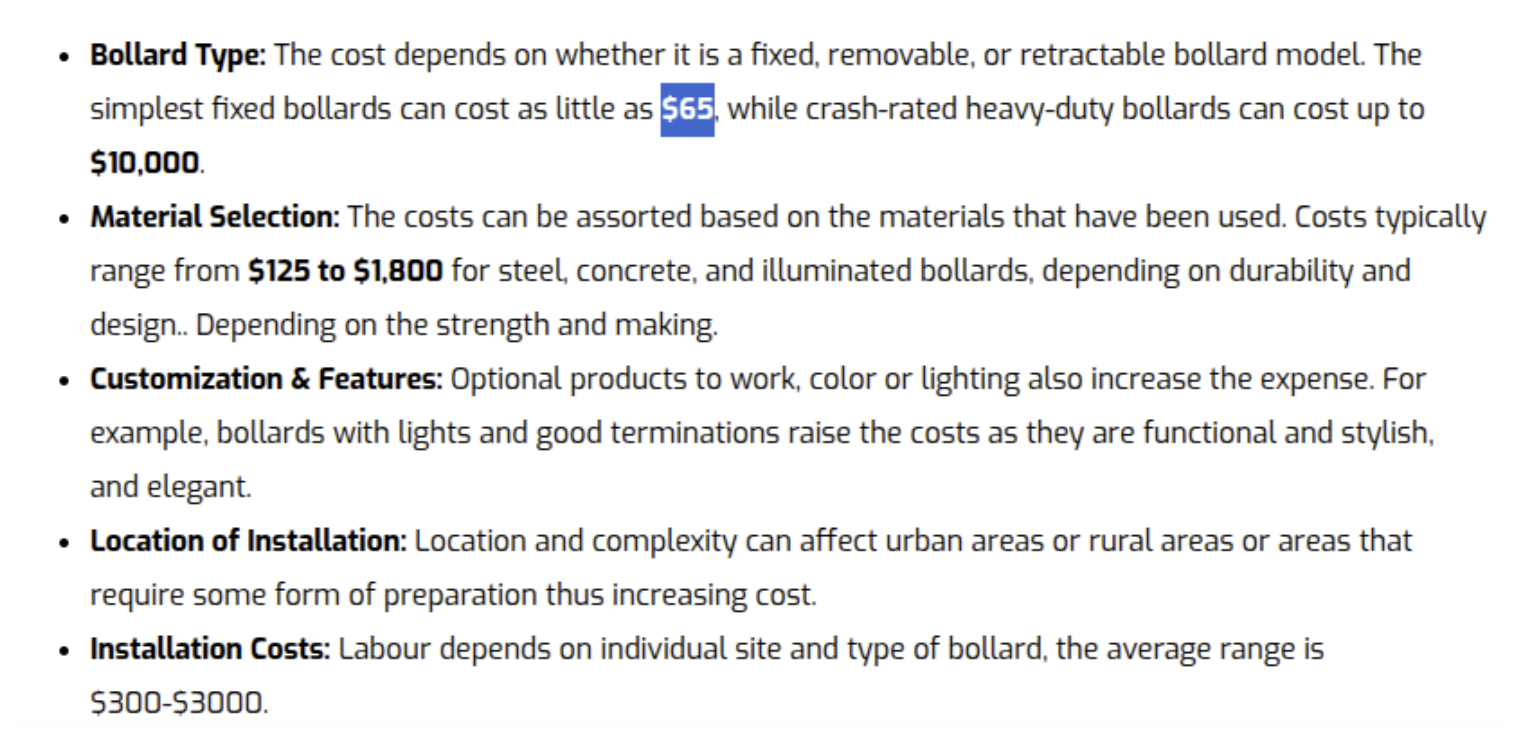}
    \caption{Source data for Bollard Damage Cost ($P_{\text{PropBollardDamage}}$)}
    \label{fig:p_propbollarddamage_source}
\end{figure}

\begin{figure}[htbp!]
    \centering
    \includegraphics[width=0.8\textwidth]{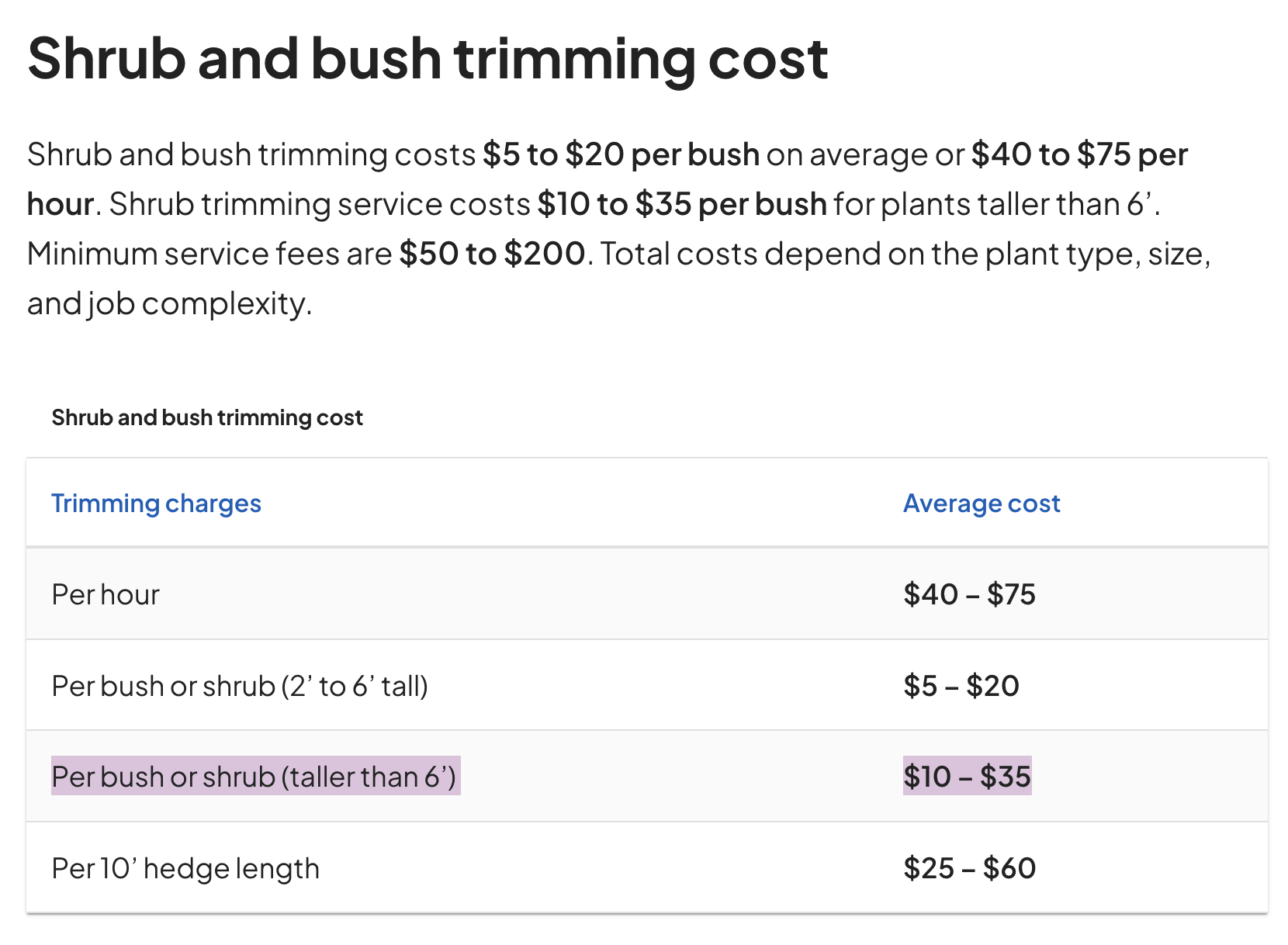}
    \caption{Source data for Plant Damage Cost ($P_{\text{PropPlantDamage}}$)}
    \label{fig:p_propplantdamage_source}
\end{figure}

\begin{figure}[htbp!]
    \centering
    \includegraphics[width=0.8\textwidth]{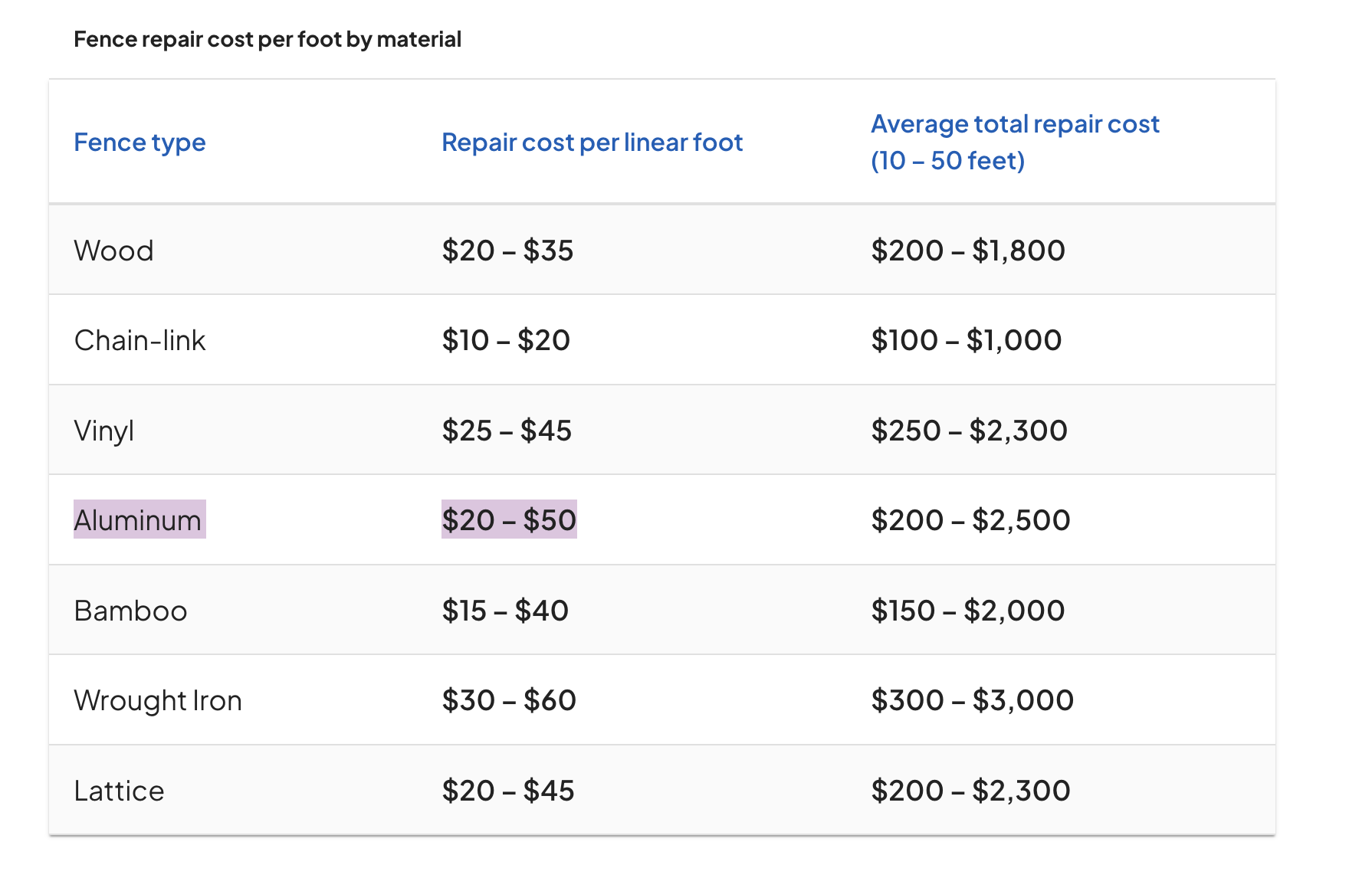}
    \caption{Source data for Fence Damage Cost ($P_{\text{PropFenceDamage}}$)}
    \label{fig:p_propfencedamage_source}
\end{figure}

\begin{figure}[htbp!]
    \centering
    \includegraphics[width=0.8\textwidth]{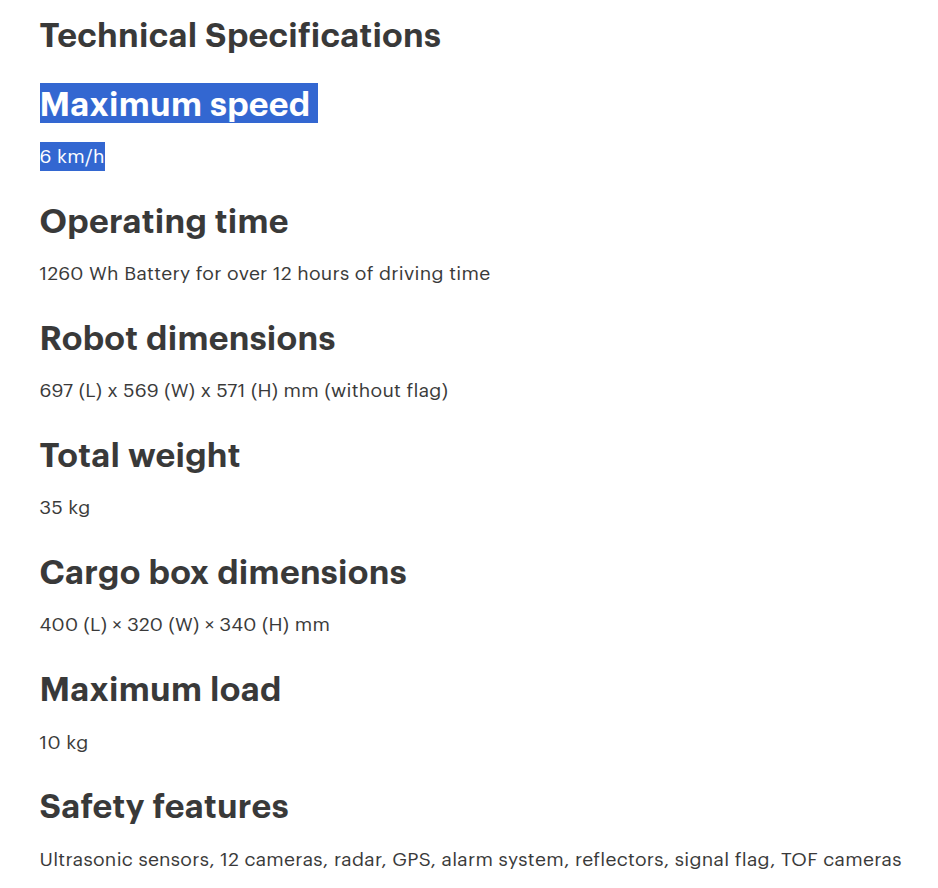}
    \caption{Source data for Robot Speed ($C_{\text{RobotSpeed}}$)}
    \label{fig:c_robotspeed_source}
\end{figure}

\begin{figure}[htbp!]
    \centering
    \includegraphics[width=0.8\textwidth]{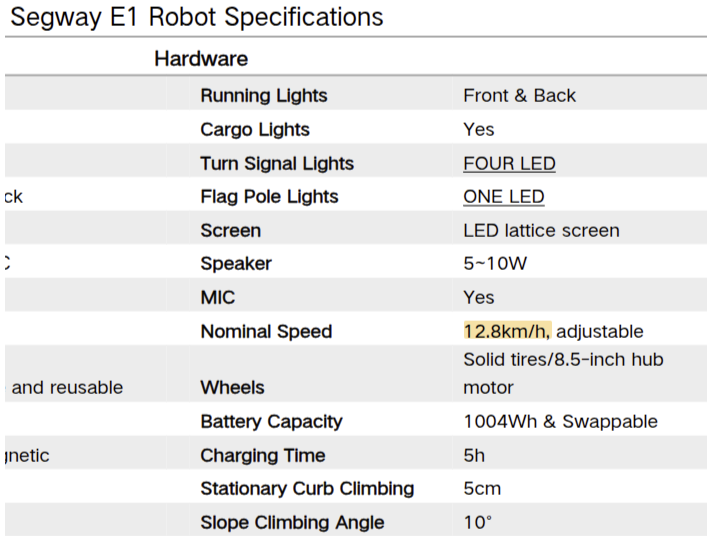}
    \caption{Source data for Robot Max Speed ($C_{\text{RobotMaxSpeed}}$)}
    \label{fig:c_robotmaxspeed_source}
\end{figure}

\begin{figure}[htbp!]
    \centering
    \includegraphics[width=0.8\textwidth]{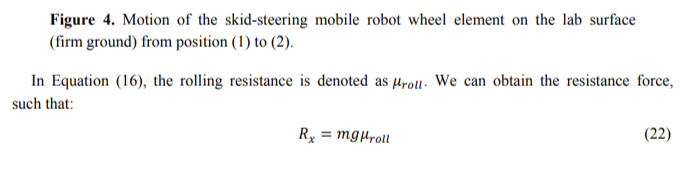}
    \caption{Source data for Rolling Resistance Force ($C_{\text{RollingResistanceForce}}$)}
    \label{fig:c_rollingresistanceforce_source}
\end{figure}

\begin{figure}[htbp!]
    \centering
    \includegraphics[width=0.8\textwidth]{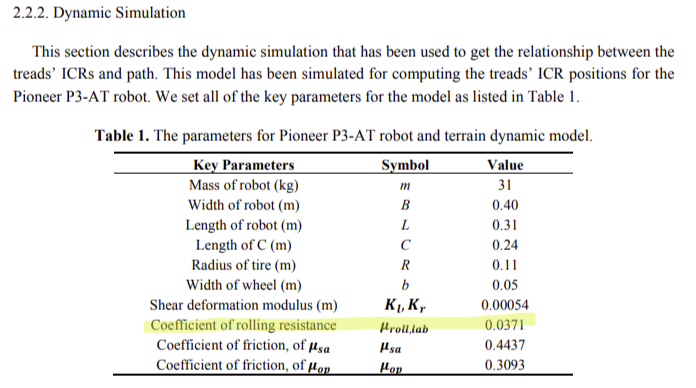}
    \caption{Source data for Rolling Resistance Coefficient ($C_{\text{RollingResistanceCoeff}}$)}
    \label{fig:c_rollingresistancecoeff_source}
\end{figure}

\begin{figure}[htbp!]
    \centering
    \includegraphics[width=0.8\textwidth]{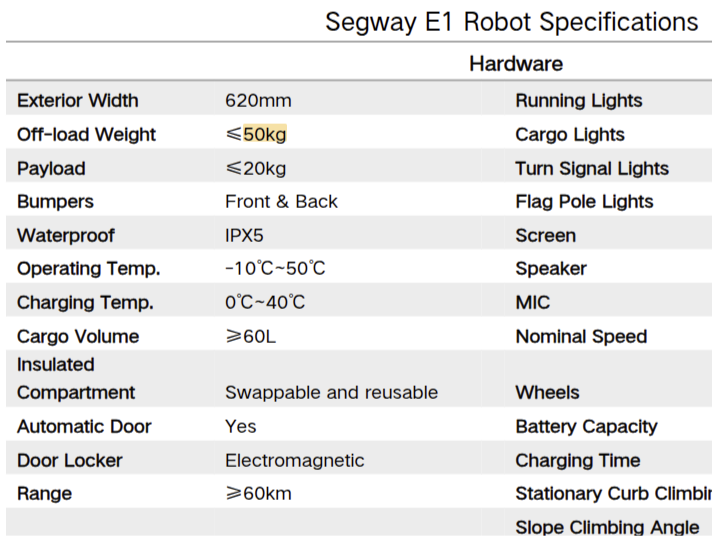}
    \caption{Source data for Robot Weight ($C_{\text{RobotWeight}}$)}
    \label{fig:c_robotweight_source}
\end{figure}

\begin{figure}[htbp!]
    \centering
    \includegraphics[width=0.8\textwidth]{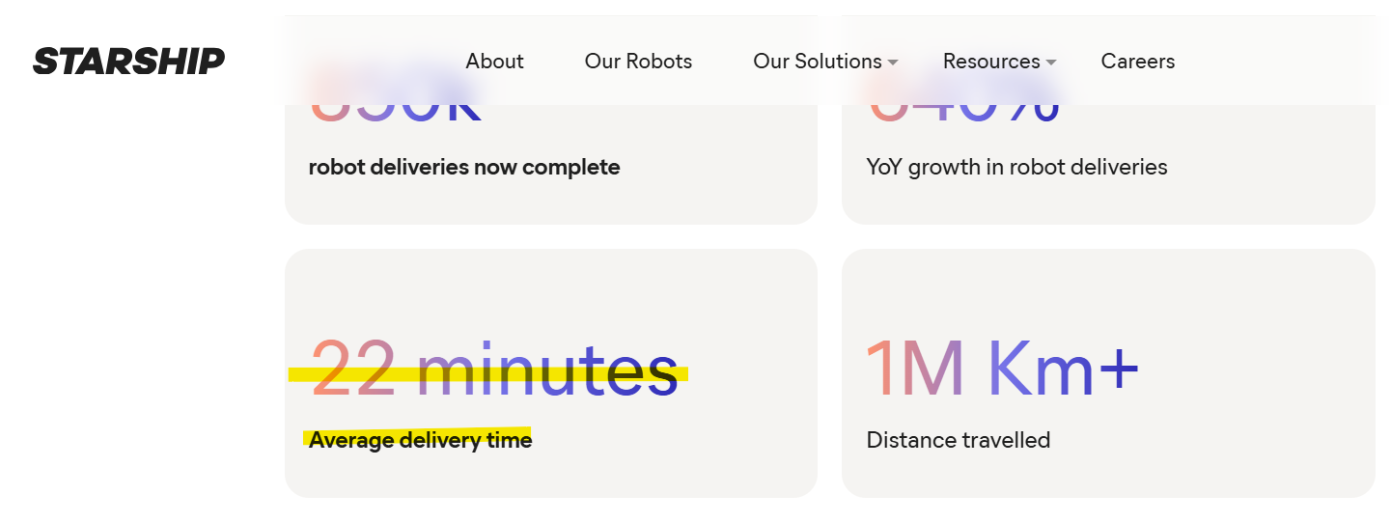}
    \caption{Source data for Avg Delivery Run Time ($T_{\text{AvgDeliveryTime}}$)}
    \label{fig:t_avgdeliverytime_source}
\end{figure}

\begin{figure}[htbp!]
    \centering
    \includegraphics[width=0.8\textwidth]{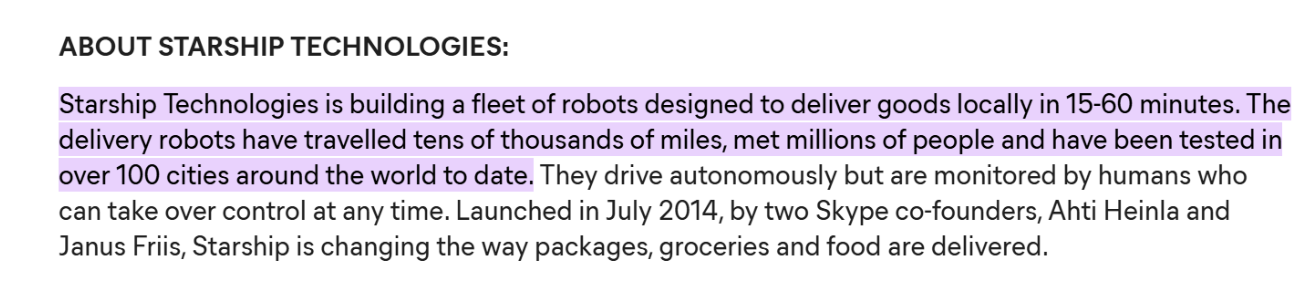}
    \caption{Source data for Max Delivery Run Time 1 ($T_{\text{MaxDeliveryTime}}$)}
    \label{fig:t_maxdeliverytime_1_source}
\end{figure}

\begin{figure}[htbp!]
    \centering
    \includegraphics[width=0.8\textwidth]{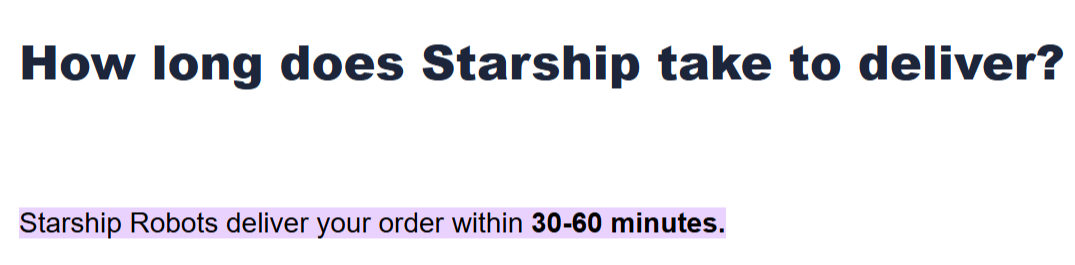}
    \caption{Source data for Max Delivery Run Time 2 ($T_{\text{MaxDeliveryTime}}$)}
    \label{fig:t_maxdeliverytime_2_source}
\end{figure}

\begin{figure}[htbp!]
    \centering
    \includegraphics[width=0.8\textwidth]{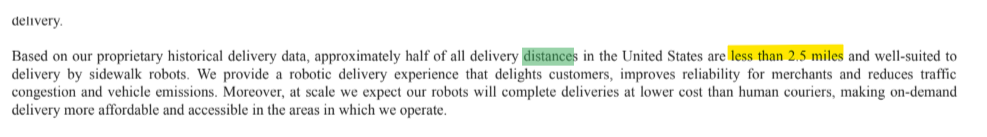}
    \caption{Source data for Avg distance per Run ($T_{\text{AvgDeliveryDistance}}$)}
    \label{fig:t_avgdeliverydistance_source}
\end{figure}

\begin{figure}[htbp!]
    \centering
    \includegraphics[width=0.8\textwidth]{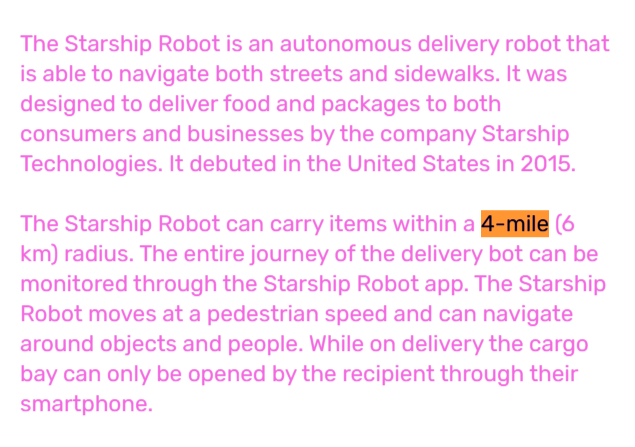}
    \caption{Source data for Max Delivery Distance ($C_{\text{AvgDeliveryDistance}}$)}
    \label{fig:c_avgdeliverydistance_source}
\end{figure}

\begin{figure}[htbp!]
    \centering
    \includegraphics[width=0.8\textwidth]{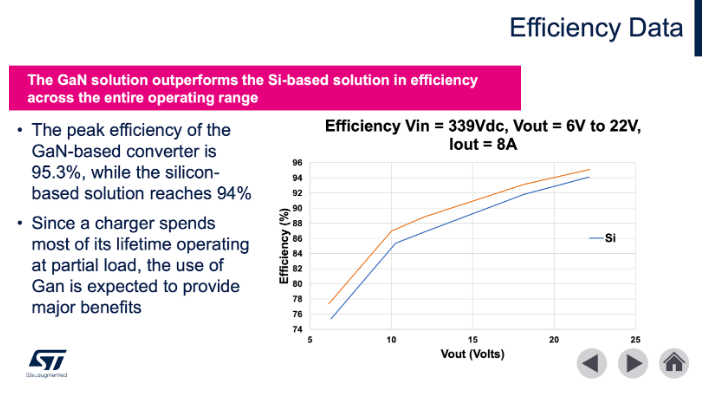}
    \caption{Source data for Energy Convert  ($C_{\text{ElectroMechanicalEff}}$) : $\eta_{\text{charge}}$}
    \label{fig:c_energyconvert_source_1}
\end{figure}

\begin{figure}[htbp!]
    \centering
    \includegraphics[width=0.8\textwidth]{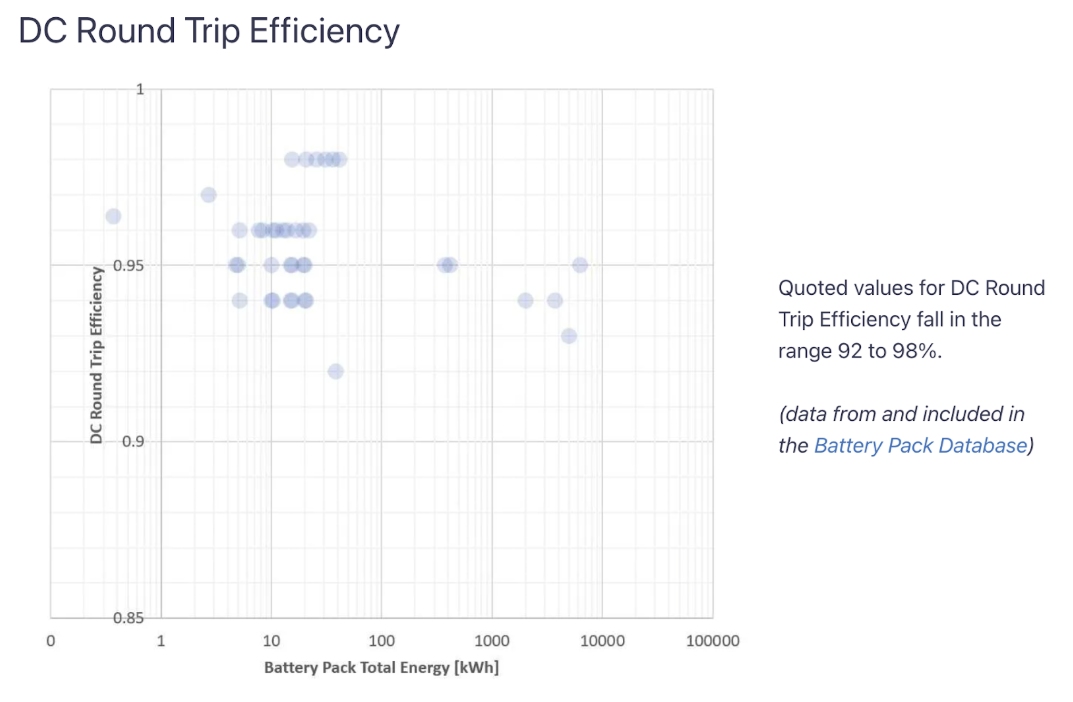}
    \caption{Source data for Energy Convert ($C_{\text{ElectroMechanicalEff}}$) : $\eta_{\text{batteryroundtrip}}$}
    \label{fig:c_energyconvert_source_2}
\end{figure}

\begin{figure}[htbp!]
    \centering
    \includegraphics[width=0.8\textwidth]{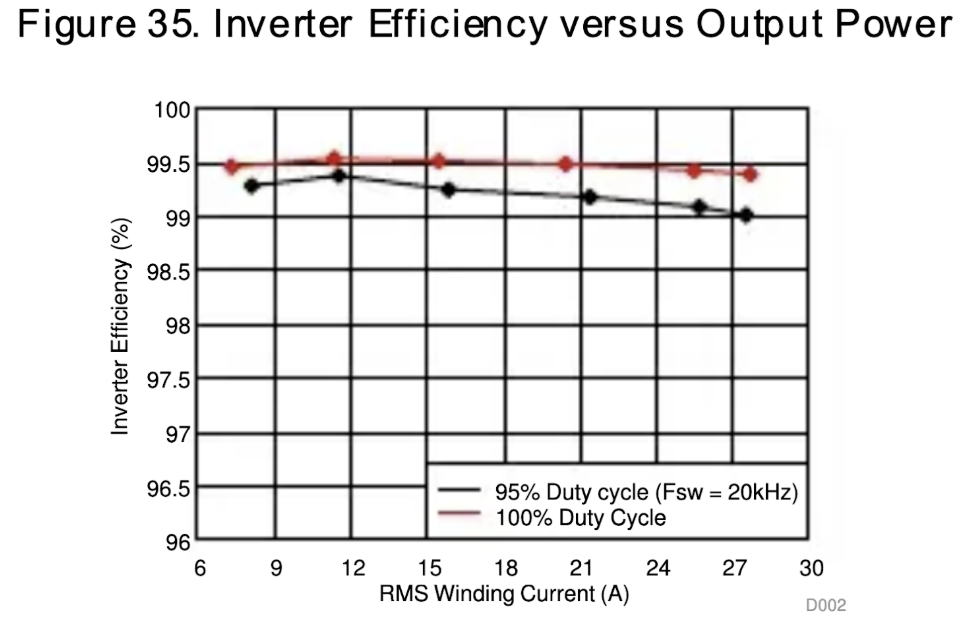}
    \caption{Source data for Energy Convert ($C_{\text{ElectroMechanicalEff}}$) : $\eta_{\text{inverter}}$}
    \label{fig:c_energyconvert_source_3}
\end{figure}

\begin{figure}[htbp!]
    \centering
    \includegraphics[width=0.8\textwidth]{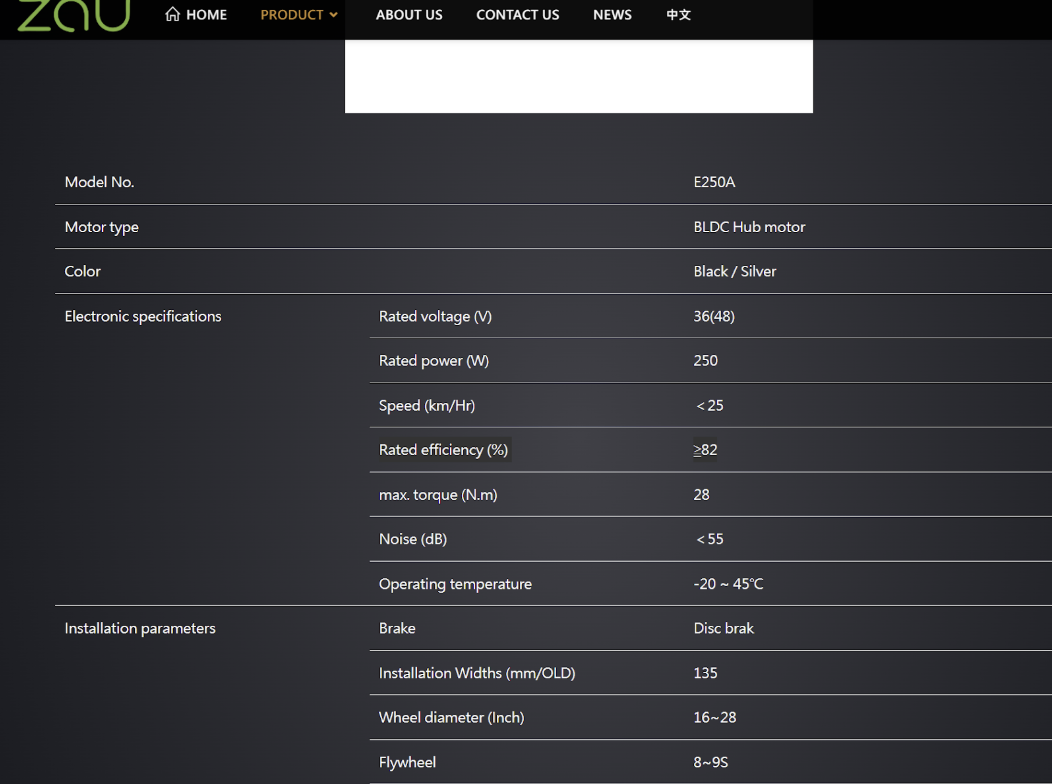}
    \caption{Source data for Energy Convert ($C_{\text{ElectroMechanicalEff}}$) : $\eta_{\text{motor}}$}
    \label{fig:c_energyconvert_source_4}
\end{figure}

\clearpage

\section{Baseline Details}
\label{sec:baseline_details}

Table~\ref{tab:baselines} summarizes the sensor inputs, pretraining data, CostNav finetuning, and goal conditioning for the seven evaluated baselines: the rule-based Nav2~\cite{nav2} variants with AMCL~\cite{amcl} and GPS-only localization, and the learning-based GNM~\cite{shah2023gnm}, ViNT~\cite{shah2023vint}, NoMaD~\cite{sridhar2023nomad}, NavDP~\cite{cai2025navdp}, and CANVAS~\cite{choi2025canvas} policies. The remaining subsections detail each configuration.

\begin{table}[t]
    \centering
    \small
    \setlength{\tabcolsep}{4pt}%
    \caption{\textbf{Baseline Configurations for Simulation Evaluation.} Sensor inputs, pretraining data, additional finetuning on our \SDataCollectionTime\ hr CostNav teleoperation dataset, and goal conditioning for the seven evaluated baselines in the simulation experiments.}
    \label{tab:baselines}
    \resizebox{\textwidth}{!}{%
    \begin{tabular}{@{}lcccccccc@{}}
        \toprule
                                & \multicolumn{2}{c}{\textbf{Rule-Based}}                                & \multicolumn{5}{c}{\textbf{Learning-Based}}                                                                  \\
        \cmidrule(lr){2-3} \cmidrule(lr){4-8}
                                & \textbf{Nav2 w/ AMCL}                & \textbf{Nav2 w/ GPS}            & \textbf{GNM}  & \textbf{ViNT} & \textbf{NoMaD}  & \textbf{NavDP}    & \textbf{CANVAS} \\
        \midrule
        Sensor Inputs            & LiDAR, GPS                           & LiDAR, GPS                      & RGB           & RGB           & RGB             & RGB, Depth        & RGB, GPS                \\
        Pretraining Data Size    & --                                   & --                              & 70 hr (real)  & 80 hr (real)  & 100+ hr (real)  & 363 km (sim)      & 48 hr / 219 km (sim)    \\
        Finetuned on CostNav     & \xmark                               & \xmark                          & \cmark        & \cmark        & \cmark          & \cmark            & \cmark                  \\
        Goal Conditioning        & Point                                & Point                           & Topo. Image   & Topo. Image   & Topo. Image     & Image + Point     & Sketch                  \\
        \bottomrule
    \end{tabular}%
    }
\end{table}

\subsection{Rule-Based Navigation Baseline}
\label{sec:rule_based_baseline}

Our rule-based baseline uses Nav2~\cite{nav2}, which performs classical path planning over a prebuilt 2D occupancy map and uses LiDAR-based costmaps for online obstacle avoidance. We evaluate it under two localization configurations: AMCL~\cite{amcl}, which fuses LiDAR scan matching against the prior map with GPS for global drift correction, and a GPS-only variant that omits the scan-matching step and relies solely on GPS measurements.

\subsection{Learning-Based Baselines: Pretraining Source Datasets}
\label{sec:pretraining_sources}

Table~\ref{tab:pretraining_sources} lists the source datasets used to pretrain each learning baseline. GNM, ViNT, and NoMaD share many of the same real-world multi-robot navigation datasets, while NavDP and CANVAS each use a single self-collected simulation dataset.

\begin{table}[h]
    \centering
    \small
    \setlength{\tabcolsep}{4pt}
    \caption{\textbf{Pretraining Source Datasets per Learning Baseline.} \cmark\ indicates that the source dataset is included in the baseline's pretraining corpus, as reported in the original publications.}
    \label{tab:pretraining_sources}
    \begin{tabular}{@{}lccccc@{}}
        \toprule
        \textbf{Source Dataset} & \textbf{GNM} & \textbf{ViNT} & \textbf{NoMaD} & \textbf{NavDP} & \textbf{CANVAS}      \\
        \midrule
        GoStanford              & \cmark       & \cmark        & \cmark         & --             & --                   \\
        RECON                   & \cmark       & \cmark        & \cmark         & --             & --                   \\
        CoryHall                & \cmark       & \cmark        & \cmark         & --             & --                   \\
        Berkeley                & \cmark       & \cmark        & \cmark         & --             & --                   \\
        SCAND-S/J               & \cmark       & \cmark        & \cmark         & --             & --                   \\
        Seattle                 & \cmark       & \cmark        & \cmark         & --             & --                   \\
        TartanDrive             & \cmark       & \cmark        & \cmark         & --             & --                   \\
        NeBula                  & \cmark       & \cmark        & \cmark         & --             & --                   \\
        SACSoN                  & --           & \cmark        & \cmark         & --             & --                   \\
        BDD                     & --           & \cmark        & --             & --             & --                   \\
        NavDP Sim               & --           & --            & --             & \cmark         & --                   \\
        COMMAND                 & --           & --            & --             & --             & \cmark               \\
        \midrule
        \textbf{Total Coverage} & 70 hr (real) & 80 hr (real)  & 100+ hr (real) & 363 km (sim)   & 48 hr / 219 km (sim) \\
        \bottomrule
    \end{tabular}
\end{table}

\subsection{Finetuning Procedure}
\label{sec:finetuning_procedure}

All five learning baselines are finetuned on our \SDataCollectionTime\ hr CostNav teleoperation dataset starting from each baseline's official pretrained checkpoint. The training code, hyperparameter configurations, and dataset preprocessing scripts are provided in the supplementary materials.

\subsection{Goal Conditioning Specifications}
\label{sec:goal_conditioning_details}

Each baseline accepts a distinct goal modality:
\begin{itemize}
    \item \textbf{Nav2 (AMCL, GPS-only).} A point goal expressed as $(x, y)$ in the map frame.
    \item \textbf{GNM, ViNT, NoMaD.} An image goal sampled from a topological map of the environment along the shortest path to the destination. The policy is conditioned on a sequence of intermediate sub-goal images that guides long-horizon navigation.
    \item \textbf{NavDP.} An image goal paired with an $(x, y)$ point goal in the local frame.
    \item \textbf{CANVAS.} A sketch rendered on an orthographic top-down view of the environment, similar to a satellite map, tracing the simulation shortest path to the destination.
\end{itemize}

\clearpage
\raggedbottom

\begin{landscape}

\section{Simulation Details}
\label{sec:simulation_evaluation}

The following table presents the complete simulation evaluation metrics, including symbols, simulated values, and units.
{
    \footnotesize
    \setlength{\tabcolsep}{2.5pt}
    \begin{longtable}{@{}p{0.08\linewidth} l r r r r r r r r l >{\raggedright\arraybackslash}p{0.14\linewidth}@{}}
        \caption{Simulation Evaluation Metrics} \label{tab:sim_eval_metrics} \\
        \toprule
        \textbf{Variable} & \textbf{Symbol} & \textbf{AMCL} &
        \textbf{GPS} & \textbf{GNM} & \textbf{ViNT} & \textbf{NoMaD} & \textbf{NavDP} & \textbf{CANVAS} & \textbf{Common} & \textbf{Unit} & \textbf{Rationale} \\
        \midrule
        \endfirsthead

        \multicolumn{12}{c}%
        {{\bfseries \tablename\ \thetable{} -- continued from previous page}} \\
        \toprule
        \textbf{Variable} & \textbf{Symbol} & \textbf{AMCL} &
        \textbf{GPS} & \textbf{GNM} & \textbf{ViNT} & \textbf{NoMaD} & \textbf{NavDP} & \textbf{CANVAS} & \textbf{Common} & \textbf{Unit} & \textbf{Rationale} \\
        \midrule
        \endhead

        \midrule
        \multicolumn{12}{r}{{Continued on next page}} \\
        \bottomrule
        \endfoot

        \bottomrule
        \endlastfoot

        \multicolumn{12}{l}{\textit{Physical Metrics}} \\[4pt]
        Avg Velocity & $S_{\text{AvgVelocity}}$ & \SAvgVelocityAMCL & \SAvgVelocityGPS & \SAvgVelocityGNM & \SAvgVelocityViNT & \SAvgVelocityNoMaD & \SAvgVelocityNavDP & \SAvgVelocityCANVAS & & m/s & Measured from simulation evaluation episodes \\[4pt]
        \hdashline[0.5pt/2pt]
        \noalign{\vskip 4pt}
        Avg Mechanical Power & $S_{\text{AvgMechanicalPower}}$ & \SAvgMechanicalPowerAMCL & \SAvgMechanicalPowerGPS & \SAvgMechanicalPowerGNM & \SAvgMechanicalPowerViNT & \SAvgMechanicalPowerNoMaD & \SAvgMechanicalPowerNavDP & \SAvgMechanicalPowerCANVAS & & kW & $C_{\text{RollingResistanceForce}} \cdot S_{\text{AvgVelocity}}$ \\[4pt]
        \hdashline[0.5pt/2pt]
        \noalign{\vskip 4pt}
        Avg Collision Impulse & $S_{\text{CollisionImpulse}}$ & \SCollisionImpulseAMCL & \SCollisionImpulseGPS & \SCollisionImpulseGNM & \SCollisionImpulseViNT & \SCollisionImpulseNoMaD & \SCollisionImpulseNavDP & \SCollisionImpulseCANVAS & & N$\cdot$s & -- \\[4pt]
        \hdashline[0.5pt/2pt]
        \noalign{\vskip 4pt}
        Avg Collision $\Delta v$ & $S_{\text{CollisionDeltaV}}$ & \SCollisionDeltaVAMCL & \SCollisionDeltaVGPS & \SCollisionDeltaVGNM & \SCollisionDeltaVViNT & \SCollisionDeltaVNoMaD & \SCollisionDeltaVNavDP & \SCollisionDeltaVCANVAS & & m$/$s & -- \\[4pt]
        \midrule

        \multicolumn{12}{l}{\textit{Episode Arrival Rates}} \\[4pt]
        SLA Compliance & $S_{\text{EpisodeTermSLA}}$ & \SEpisodeTermSLAAMCL & \SEpisodeTermSLAGPS & \SEpisodeTermSLAGNM & \SEpisodeTermSLAViNT & \SEpisodeTermSLANoMaD & \SEpisodeTermSLANavDP & \SEpisodeTermSLACANVAS & & /run & Safe delivery with intact food \\[4pt]
        \hdashline[0.5pt/2pt]
        \noalign{\vskip 4pt}
        Food Spoiled & $S_{\text{EpisodeTermSpoiled}}$ & \SEpisodeTermSpoiledAMCL & \SEpisodeTermSpoiledGPS & \SEpisodeTermSpoiledGNM & \SEpisodeTermSpoiledViNT & \SEpisodeTermSpoiledNoMaD & \SEpisodeTermSpoiledNavDP & \SEpisodeTermSpoiledCANVAS & & /run & Delivery Completed but food spoiled \\[4pt]
        \hdashline[0.5pt/2pt]
        \noalign{\vskip 4pt}
        Timeout & $S_{\text{EpisodeTermTimeoutRobot}}$ & \SEpisodeTermTimeoutRobotAMCL & \SEpisodeTermTimeoutRobotGPS & \SEpisodeTermTimeoutRobotGNM & \SEpisodeTermTimeoutRobotViNT & \SEpisodeTermTimeoutRobotNoMaD & \SEpisodeTermTimeoutRobotNavDP & \SEpisodeTermTimeoutRobotCANVAS & & /run & Delivery not completed within timeout limit \\[4pt]
        \hdashline[0.5pt/2pt]
        \noalign{\vskip 4pt}
        Physical Assistance & $S_{\text{EpisodeTermPhysicalAssistance}}$ & \SEpisodeTermPhysicalAssistanceAMCL & \SEpisodeTermPhysicalAssistanceGPS & \SEpisodeTermPhysicalAssistanceGNM & \SEpisodeTermPhysicalAssistanceViNT & \SEpisodeTermPhysicalAssistanceNoMaD & \SEpisodeTermPhysicalAssistanceNavDP & \SEpisodeTermPhysicalAssistanceCANVAS & & /run & Bad Robot Orientation or Collision Impulse Health Depletion derived from $S_{\text{RobotMaxMomentum}}$ \\[4pt]
        \midrule

        \multicolumn{12}{l}{\textit{Safety \& Contact Events}} \\[4pt]
        MAIS injury adjustment & $K_{\text{InjuryAdjustment}}$ & & & & & & & &
        \num[round-mode=places, round-precision=4]{\KInjuryAdjustment} & -- & Derived: $C_{\text{RobotWeight}}$ / $P_{\text{MAISReportVehicleWeight}}$ \\[4pt]
        \hdashline[0.5pt/2pt]
        \noalign{\vskip 4pt}
        Pedestrian Injury Cost & $S_{\text{PedInjury}}$ & \SPedInjuryAMCL & \SPedInjuryGPS & \SPedInjuryGNM & \SPedInjuryViNT & \SPedInjuryNoMaD & \SPedInjuryNavDP & \SPedInjuryCANVAS & & \$/run & Derived: $\sum_{\text{AIS}} P(\text{AIS} \mid \Delta v) \cdot P_{\text{PedDamageAIS}} \cdot K_{\text{InjuryAdjustment}}$, where $\Delta v$ from simulation, $P(\text{AIS} \mid \Delta v)$ from~\cite{wang2022_mais0508_deltav} \\[4pt]
        \hdashline[0.5pt/2pt]
        \noalign{\vskip 4pt}
        Property Contact & $S_{\text{PropContact}}$ & & & & & & & & & event/run & Count of collisions greater than 100 $N \cdot s$ \\[4pt]
        -- Bollard & $S_{\text{PropBollardContact}}$ & \SPropBollardContactAMCL & \SPropBollardContactGPS & \SPropBollardContactGNM & \SPropBollardContactViNT & \SPropBollardContactNoMaD & \SPropBollardContactNavDP & \SPropBollardContactCANVAS & & event/run & -- \\[4pt]
        -- Building Glass & $S_{\text{PropBuildingGlassContact}}$ & \SPropBuildingGlassContactAMCL & \SPropBuildingGlassContactGPS & \SPropBuildingGlassContactGNM & \SPropBuildingGlassContactViNT & \SPropBuildingGlassContactNoMaD & \SPropBuildingGlassContactNavDP & \SPropBuildingGlassContactCANVAS & & event/run & -- \\[4pt]
        -- Mail Box & $S_{\text{PropMailBoxContact}}$ & \SPropMailBoxContactAMCL & \SPropMailBoxContactGPS & \SPropMailBoxContactGNM & \SPropMailBoxContactViNT & \SPropMailBoxContactNoMaD & \SPropMailBoxContactNavDP & \SPropMailBoxContactCANVAS & & event/run & -- \\[4pt]
        -- Trash Bin & $S_{\text{PropTrashBinContact}}$ & \SPropTrashBinContactAMCL & \SPropTrashBinContactGPS & \SPropTrashBinContactGNM & \SPropTrashBinContactViNT & \SPropTrashBinContactNoMaD & \SPropTrashBinContactNavDP & \SPropTrashBinContactCANVAS & & event/run & -- \\[4pt]
        \midrule

        \multicolumn{12}{l}{\textit{Map \& Performance Parameters}} \\[4pt]
        Max Deliv Robot Speed & $S_{\text{RobotSpeed}}$ & & & & & & & & \SRobotSpeed & km/h & Based on actual robot speed $C_{\text{RobotSpeed}}$ \\[4pt]
        \hdashline[0.5pt/2pt]
        \noalign{\vskip 4pt}
        Eval Timeout per delivery & $S_{\text{Timeout}}$ & & & & & & & & \STimeout & hr/run & Derived: $T_{\text{MaxDeliveryTime}} \cdot \frac{S_{\text{MaxDeliveryDistance}}}{C_{\text{MaxDeliveryDistance}}} $ \\[4pt]
        \hdashline[0.5pt/2pt]
        \noalign{\vskip 4pt}
        Avg Runtime per delivery & $S_{\text{AvgRunTime}}$ & \num[round-mode=places, round-precision=4]\SAvgRunTimeAMCL & \num[round-mode=places, round-precision=4]\SAvgRunTimeGPS & \num[round-mode=places, round-precision=4]\SAvgRunTimeGNM & \num[round-mode=places, round-precision=4]\SAvgRunTimeViNT & \num[round-mode=places, round-precision=4]\SAvgRunTimeNoMaD & \num[round-mode=places, round-precision=4]\SAvgRunTimeNavDP & \num[round-mode=places, round-precision=4]\SAvgRunTimeCANVAS & & hr/run & Measured from simulation evaluation episodes \\[4pt]
        \hdashline[0.5pt/2pt]
        \noalign{\vskip 4pt}
        Avg Delivery Distance & $S_{\text{AvgDeliveryDistance}}$ & \SAvgDeliveryDistanceAMCL & \SAvgDeliveryDistanceGPS & \SAvgDeliveryDistanceGNM & \SAvgDeliveryDistanceViNT & \SAvgDeliveryDistanceNoMaD & \SAvgDeliveryDistanceNavDP & \SAvgDeliveryDistanceCANVAS & & km/run & Measured from simulation evaluation episodes \\[4pt]
        \hdashline[0.5pt/2pt]
        \noalign{\vskip 4pt}
        Max Delivery Distance & $S_{\text{MaxDeliveryDistance}}$ & & & & & & & & \SMaxDeliveryDistance & km/run & Maximum taxi distance of 200\,m $\times$ 200\,m map \\[4pt]
        \hdashline[0.5pt/2pt]
        \noalign{\vskip 4pt}
        Robot Max Momentum & $S_{\text{RobotMaxMomentum}}$ & & & & & & & & \SRobotMaxMomentum & kg$\cdot$km/h & $C_{\text{RobotWeight}} \cdot C_{\text{RobotMaxSpeed}}$ \\[4pt]
        \midrule
        \multicolumn{12}{l}{\textit{Data Collection Parameters}} \\[4pt]
        Data Collector Working Time & $S_{\text{DataCollectorWorkingTime}}$ & & & & & & & & \SDataCollectorWorkingTime & hr & \\[4pt]
        \hdashline[0.5pt/2pt]
        \noalign{\vskip 4pt}
        Collected Data Time & $S_{\text{DataCollectionTime}}$ & & & & & & & & \SDataCollectionTime & hr & \\[4pt]
        \hdashline[0.5pt/2pt]
        \noalign{\vskip 4pt}
        Data Collection Efficiency & $S_{\text{DataCollectionEff}}$ & & & & & & & & \num[round-mode=places, round-precision=4]{\SDataCollectionEff} & unitless & $\frac{S_{\text{DataCollectionTime}}}{S_{\text{DataCollectorWorkingTime}}}$ \\[4pt]
        
    
    \end{longtable}
}
\end{landscape}

\clearpage

\begin{landscape}

\section{Aggregated Cost Formula}
\label{sec:cost_formula}
{
\footnotesize
\setlength{\tabcolsep}{1.5pt}
\renewcommand{\arraystretch}{1.2}

\begin{longtable}{@{}l l l r r r r r r r l >{\raggedright\arraybackslash}p{0.13\linewidth}@{}}
    \caption{Aggregated Cost Formula}
    \label{tab:cost_formula} \\

    \toprule
    \textbf{Category} & \textbf{Variable} & \textbf{Symbol} &
    \textbf{AMCL} & \textbf{GPS} & \textbf{GNM} & \textbf{ViNT} & \textbf{NoMaD} & \textbf{NavDP} & \textbf{CANVAS} &
    \textbf{Unit} & \textbf{Reference/Rationale} \\
    \midrule
    \endfirsthead

    \multicolumn{12}{c}{{\bfseries \tablename\ \thetable{} -- continued from previous page}} \\
    \toprule
    \textbf{Category} & \textbf{Variable} & \textbf{Symbol} &
    \textbf{AMCL} & \textbf{GPS} & \textbf{GNM} & \textbf{ViNT} & \textbf{NoMaD} & \textbf{NavDP} & \textbf{CANVAS} &
    \textbf{Unit} & \textbf{Reference/Rationale} \\
    \midrule
    \endhead

    \midrule
    \multicolumn{12}{r}{{Continued on next page}} \\
    \bottomrule
    \endfoot

    \bottomrule
    \endlastfoot

    CAPEX & Hardware Cost & $C_{\text{Hardware}}$
    & \CHardware
    & \CHardwareGPS & \CHardwareGNM & \CHardwareViNT & \CHardwareNoMaD & \CHardwareNavDP & \CHardwareCANVAS & \$ &
    Nav2: $P_{\text{Robot}} + P_{\text{LiDAR}} + P_{\text{GPS}}$; CANVAS: $P_{\text{Robot}} + P_{\text{GPS}}$; Others: $P_{\text{Robot}}$ \\[4pt]
    \hdashline[0.5pt/2pt]
    \noalign{\vskip 4pt}
    CAPEX & Data Collection Cost & $C_{\text{DataCollection}}$
    & -- & -- & \CDataCollection & \CDataCollection & \CDataCollection & \CDataCollection & \CDataCollection & \$ &
    $P_{\text{DataCollector}} \cdot S_{\text{DataCollectorWT}}$ \\[4pt]
    \hdashline[0.5pt/2pt]
    \noalign{\vskip 4pt}
    OPEX & Electricity Cost & $C_{\text{ElectricityRun}}$
    & \num[round-mode=places, round-precision=4]\CElectricityRunAMCL
    & \num[round-mode=places, round-precision=4]\CElectricityRunGPS
    & \num[round-mode=places, round-precision=4]\CElectricityRunGNM
    & \num[round-mode=places, round-precision=4]\CElectricityRunViNT
    & \num[round-mode=places, round-precision=4]\CElectricityRunNoMaD
    & \num[round-mode=places, round-precision=4]\CElectricityRunNavDP
    & \num[round-mode=places, round-precision=4]\CElectricityRunCANVAS
    & \$/run &
    $P_{\text{Elec}} \cdot \frac{S_{\text{AvgPower}}}{C_{\text{EnergyConvert}}}
     \cdot S_{\text{AvgRunTime}}$ \\[4pt]
    \hdashline[0.5pt/2pt]
    \noalign{\vskip 4pt}
    OPEX & Service Compensation Cost & $C_{\text{ServiceCompRun}}$
    & \num[round-mode=places, round-precision=2]\CServiceCompRunAMCL
    & \num[round-mode=places, round-precision=2]\CServiceCompRunGPS
    & \num[round-mode=places, round-precision=2]\CServiceCompRunGNM
    & \num[round-mode=places, round-precision=2]\CServiceCompRunViNT
    & \num[round-mode=places, round-precision=2]\CServiceCompRunNoMaD
    & \num[round-mode=places, round-precision=2]\CServiceCompRunNavDP
    & \num[round-mode=places, round-precision=2]\CServiceCompRunCANVAS
    & \$/run &
    $(S_{\text{EpisodeTermSpoiled}})\cdot P_{\text{MktFood}}
     + (S_{\text{EpisodeTermTimeout}}+S_{\text{EpisodeTermPhys}})
     \cdot P_{\text{MktRobotDeli}}$ \\[4pt]
    \hdashline[0.5pt/2pt]
    \noalign{\vskip 4pt}
    OPEX & Pedestrian Safety Cost & $C_{\text{PedestrianRun}}$
    & \num[round-mode=places, round-precision=2]\CPedestrianRunAMCL
    & \num[round-mode=places, round-precision=2]\CPedestrianRunGPS
    & \num[round-mode=places, round-precision=2]\CPedestrianRunGNM
    & \num[round-mode=places, round-precision=2]\CPedestrianRunViNT
    & \num[round-mode=places, round-precision=2]\CPedestrianRunNoMaD
    & \num[round-mode=places, round-precision=2]\CPedestrianRunNavDP
    & \num[round-mode=places, round-precision=2]\CPedestrianRunCANVAS
    & \$/run &
    $S_{\text{PedInjury}}$ \\[4pt]
    \hdashline[0.5pt/2pt]
    \noalign{\vskip 4pt}
    OPEX & Property Damage Cost & $C_{\text{PropertyRun}}$
    & \num[round-mode=places, round-precision=2]\CPropertyRunAMCL
    & \num[round-mode=places, round-precision=2]\CPropertyRunGPS
    & \num[round-mode=places, round-precision=2]\CPropertyRunGNM
    & \num[round-mode=places, round-precision=2]\CPropertyRunViNT
    & \num[round-mode=places, round-precision=2]\CPropertyRunNoMaD
    & \num[round-mode=places, round-precision=2]\CPropertyRunNavDP
    & \num[round-mode=places, round-precision=2]\CPropertyRunCANVAS
    & \$/run &
    $\sum S_{\text{PropContact}} \cdot P_{\text{PropDamage}}$ \\[4pt]
    \hdashline[0.5pt/2pt]
    \noalign{\vskip 4pt}
    OPEX & Repair Cost & $C_{\text{RepairRun}}$
    & \num[round-mode=places, round-precision=2]\CRepairRunAMCL
    & \num[round-mode=places, round-precision=2]\CRepairRunGPS
    & \num[round-mode=places, round-precision=2]\CRepairRunGNM
    & \num[round-mode=places, round-precision=2]\CRepairRunViNT
    & \num[round-mode=places, round-precision=2]\CRepairRunNoMaD
    & \num[round-mode=places, round-precision=2]\CRepairRunNavDP
    & \num[round-mode=places, round-precision=2]\CRepairRunCANVAS
    & \$/run &
    $\frac{P_{\text{Robot}}}{N_{\text{RobotLifeRun}}}
     \cdot C_{\text{Repair}}
     \cdot \frac{S_{\text{EpisodeTermPhys}}}{C_{\text{PhysicalAssistance}}}$ \\[4pt]

     \midrule
     \noalign{\vskip 4pt}
    Total CAPEX & & $C_{\text{CAPEX}}$ & \CCAPEXAMCL & \CCAPEXGPS & \CCAPEXGNM & \CCAPEXViNT & \CCAPEXNoMaD & \CCAPEXNavDP & \CCAPEXCANVAS & \$ & $C_{\text{Hardware}} + C_{\text{DataCollection}}$ \\[4pt]
    \hdashline[0.5pt/2pt]
    \noalign{\vskip 4pt}
    OPEX Per Run & & $C_{\text{OPEX}}$ & \num[round-mode=places, round-precision=2]\COPEXRunAMCL & \num[round-mode=places, round-precision=2]\COPEXRunGPS & \num[round-mode=places, round-precision=2]\COPEXRunGNM & \num[round-mode=places, round-precision=2]\COPEXRunViNT & \num[round-mode=places, round-precision=2]\COPEXRunNoMaD & \num[round-mode=places, round-precision=2]\COPEXRunNavDP & \num[round-mode=places, round-precision=2]\COPEXRunCANVAS & \$/run & $\sum C_{\text{OPEX components}}$ \\[4pt]
    \hdashline[0.5pt/2pt]
    \noalign{\vskip 4pt}
    Revenue Per Run & & $R$ & \num[round-mode=places, round-precision=2]\CRevenueRunAMCL & \num[round-mode=places, round-precision=2]\CRevenueRunGPS & \num[round-mode=places, round-precision=2]\CRevenueRunGNM & \num[round-mode=places, round-precision=2]\CRevenueRunViNT & \num[round-mode=places, round-precision=2]\CRevenueRunNoMaD & \num[round-mode=places, round-precision=2]\CRevenueRunNavDP & \num[round-mode=places, round-precision=2]\CRevenueRunCANVAS & \$/run &
    $P_{\text{MktRobotDeli}} \cdot S_{\text{EpisodeTermSLA}}$ \\[4pt]
    \hdashline[0.5pt/2pt]
    \noalign{\vskip 4pt}
    Contribution Margin & & $C_{\text{ContributionMargin}}$ & \num[round-mode=places, round-precision=2]\CMarginAMCL & \num[round-mode=places, round-precision=2]\CMarginGPS & \num[round-mode=places, round-precision=2]\CMarginGNM & \num[round-mode=places, round-precision=2]\CMarginViNT & \num[round-mode=places, round-precision=2]\CMarginNoMaD & \num[round-mode=places, round-precision=2]\CMarginNavDP & \num[round-mode=places, round-precision=2]\CMarginCANVAS & \$/run &
    $R - C_{\text{OPEX}}$ \\[4pt]
    \hdashline[0.5pt/2pt]
    \noalign{\vskip 4pt}
    Break Even Point & & $\text{BEP}$ & -- & -- & -- & -- & -- & -- & -- & run &
    $\frac{C_{\text{CAPEX}}}{C_{\text{ContributionMargin}}}$ \\[4pt]
    \hdashline[0.5pt/2pt]
    \noalign{\vskip 4pt}
    Break Even Point (days) & & $\text{BEP}_{\text{days}}$ & -- & -- & -- & -- & -- & -- & -- & days &
    $\frac{\text{BEP}}{N_{\text{Delivery}}}$ \\

    \end{longtable}
}
\end{landscape}

\clearpage

\section{Per-Component Break-Even Thresholds}
\label{sec:bep_requirements}

For each OPEX component, we report the rate above which its cost alone exceeds the \num[round-mode=places, round-precision=2]{\PMktRobotDeli}\$/run revenue, assuming all other components are zero.

\begin{itemize}
    \item \textbf{Pedestrian-safety cost} $> \num[round-mode=places, round-precision=2]{\PMktRobotDeli}$\$/run. From Sec.~\ref{sec:pedestrian_cost}, each $1\%$ of contacts at AIS severity $i$ adds \num[round-mode=places, round-precision=2]{\fpeval{0.01*\KInjuryAdjustment*\PPedDamageAISOne}}\$/run ($i{=}1$), \num[round-mode=places, round-precision=2]{\fpeval{0.01*\KInjuryAdjustment*\PPedDamageAISTwo}}\$/run ($i{=}2$), and \num[round-mode=places, round-precision=2]{\fpeval{0.01*\KInjuryAdjustment*\PPedDamageAISThree}}\$/run ($i{=}3$), so the policy is unprofitable above AIS 1 $> 3.7\%$, AIS 2 $> 0.52\%$, or AIS 3 $> 0.12\%$.
    \item \textbf{Spoilage rate} $> 9.8\%$. Each $1\%$ costs $\num[round-mode=places, round-precision=2]{\fpeval{(\PMktFood + \PMktRobotDeli)*0.01}}$\$/run (food refund plus lost delivery-fee revenue), so the break-even rate is $\num[round-mode=places, round-precision=2]{\PMktRobotDeli}/\num[round-mode=places, round-precision=2]{\fpeval{\PMktFood + \PMktRobotDeli}} = 9.8\%$.
    \item \textbf{Combined timeout and physical-assistance rate} $> 50\%$. Each $1\%$ costs $\num[round-mode=places, round-precision=4]{\fpeval{2*\PMktRobotDeli*0.01}}$\$/run (refund $+$ lost revenue), so the break-even rate is $1/2 = 50\%$.
    \item \textbf{Property contact rate} $> \num[round-mode=places, round-precision=1]{\fpeval{100*\PMktRobotDeli/((\PPropBollardDamage + \PPropMailBoxDamage + \PPropTrashBinDamage + \PPropBuildingGlassDamage)/4)}}\%$. The mean damage across mailbox, trash bin, bollard, and building glass is $\num[round-mode=places, round-precision=2]{\fpeval{(\PPropBollardDamage + \PPropMailBoxDamage + \PPropTrashBinDamage + \PPropBuildingGlassDamage)/4}}$\$/contact, so a contact rate above $\num[round-mode=places, round-precision=2]{\PMktRobotDeli}/\num[round-mode=places, round-precision=2]{\fpeval{(\PPropBollardDamage + \PPropMailBoxDamage + \PPropTrashBinDamage + \PPropBuildingGlassDamage)/4}} \approx 3.0\%$ exhausts the budget.
    \item \textbf{Physical-assistance rate} $> 10.6\%$ for repair alone. Repair scales as $(P_{\text{Robot}}/N_{\text{RobotLifeRun}}) \cdot C_{\text{RepairCalc}} \cdot (\text{PA}/C_{\text{PhysicalAssistance}}) \approx \num[round-mode=places, round-precision=2]{\fpeval{(\PRobot/\NRobotLifeRun)*\CRepairCalc/\CPhysicalAssistance}}$\$/run per unit PA, so PA $> \num[round-mode=places, round-precision=2]{\PMktRobotDeli}/\num[round-mode=places, round-precision=2]{\fpeval{(\PRobot/\NRobotLifeRun)*\CRepairCalc/\CPhysicalAssistance}} \approx 10.6\%$ exhausts the budget through repair.
\end{itemize}

\section{Sim-to-Real Experiment: Per-Scenario Details}
\label{sec:sim_to_real_appendix}

\begin{figure}[ht]
    \centering
    \includegraphics[width=\textwidth]{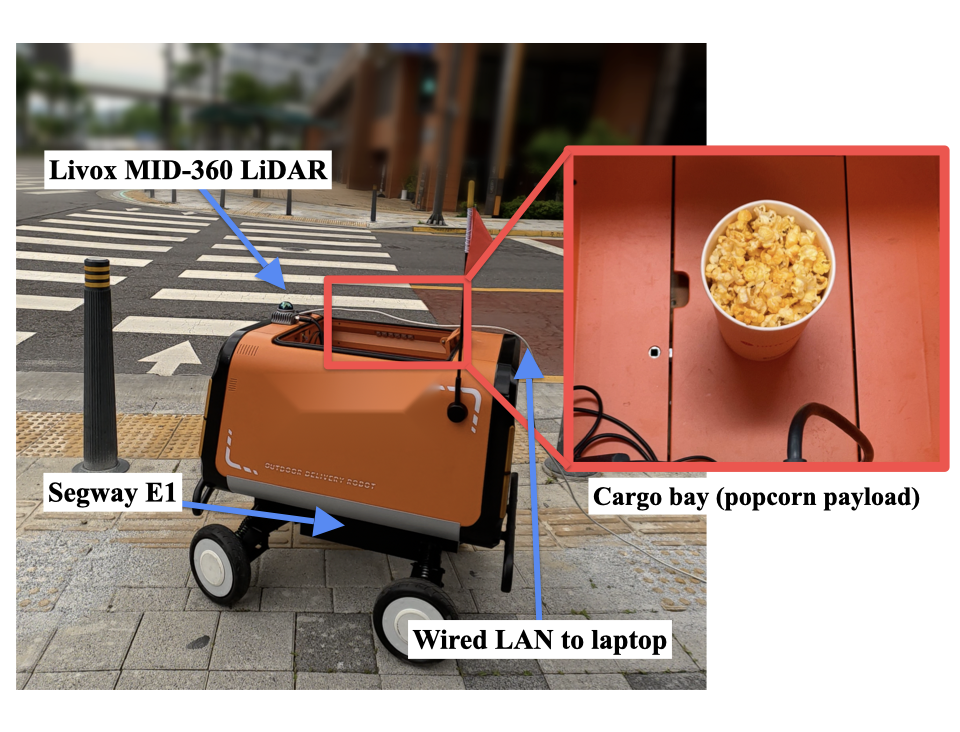}
    \caption{\textbf{Real-world experimental platform.} A Segway E1 sidewalk delivery robot with a Livox MID-360 360$^\circ$\,$\times$\,59$^\circ$ hemispherical LiDAR streams sensor data over a wired LAN tether to an off-board laptop that runs the sim-trained CANVAS policy. The laptop carries an Intel Core Ultra 9 275HX 24-core CPU, \SI{64}{\giga\byte} RAM, \SI{1}{\tera\byte} storage, and an NVIDIA RTX 5090 Laptop GPU. Right inset: cargo bay with a popcorn-cup payload.}
    \label{fig:real_robot_setup}
\end{figure}

We deployed the sim-trained CANVAS policy on the physical platform shown in Figure~\ref{fig:real_robot_setup} along an outdoor sidewalk route in an urban area. \REvalEpisode\ scenarios were attempted; outcomes and physical measurements are summarized below.

Distances are computed from the LiDAR-localized pose stream. Contact occurred only in S3, S5, and S7. The reported impulse for S3 is the directly measured \texttt{/odom} velocity drop multiplied by the 50~kg robot mass. For S5 and S7 a human safety operator intervened to halt the robot before the full collision had run its course, so the raw \texttt{/odom} signal only captures the pre-intervention deceleration and underestimates how severe the un-prevented impact would have been. We therefore use \emph{physics-based estimates} of the would-be impulse---the impulse the collision would have produced had the operator not stopped the robot. For S5 we adopt the same impulse as S3's plant hit ($50$~N$\cdot$s). For S7 we use $100$~N$\cdot$s, corresponding to an upper-bound near-elastic collision of a $50$~kg robot at $\approx 1$~m/s ($2 m v = 100$~N$\cdot$s).
Average mechanical power (\num[round-mode=places, round-precision=4]\RAvgMechanicalPowerCANVAS~kW = \SI{138.3}{\watt}) is the mean battery draw $P=V\cdot I$ over a separate 10-minute teleop run on the same robot.

\begin{figure}[ht]
    \centering
    \includegraphics[width=\textwidth]{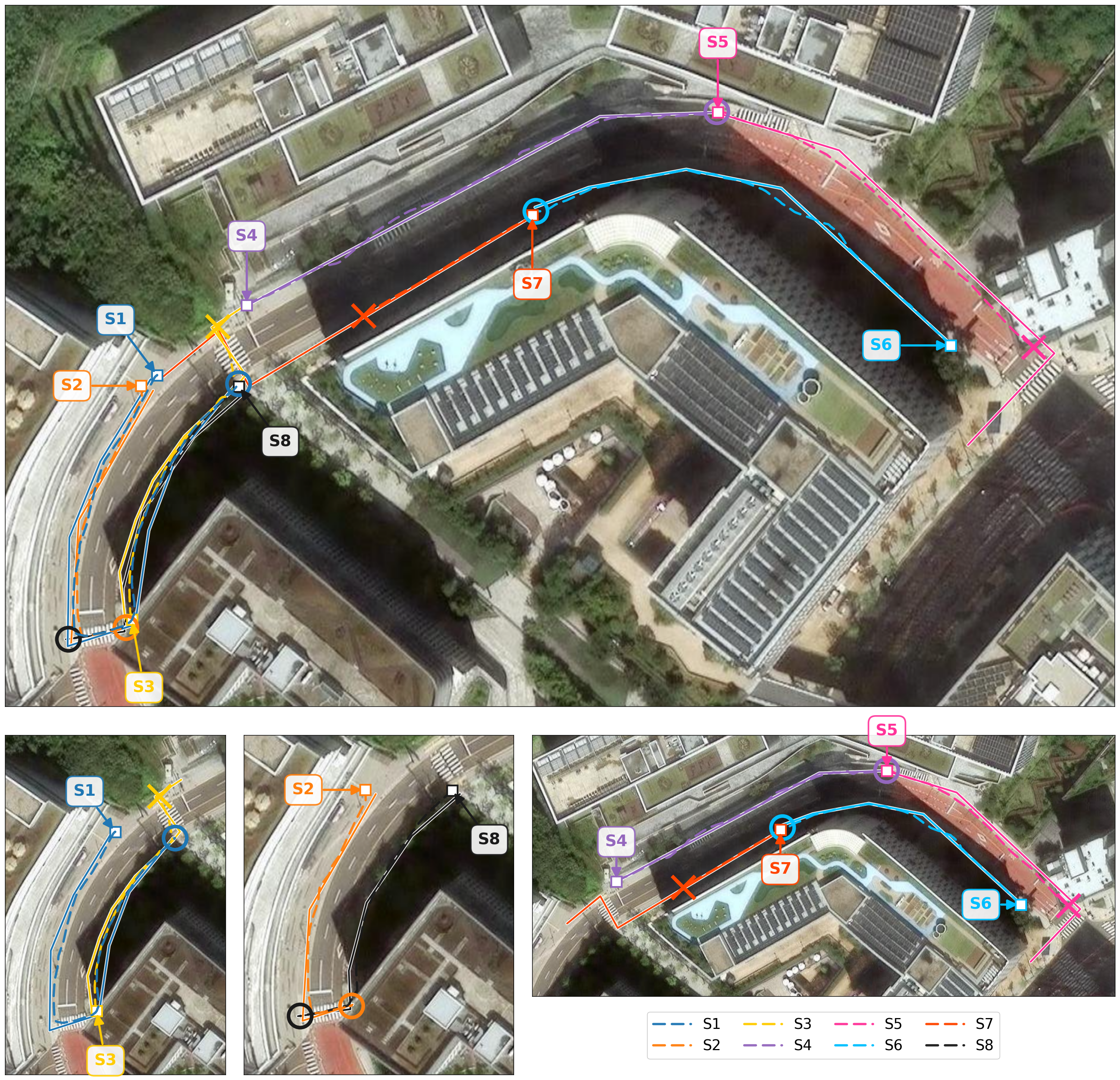}
    \caption{\textbf{Sim-to-Real CANVAS — all 8 scenarios on the outdoor sidewalk route.} Top: overview showing all eight planned and actual paths on the same map. Bottom row, from left to right: zoomed views for (S1, S3), (S2, S8), and (S4, S5, S6, S7). Solid lines: planned waypoint paths. Dashed lines: actual robot path (LiDAR-localized pose). Each scenario's label sits next to its start position. Markers (in the scenario's own color): $\blacksquare$ start, $\bigcirc$ successful arrival, $\times$ crash (S3 plant, S5 plant with spoiled cargo, S7 fence). S3, S5, S7 paths terminate at the impact location rather than at the planned destination.}
    \label{fig:sim_to_real_scenarios}
\end{figure}

\begin{table}[ht]
    \centering
    \small
    \caption{\textbf{Per-Scenario Real-World Results (CANVAS, sim-trained).} S=Success, F=Fail. Contact column lists the obstacle struck. Distance from LiDAR-localized pose. Impulse uses 50~kg robot mass.}
    \label{tab:sim_to_real_per_scenario}
    \resizebox{\textwidth}{!}{%
        \begin{tabular}{@{}lccccccc@{}}
            \toprule
            \textbf{Scenario}                 & \textbf{Outcome}     & \textbf{Contact}     & \textbf{Runtime (s)} & \textbf{Distance (m)} & \textbf{Avg V (m/s)} & \textbf{Avg Impulse (N$\cdot$s)} & \textbf{Avg $\Delta v$ (m/s)} \\
            \midrule
            S1                                & S                    & --                   & \RSDurationOne   & \RSDistanceOne   & \RSAvgVelocityOne   & --                 & --                \\
            S2                                & S                    & --                   & \RSDurationTwo   & \RSDistanceTwo   & \RSAvgVelocityTwo   & --                 & --                \\
            S3                                & F (physical)         & Plant                & \RSDurationThree & \RSDistanceThree & \RSAvgVelocityThree & \RSAvgImpulseThree & \RSAvgDeltaVThree \\
            S4                                & S                    & --                   & \RSDurationFour  & \RSDistanceFour  & \RSAvgVelocityFour  & --                 & --                \\
            S5                                & F (spoiled)          & Plant (+spilled food)& \RSDurationFive  & \RSDistanceFive  & \RSAvgVelocityFive  & \RSAvgImpulseFive  & \RSAvgDeltaVFive  \\
            S6                                & S                    & --                   & \RSDurationSix   & \RSDistanceSix   & \RSAvgVelocitySix   & --                 & --                \\
            S7                                & F (physical)         & Fence                & \RSDurationSeven & \RSDistanceSeven & \RSAvgVelocitySeven & \RSAvgImpulseSeven & \RSAvgDeltaVSeven \\
            S8                                & S                    & --                   & \RSDurationEight & \RSDistanceEight & \RSAvgVelocityEight & --                 & --                \\
            \midrule
            \textbf{Mean}                     &                      &                      & 106.83           & 116.28           & \RAvgVelocityCANVAS & \RCollisionImpulseCANVAS & \RCollisionDeltaVCANVAS \\
            \bottomrule
        \end{tabular}%
    }
\end{table}

\paragraph{Aggregated Variable Reference.}
Table~\ref{tab:real_variable_reference} lists all aggregated \texttt{R*} macros defined for the real-world CANVAS run (mirrors the simulation \texttt{S*} schema in \S\ref{sec:cost_parameters}). These values feed the comparison in Table~\ref{tab:sim_to_real} and are derived from the raw values for each scenario above.

\begin{table}[ht]
    \centering
    \small
    \caption{\textbf{Real-World CANVAS Aggregated Variables.} All values are means over \REvalEpisode\ scenarios. \texttt{R*} prefix denotes real-world (parallel to \texttt{S*} for simulation).}
    \label{tab:real_variable_reference}
    {\small
    \setlength{\tabcolsep}{3pt}
    \begin{tabular}{l p{0.18\textwidth} l r l p{0.30\textwidth}}
        \toprule
        \textbf{Category} & \textbf{Variable} & \textbf{Symbol} & \textbf{Value} & \textbf{Unit} & \textbf{Reference/Rationale} \\
        \midrule
        \emph{Episode Outcomes} & Episodes              & $N_\text{eval}^\text{R}$    & \REvalEpisode                                                       & episodes   & S1--S8                                                                                       \\
        \emph{Episode Outcomes} & SLA Compliance        & $R_\text{SLA}$              & \REpisodeTermSLACANVAS                                              & rate       & S1, S2, S4, S6, S8                                                                            \\
        \emph{Episode Outcomes} & Spoiled               & $R_\text{spoiled}$          & \REpisodeTermSpoiledCANVAS                                          & rate       & S5                                                                                            \\
        \emph{Episode Outcomes} & Timeout               & $R_\text{timeout}$          & \REpisodeTermTimeoutRobotCANVAS                                     & rate       & --                                                                                            \\
        \emph{Episode Outcomes} & Physical Assist       & $R_\text{phys}$             & \REpisodeTermPhysicalAssistanceCANVAS                               & rate       & S3, S7                                                                                        \\
        \midrule
        \emph{Physical Metrics} & Avg Velocity          & $R_\text{AvgVelocity}$      & \RAvgVelocityCANVAS                                                 & m/s        & Mean of linear speed                                                           \\
        \emph{Physical Metrics} & Avg Mechanical Power  & $R_\text{AvgPower}$         & \RAvgMechanicalPowerCANVAS                                          & kW         & $P=V\cdot I$, 10-min teleop                                                                   \\
        \emph{Physical Metrics} & Avg Collision Impulse & $R_\text{CollImpulse}$      & \RCollisionImpulseCANVAS                                            & N$\cdot$s  & $m_\text{robot}\cdot R_\text{CollDeltaV}$                                                     \\
        \emph{Physical Metrics} & Avg Collision $\Delta v$ & $R_\text{CollDeltaV}$    & \RCollisionDeltaVCANVAS                                             & m/s        & See Table~\ref{tab:sim_to_real_per_scenario}                                                  \\
        \midrule
        \emph{Safety Events}    & Pedestrian Injury     & $R_\text{PedInjury}$        & \RPedInjuryCANVAS                                                   & \$/run     & Pedestrians present, no collisions                                                            \\
        \emph{Safety Events}    & Plant Contact         & $R_\text{PropPlant}$        & \RPropPlantContactCANVAS                                            & events/run & S3, S5                                                                                        \\
        \emph{Safety Events}    & Fence Contact         & $R_\text{PropFence}$        & \RPropFenceContactCANVAS                                            & events/run & S7                                                                                            \\
        \midrule
        \emph{Performance}      & Avg Runtime           & $R_\text{AvgRunTime}$       & \num[round-mode=places, round-precision=4]\RAvgRunTimeCANVAS        & hr/run     & Mean episode duration                                                                         \\
        \emph{Performance}      & Avg Distance          & $R_\text{AvgDistance}$      & \num[round-mode=places, round-precision=4]\RAvgDeliveryDistanceCANVAS & km/run  & From LiDAR-localized pose                                                                     \\
        \midrule
        \emph{CAPEX}            & Hardware              & $R_\text{Hardware}$         & \RHardwareCANVAS                                                    & \$         & $P_\text{Robot}+P_\text{Lidar}$                                                                \\
        \emph{CAPEX}            & Data Collection       & $C_\text{DataCollection}$   & \CDataCollection                                                    & \$         & $P_\text{DataCollector}\cdot S_\text{DataCollectorWorkingTime}$                                \\
        \emph{CAPEX}            & Total CAPEX           & $R_\text{CAPEX}$            & \RCAPEXCANVAS                                                       & \$         & $R_\text{Hardware}+C_\text{DataCollection}$                                                    \\
        \midrule
        \emph{OPEX}             & Electricity           & $R_\text{ElectricityRun}$   & \num[round-mode=places, round-precision=4]{\RElectricityRunCANVAS}  & \$/run     & $P_\text{Elec}\cdot(R_\text{AvgPower}/C_\text{EnergyConvert})\cdot R_\text{AvgRunTime}$        \\
        \emph{OPEX}             & Service Compensation  & $R_\text{ServiceCompRun}$   & \num[round-mode=places, round-precision=4]{\RServiceCompRunCANVAS}  & \$/run     & $R_\text{spoiled}\cdot P_\text{MktFood}+(R_\text{timeout}+R_\text{phys})\cdot P_\text{MktRobotDeli}$ \\
        \emph{OPEX}             & Pedestrian Safety     & $R_\text{PedestrianRun}$    & \RPedestrianRunCANVAS                                               & \$/run     & $R_\text{PedInjury}$                                                                          \\
        \emph{OPEX}             & Property Damage       & $R_\text{PropertyRun}$      & \num[round-mode=places, round-precision=2]{\RPropertyRunCANVAS}     & \$/run     & $R_\text{PropPlant}\cdot P_\text{PropPlantDamage}+R_\text{PropFence}\cdot P_\text{PropFenceDamage}$ \\
        \emph{OPEX}             & Repair                & $R_\text{RepairRun}$        & \num[round-mode=places, round-precision=2]{\RRepairRunCANVAS}       & \$/run     & $(P_\text{Robot}/N_\text{RobotLifeRun})\cdot C_\text{Repair}\cdot(R_\text{phys}/C_\text{PhysicalAssistance})$ \\
        \emph{OPEX}             & Total OPEX            & $R_\text{OPEX}$             & \num[round-mode=places, round-precision=2]{\ROPEXRunCANVAS}         & \$/run     & $\sum R_\text{OPEX components}$                                                                \\
        \midrule
        \emph{Profitability}    & Revenue               & $R_\text{RevenueRun}$       & \num[round-mode=places, round-precision=4]{\RRevenueRunCANVAS}      & \$/run     & $P_\text{MktRobotDeli}\cdot R_\text{SLA}$                                                     \\
        \emph{Profitability}    & Contribution Margin   & $R_\text{Margin}$           & \num[round-mode=places, round-precision=2]{\RMarginCANVAS}          & \$/run     & $R_\text{RevenueRun}-R_\text{OPEX}$                                                            \\
        \emph{Profitability}    & BEP                   & $R_\text{BEP}$              & --                                                                  & runs       & $R_\text{CAPEX}/R_\text{Margin}$ (undefined; $R_\text{Margin}<0$)                              \\
        \bottomrule
    \end{tabular}%
    }
\end{table}

\end{document}